\documentclass[smallextended]{svjour3}   
\smartqed

\usepackage{amsmath,amssymb,graphicx}
\usepackage{epsfig}
\usepackage{lscape}
\usepackage{footmisc}
\usepackage{times}
\usepackage{graphicx}
\usepackage{multirow}
\usepackage{lscape}
\newtheorem{de}{Definition}

\begin{document}
\title{On the Role and the Importance of Features for Background Modeling and Foreground Detection}

\author{Thierry Bouwmans  \and Caroline Silva \and Cristina Marghes \and Mohammed Sami Zitouni \and Harish Bhaskar \and Carl Frelicot}

\institute{T. Bouwmans \at
              Lab. MIA, Univ. La Rochelle, France \\
              Tel.: +05.46.45.72.02\\
              \email{tbouwman@univ-lr.fr} 
              \and   
              C. Silva  \at
              Lab. MIA, Univ. La Rochelle, France \\ 
              \email{lolyne.pacheco@gmail.com}
              \and 
              C. Marghes \at
              Movidius \\
              \email{cristina.marghes@gmail.com} 
              \and 
              M. Zitouni \at
              Visual Signal Analysis and Processing (VSAP) Research Center \\
              Department of Electrical and Computer Engineering, Khalifa University, U.A.E
              \email{mohammad.zitouni@kustar.ac.ae}
              \and 
              H. Bhaskar \at
              Visual Signal Analysis and Processing (VSAP) Research Center \\
              Department of Electrical and Computer Engineering, Khalifa University, U.A.E
              \email{harish.bhaskar@kustar.ac.ae}   
              \and 
              C. Frelicot \at
              Lab. MIA, Univ. La Rochelle, France \\
              \email{cfrelicot@univ-lr.fr}        
              }

\date{Received: date / Accepted: date}

\maketitle

\begin{abstract}
Background modeling has emerged as a popular foreground detection technique for various applications in video surveillance. Background modeling methods have become increasing efficient in robustly modeling the background and hence detecting moving objects in any visual scene. Although several background subtraction and foreground detection have been proposed recently, no traditional algorithm today still seem to be able to simultaneously address all the key challenges of illumination variation, dynamic camera motion, cluttered background and occlusion. This limitation can be attributed to the lack of systematic investigation concerning the role and importance of features within background modeling and foreground detection. With the availability of a rather large set of invariant features, the challenge is in determining the best combination of features that would improve accuracy and robustness in detection. The purpose of this study is to initiate a rigorous and comprehensive survey of features used within background modeling and foreground detection. Further, this paper presents a systematic experimental and statistical analysis of techniques that provide valuable insight on the trends in background modeling and use it to draw meaningful recommendations for practitioners. In this paper, a preliminary review of the key characteristics of features based on the types and sizes is provided in addition to investigating their intrinsic spectral, spatial and temporal properties. Furthermore, improvements using statistical and fuzzy tools are examined and techniques based on multiple features are benchmarked against reliability and selection criterion. Finally, a description of the different resources available such as datasets and codes is provided.
\keywords{Background modeling \and Foreground Detection \and Features \and Local Binary Patterns}
\end{abstract}

\section{Introduction}
\label{sec:Introduction}
\begin{figure}
\begin{center}
\includegraphics[width=2.5cm]{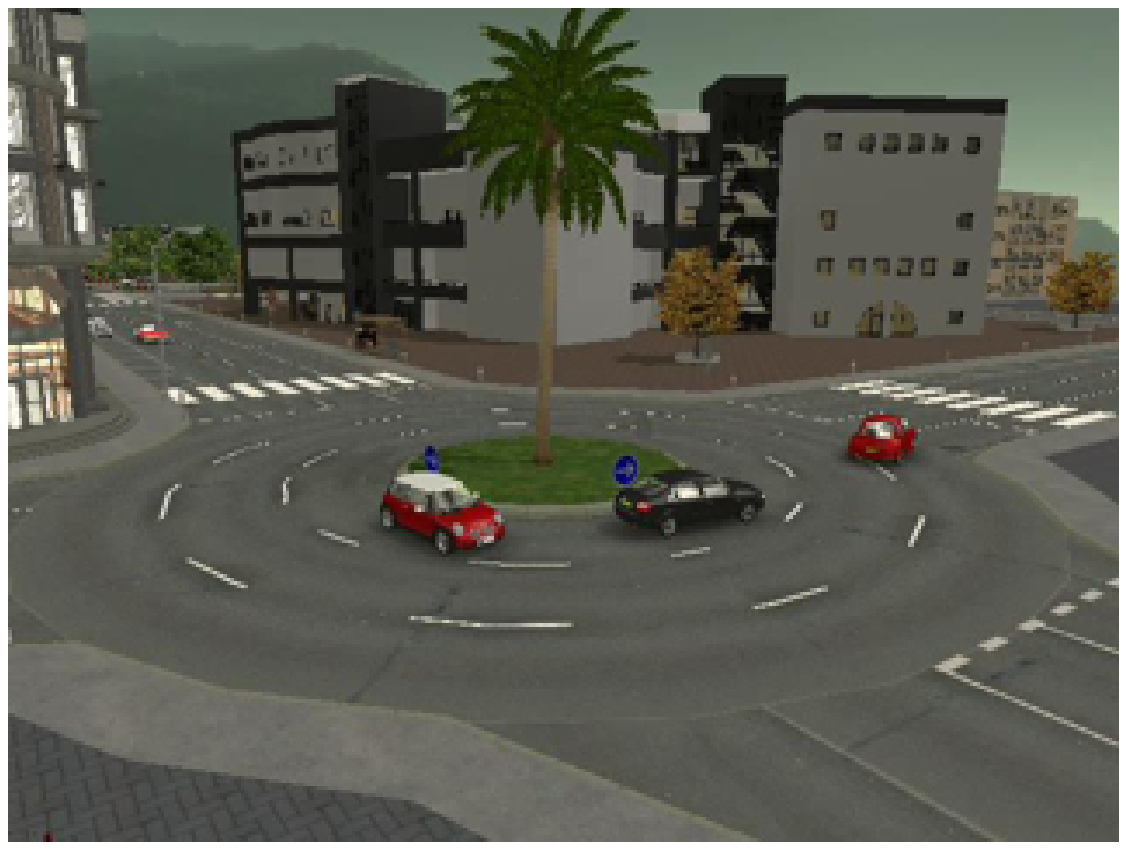}
\includegraphics[width=2.5cm]{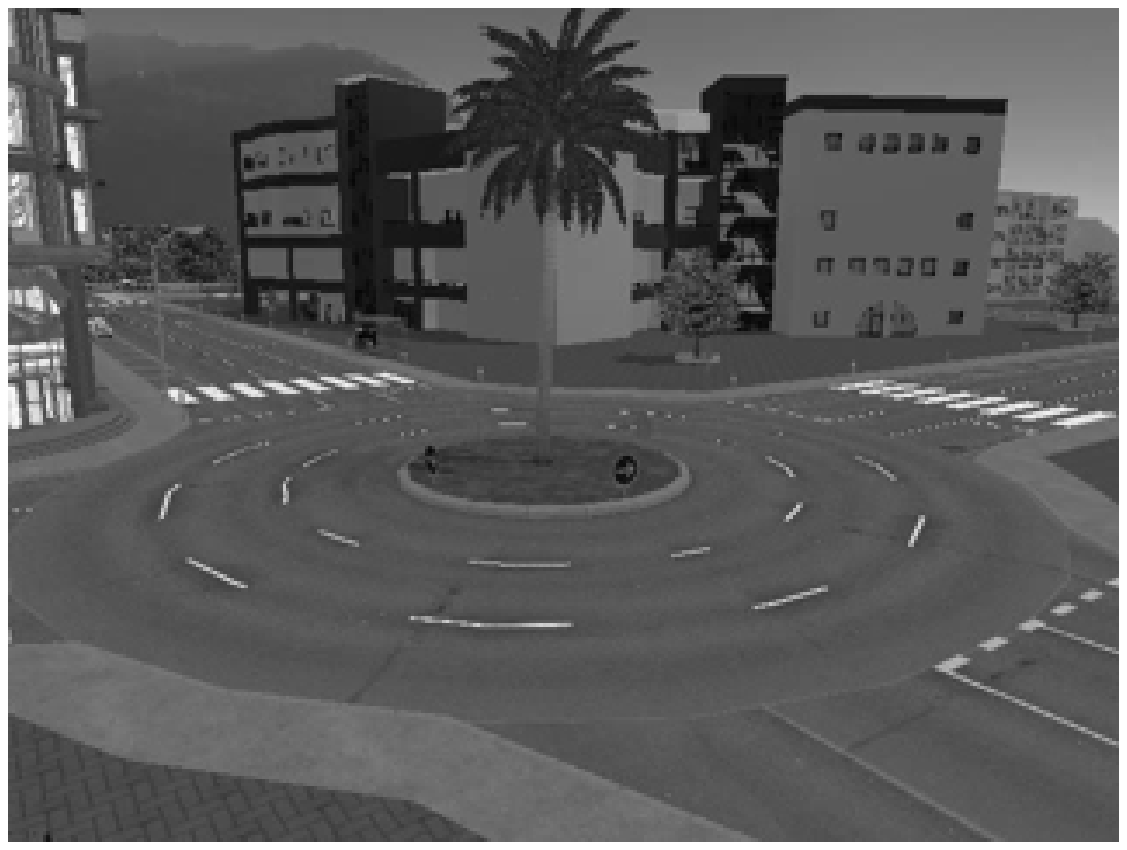}
\includegraphics[width=2.5cm]{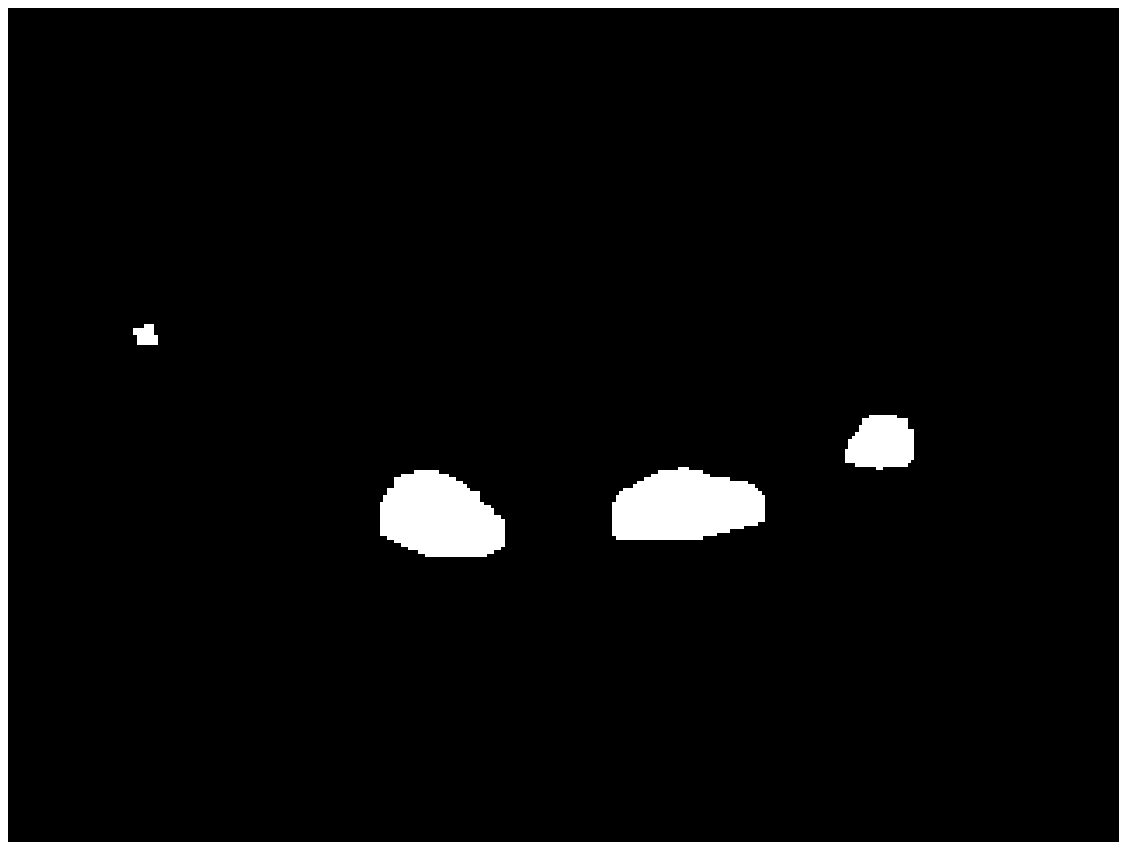} \\
\caption{Background Modeling and Foreground Detection: Original image (309), reconstructed background image, foreground mask (Sequences from BMC 2012 dataset \cite{903}).} 
\label{BMC2012}
\end{center}
\end{figure}

Background modeling and foreground detection are important steps for video processing applications in video-surveillance \cite{1}, optical motion capture \cite{2}, multimedia \cite{3}, teleconferencing and human-computer interface. The aim is to separate the moving objects, called "foreground", from the static information, called "background". For example, Fig. \ref{BMC2012} shows an original frame of
a sequence from the BMC 2012 dataset \cite{903}, the reconstructed background image and the moving objects mask obtained from a decomposition into the low-rank matrix and sparse matrix based model \cite{JournalCOSREV2014-2}. Conventional background modeling methods exploit the temporal variation of each pixel to model the background and hence use it in conjunction with change detection for foreground extraction. The last decade witnessed very significant contributions to this field \cite{JournalRPCS2008}\cite{JournalRPCS2009}\cite{JournalRPCS2011} \cite{JournalCOSREV2014-1}\cite{JournalCOSREV2014-2}\cite{JournalCVIU2014-1}\cite{ChapterHPRCV2010}\cite{ChapterCRC2012}\cite{ChapterINTECH2012} 
\cite{BookCRC2014}. Despite the advances to background modeling and foreground detection, the dynamic nature of visual scenes attributed by changing illumination conditions, occlusion, background clutter and noise have challenged the robustness of such techniques. Under this pretext, focus has shifted towards the investigation of features and their role in improving both the accuracy and robustness of background modeling and foreground detection. Although fundamental low-level features such as color, edge, texture, motion and stereo have reported reasonable success, recent visual applications using mobile devices and internet videos where the background is non-static, require more complex representations to guarantee robust moving object detection \cite{SIMVA2014}. Furthermore, in order to generalize existing background modeling and foreground detection schemes to real-life scenes where dynamic variations are inevitable and the pose of the camera is little known, automatic feature selection, model selection and adaptation for such schemes are often desired. \\ 

\indent Considering the needs and challenges aforementioned, in this paper, a comprehensive review of features used in background modeling and foreground detection is initiated for benchmarking them against the complexities of typical dynamic scenes. Thus, the aim of this survey is then to provide a first complete overview of the role and the importance of features in background modeling and foreground detection by reviewing both existing and new ideas for (1) novices who could be students or engineers beginning in the field of computer vision, (2) experts as we put forward the recent advances that need to be improved, and (3) reviewers to evaluate papers in journals, conferences, and workshop. In addition, this survey gives a complete overview  Moreover, an accompanying website called the Features Website\protect\footnotemark[1] is provided. It allows the reader to have a quick access to the main resources, and codes in the field. So, this survey is intended to be a reference for researchers and developers in industries, as well as graduate students, interested in robust background modeling and foreground detection in challenging environments. \\

\footnotetext[1]{{https://sites.google.com/site/featuresbackgroundforeground/}}

\indent Some of the main contributions of this paper can be summarized as follows:
\begin{itemize}
\item \textbf{A review regarding feature concepts:} A first complete overview of features used in background modeling and foreground detection over
the last decade concerning more than 600 papers. After a preliminary overview on the key concepts in the field of features in Section \ref{sec:PreliminaryOverview}, a survey of spectral features including color features are detailed in Section \ref{sec:ColorFeatures}. Then, spatial features such as edge, texture and stereo features are studied in Section \ref{sec:EdgeFeatures}, Section \ref{sec:TextureFeatures} and Section \ref{sec:StereoFeatures}, respectively. Temporal features such as motion features are reviewed in Section \ref{sec:MotionFeatures}. In Section \ref{sec:TransformDomainFeatures}, features that are extracted in alternative domains other than the pixel domain are described. Finally, the different strategies of combining multiple features using fusion operators and feature selection mechanisms are discussed in Section \ref{sec:MultipleFeatures} and Section \ref{sec:FeatureSelection}. \\
\item \textbf{A description of the different resources available} to allow fair comparisons of the features. We present the color datasets and recent RGB-D datasets with accurate ground-truth providing a balanced coverage of the range of challenges which are present in the real world. Furthermore, we present the LBP Library which provide a common framework for the implementation of the local texture patterns. \\
\end{itemize}

\noindent The rest of this paper is organized as follows. First, a preliminary overview that investigates the classification of features by size and type of the features are considered within the context of background modeling and foreground detection in Section \ref{sec:PreliminaryOverview}. Moreover, basic concepts on feature reliability, feature fusion and feature selection are detailed in Section \ref{subsec:FReliability}, Section \ref{subsec:FFusion} and Section \ref{subsec:FSelection}, respectively. Further, each individual feature is reviewed with their crisp description in Section \ref{sec:ColorFeatures} to Section \ref{sec:MotionFeatures}. For each feature, the paper shall also present an investigation of their intrinsic properties that facilitate robustness against the challenges in real-life videos. Also, the paper shall provide an insight on strategies to combine multiple features using fusion operators (Section \ref{sec:MultipleFeatures}) and hence apply feature selection using boosting algorithms (Section \ref{sec:FeatureSelection}). Finally, a description of the different resources available such as datasets and codes is provided in Section \ref{sec:Resources}. Section \ref{sec:Conclusion} concludes with remarks on future research directions.\section{A Preliminary Overview}
\label{sec:PreliminaryOverview}
Features (descriptors or signatures) characterize a picture element captured in the current frame of a video sequence and are compared against a known background model to  classify it as either foreground or background. Feature representations can take multiple forms and can be computed for and from: a pixel, a block around the central pixel and a cluster (a region with the same value of feature than the current pixel). Practically, there are several types of features which can be computed either in the spatial or transform domains. Some of the features commonly used within the background modeling literature includes: color features, edge features, stereo features, motion features, texture features, local histogram features and Haar-like features \cite{TDF-3000}. These different features have intrinsic properties that allow the model to take into account spectral, spatial and/or temporal characteristics. Furthermore, these features use mathematical concepts in their design that facilitate computing them using well-known statistical or fuzzy concepts. Thus, features used in background/foreground separation can be classified from four different view points: their size, their type in a specific domain, their intrinsic properties and their mathematical concepts (Section \ref{subsec:CSize} to Section \ref{subsec:CMathematicalConcepts}). Then, we investigate how features can be used in terms of reliability, fusion and selection in Section \ref{subsec:FUsage}.

\subsection{Classification by Size} 
\label{subsec:CSize}
The size of the picture element chosen for interpreting necessary features that faithfully represent its characteristics plays a crucial role in modeling. As mentioned earlier, features can be computed from and for a pixel \cite{CF-1}, a block \cite{FSI-1} or a cluster \cite{FSI-10}. That is, the size of the picture element that is used to model the background and hence for comparing the current image frame to the background model, can either be a pixel \cite{CF-1}, a block \cite{FSI-1}, a region (Regions of difference \cite{FSI-200}, shape \cite{FSI-201}, behavior \cite{FSI-210}, cluster \cite{FSI-10}, superpixel \cite{FSI-100}) with a feature value. During practical implementations, a feature value at a given pixel can either depend on the feature value at the pixel itself or on the feature values around a predefined neighbourhood in the form of a block or a cluster. 

\begin{itemize}
\item \textbf{Pixel-based Features:} These features, otherwise known as point features, concern only the pixel at a given location $(x,y)$. This is the case of intensity and color features but in some cases include stereo features too. The background model applied in this case of pixel-based modeling and comparison is an independent process on each individual pixel. Practically, these features are used in uni-modal or multi-modal pixel-wise background modeling and foreground detection. Furthermore, pixel-based feature can be used to compute the mean or an other statistic value over spatial and/or temporal neighborhood to take into account spatial and/or temporal constraints. Then, the statistic value is assigned to the central pixel. For example, Varadarajan et al. \cite{FSI-220}\cite{FSI-221} proposed a region-based Mixture of Gaussians called (R-MOG) instead of a pixel based MOG. Each region is a square neighborhood which is effectively a block of size $4 \times 4$. Then, the color mean obtained from the neighborhood is assigned to the central pixel.\\
\item \textbf{Block-based Features:} This category of features is a generalization of the pixel-type, where in the element size a block of $1 \times 1$ or any aribrary block size $m \times n$ it represents an individual feature. In contrast to the previous case of pixel-based feature, which equally applies, spatial and/or temporal informations can also be computed depending on the spatial and temporal interaction of the element to its neighbourhood as in edge, texture and motion features. To completly exploit their potential, the spatial and/or temporal properties of these features need to be taken into account in all the background subtraction steps to be fully addressed. Practically, these block-based features can be assigned to a central pixel of a block (or neighborhood), or to all the block. For example, textures such as Local Binary Pattern can be assigned at each central pixel of a block size $3 \time 3$ by moving this block all over the frame, or to all the block as in the works ofHeikkila and Pietikainen \cite{TF-11}, and Heikkila et al. \cite{TF-10} which used a pixel-wise LBP histogram based one (LBP-P) and a block-wise LBP histogram based approach (LBP-B), respectively. Thus, the block-based features can be used in pixel-wise or block-wise background modeling and foreground detection. When the features are obtained from the video compressed domain, the approach is mandatory block-based because the block are pre-defined and thus they can not be moved over the frame. However, in block-based modeling and comparison, blocks (also called patches \cite{FSI-20}\cite{FSI-21}\cite{FSI-22}\cite{FSI-23}) can overlap or not \cite{FSI-2}. A block is usually obtained as a vector of $3 \times 3$ neighbors of the current pixel. The advantage is to take into account the spatial dimension to improve the robustness and to reduce the computation time. Furthermore, blocks can be of spatiotemporal type called spatiotemporal blocks \cite{TF-170}, spatiotemporal neighborhoods \cite{FSI-3}, spatiotemporal patches \cite{TF-15}\cite{TF-16}\cite{TF-17} or bricks \cite{TF-74}\cite{TF-75}\cite{FSI-15}) that intrinsicly encapsulate temporal information within spatial relationships of a group of pixels. In Pokrajac and Latecki \cite{TF-170}, a dimensionality reduction technique is applied to obtain a compact vector representation for each block. These blocks provide a joint representation of texture and motion patterns. One advantage is their robustness to noise and to the movement in the background. However, the disadvantage is that the detection is less precise because only blocks are detected, making them unsuitable for applications that require detailed shape information. \\
\item \textbf{Region-based Features:} Region-level (cluster-level, superpixel-level) features consider element sizes that are non-uniform across the image frame considered, and then specific features are computed on the correspondint element size. First, pixels in an image frame are grouped using an application-specific homogeniety criteria, typically exploiting partitioning mechanisms as follows: \textbf{1)} region-based mechanims as in Lin et al. \cite{FSI-200} with the notion of Regions of Difference (RoD), \textbf{2)} shape mechanims as proposed in Jacobs and Pless \cite{FSI-201}, \textbf{3)} behavior mechanims as in Jodoin et al. \cite{FSI-210}, \textbf{4)} clustering mechanisms as discussed by Bhaskar et al.  \cite{FSI-10}\cite{FSI-10-1}\cite{FSI-10-2}, and \textbf{5)} superpixel mechanims as in Sobral et al. \cite{FSI-100}. For example in Bhaskar et al. \cite{FSI-10}, each cluster contains pixels that have similar features in the color space. Then, the background model is applied on these clusters to obtain cluster of pixels classified as background or foreground. This cluster-wise approach gives less false alarms. Instead of the block-wise approach, the foreground detection is obtained at a pixel-level precision. \\
\end{itemize}
Pixel-based features need less time to be extracted than block-based or region-based features which require to be computed. In literature, in general, it can be summarized that the size of the feature and the comparison element determines the robustness of background modeling to noise and the challenges met in the videos, and often controls precision of foreground detection. A pixel-based modeling and comparison gives a pixel-based precision but it is less robust to noise compared to block-based or region-based based modeling and comparison. However, there are several works  which combined block-based (or region-based) and pixel-based approaches to reduce computation time by first using a block (or region) approach, and second to obtain a pixel precision by using a pixel-based approach, and they can be classified as follows: \textbf{(1)} multi-scales strategies \cite{SC-1}\cite{SC-2}\cite{SC-3}\cite{SC-4}\cite{SC-5}\cite{SC-6}, \textbf{(2)} multi-levels strategies \cite{500}\cite{EF-1}\cite{MF-100}\cite{MF-110}\cite{TF-191}\cite{MF-120}\cite{MF-200}\cite{MF-201}\cite{MF-202}\cite{TF-139-1}\cite{MF-300},  
\textbf{(3)} multi-resolutions strategies \cite{MR-1}\cite{MR-2}\cite{MR-3}\cite{MR-10},\textbf{(4)} multi-layers strategies \cite{ML-1}\cite{ML-2}
\cite{ML-10}\cite{ML-11}\cite{ML-12}\cite{ML-13}
\\ \cite{ML-14}\cite{ML-15}\cite{ML-100}\cite{ML-200}\cite{TF-179}\cite{StF-131}\cite{MulF-5}, \textbf{(5)}  hierarchical strategies \cite{CFS-1}\cite{CFS-2}\cite{CFS-3}\cite{CFS-10}
\\ \cite{CFS-11}\cite{CFS-12}\cite{CFS-13}\cite{CFS-14}\cite{CFS-15}\cite{TDF-3100}, and
\textbf{(6)} coarse-to-fine strategies \cite{CFS-100}\cite{CFS-101} \cite{CFS-110}\cite{CF-344}\cite{MF-17}.
The analysis of these different approaches is out of the scope of this review, and the reader can found details about these strategies in \cite{BookCRC2014}.

\subsection{Classification by Type}
\label{subsec:CType}
Features can be computed in the pixel domain or in a transform domain. In this section, features those are predominantly computed in each domain and their robustness to critical situations in real videos, are discussed.

\subsubsection{Features in the Pixel Domain}
\label{subsection:FPD}

Features are popularly computed in the pixel domain as the value of the pixel is directly available. The following features are commonly used: \\
\begin{itemize}
\item \textbf{Intensity features:} Intensity features are the most basic features that can be provide by gray-level cameras or infra-red (IR) cameras (See Section \ref{sec:IntensityFeatures}). \\
\item \textbf{Color features:} The color features in the RGB color space are most widely used because it is directly available from the sensor or the camera. But the RGB color space has an important drawback: its three components are dependent to each other which increases its sensitivity to illumination changes. For example, if a background point is covered by the shadow, the three component values at this point could be affected because the brightness and the chromaticity information are not separated. Thus, the three component values increase
or decrease together as the lighting increases or decreases, respectively \cite{FS-20}. Alternative color spaces that have also been explored in the literature include YUV or YCrCb spaces. Several comparisons between these color spaces are available in the literature including \cite{CF-110}\cite{CF-200}\cite{CF-201}\cite{CF-202}\cite{CF-203} and usually YCrCb is selected as the most appropriate color space. Although color features are often very discriminative features of objects, they have several limitations in the presence of challenges such as illumination changes, camouflage and shadows (See Section \ref{sec:ColorFeatures}). In order to solve such issues, authors have also proposed to use other features like edge, texture  and stereo features in addition to the color features. \\
\item \textbf{Edge features:} The ambient light present in the scene can significantly affect the appearance of moving objects. However, spectral features, are limited by their ability to adapt to such changes in appearance. Thus, edge features emerged as a robust alternative for moving object detection. Edge features are generally computed using a gradient approaches such as Canny, Sobel  \cite{EF-0}\cite{EF-60}\cite{FA-30}\cite{FS-20}\cite{900}
\\
\cite{OT-40}\cite{MulF-30} or Prewitt \cite{EF-2}\cite{EnF-5} edge detector. It is commonly believed that edge features can handle local illumination changes, thus eliminating the chances of leaving ghosts when foreground objects begin to move. Despite some compelling advantages, edge features (high pass filters) tend to vary more than other comparable features based on low pass filters \cite{FS-20}. For example, edge features in the horizontal and vertical directions have different reliability characteristics, since textured objects have high values in both directions, whereas homogeneous objects have low values in both directions (See Section \ref{sec:EdgeFeatures}). \\
\item \textbf{Texture features:} Texture features are appropriate to cope with illumination changes and shadows. Some common texture features that are generally used within this domain include the Local Binary Pattern (LBP) \cite{TF-10}, and the Local Ternary Pattern (LTP) \cite{TF-70}. Numerous variants of LBP and LTP can be found in the literature as can be seen summarized in Table \ref{Overview3}. Furthermore, statistical and fuzzy textures can be used as developed in Section \ref{sec:TextureFeatures}. \\
\item \textbf{Stereo features:} The extraction of stereo features rely on the need and use of specific acquisition systems such as a stereo, 3D, multiple, Time of Flight (ToF) cameras or  RGB-D cameras (Microsoft Kinect\protect\footnotemark[2], or Asus Xtion Pro Live\protect\footnotemark[3]) to obtain the disparity information that usually represent the depth in the visual scene. It has become well-known that stereo features allow the model to deal with the camouflage in color \cite{SF-10}\cite{SF-11}\cite{SF-100}\cite{SF-200}\cite{SF-300}\cite{SF-400}\cite{SF-500} (See Section \ref{sec:StereoFeatures}).\\
\item \textbf{Motion features:} Motion features are usually obtained via optical flow but with the limitation of the computational time. Motion features allow the model to deal with irrelevant background motion and clutter \cite{MF-1}\cite{MF-10}\cite{MF-11}\cite{MF-12}\cite{MF-13}\cite{MF-14}\cite{MF-15} (See Section \ref{sec:MotionFeatures}).\\
\item \textbf{Local histogram features :} Local histograms are usually computed on color features \cite{StF-1}\cite{StF-2}\cite{StF-3}\cite{StF-4}\cite{StF-5}\cite{StF-6}\cite{StF-7}\cite{StF-10}\cite{StF-11}\cite{StF-12}. But, local histograms can also be computed on edge features \cite{StF-100}\cite{StF-120}\cite{StF-130}\cite{StF-131}\cite{StF-132} to obtain Histograms of Oriented Gradients (HOG) (See Section \ref{sec:LHistogramF}).\\ 
\item \textbf{Local histon features:} Histon \cite{FF-1} is a contour plotted on the top of the histograms of three primary color components of a region in a manner that the collection of all points falling under the similar color sphere of predefined radius, called similarity threshold, belongs to one single value. The similar color sphere is the region in RGB color space such that all the colors falling in that region can be classified as one color. For every intensity value in the base histogram, the number of pixels falling under similar color sphere is calculated, and this value is added to the histogram value to get the histon value of that intensity. Histon can be extended to 3D histon and 3D Fuzzy histon as developed by Chiranjeevi and Sengupta \cite{FF-1} (See Section \ref{sec:LHistonF}).\\ 
\item \textbf{Local correlogram features:} Correlogram was originally proposed for computer vision applications like object tracking \cite{611-10}. Since, correlogram captures the inter-pixel relation of two pixels at a given distance, spatial information is obtained in addition to the color information. Thus, correlograms can efficiently alleviate the drawbacks of histograms, which only consider the pixel intensities for calculating the distribution. The main drawback of correlograms is their computation time due to their size of $256^3 \times256^3$ in RGB, and $256\times2563$ in grey level. Hence, the single channel is quantized to a finite number of levels $l$. Due to this, the correlograms' size is further reduced to $l \times l$ with $l \ll 256$. Correlogram can be extended to fuzzy correlogram \cite{FF-5} and multi-channel fuzzy correlogram \cite{FF-6} (See Section \ref{sec:LCorrelogramF}). \\
\item \textbf{Haar-like features:} Some authors \cite{FA-1}\cite{FA-2}\cite{FR-20}\cite{TDF-3100}, used the Haar-like features \cite{TDF-3000}.  
Haar-like features are features defined in real-time face detector and based on the similarity with Haar wavelets. Haar-like features are computed from adjacent rectangular areas at a given location in a detection window by adding the pixel intensities in each area and by calculating the difference between these sums. The main advantage of Haar-like features is their computation speed. With the use of integral images, Haar-like features of any size can be computed in constant time (See Section  Section \ref{sec:HaarFeatures}). \\
\item \textbf{Location features:} The location (x,y) can be used as a feature to exploit the dependency between the pixel \cite{LF-1}\cite{LF-2}\cite{LF-10}\cite{LF-11}\cite{LF-12} (See Section \ref{sec:LocationFeatures}). \\ 
\end{itemize}
Table \ref{Overview1}, Table \ref{Overview2}, Table \ref{Overview3}, Table \ref{Overview4} and Table \ref{Overview5} present an overview of the features in the pixel domain. Pixel domain features are generally robust and perform well provided more accurate representation of the visual scene is available. However, its high computational complexity restricts its real-time use in some applications. The features in the pixel domain are analyzed in details from Section \ref{sec:ColorFeatures} to Section \ref{sec:LocationFeatures}.

\footnotetext[2]{{http://www.microsoft.com/en-us/kinectforwindows/}}
\footnotetext[3]{{http://www.asus.com/Multimedia/}}

\subsubsection{Features in a Transform Domain}
In order to accomplish some of the real-time demands of visual scene analysis,  feature computation in a transform domain has gained importance. 
\begin{itemize}
\item \textbf{Frequency domain:} The frequency domain offers a good framework to detect periodic processes that appear in dynamics backgrounds such as waving trees and waves in the ocean. For this, there is a need to transform the data values in the pixel domain into the frequency domain via a transformation such as the Fourier Transform (FFT) \cite{TDF-1}\cite{TDF-1-1}, Discrete Cosine Transform (DCT) \cite{TDF-2}, Wavelet Transform \cite{WTDF-1}\cite{WTDF-2}\cite{WTDF-3}, Curvelet Transform \cite{TDF-20}, Walsh Transform \cite{TDF-30}\cite{TDF-31}\cite{TDF-32}, Hadamard Transform \cite{TDF-50}, Slant Transform \cite{TDF-200} and Gabor Transform \cite{TDF-100}\cite{TDF-101}\cite{TDF-110}. Pratically, FTT  processes blocks much faster in comparison with DCT. But, DCT outperforms slightly FFT in terms of precision, similarity and F-measure \cite{CF-344}.  \\ 
\item \textbf{Video compressed domain:} As videos are usually compressed before transmission and storage, a number of compressed domain approaches have also been developed to improve the computational complexity of feature extraction. To obtain moving objects, the compressed video stream is partially decoded. Thus, the compressed domain data, such as motion vectors (MVs), transform coefficients, are employed to extract moving objects. Initially, compressed domain algorithms focussed attention on the MPEG standard. According to the classification of moving object detection, those methods can be mainly divided into three groups: MVs based \cite{TDF-2000}, coefficients based  \cite{TDF-2010} \cite{TDF-2011}, and combining MV and coefficients based \cite{TDF-2020} \cite{TDF-2021}. Other compressed domain algorithms used the video coding standard such as the H.264/AVC which process each video frame in units of a MacroBlock (MB) \cite{TDF-2050}. Thus, these MacroBlocks could be used as features. A Rate Distortion Cost (RDCost) value for each MB which changes depending on the frame content can be used as an indicator of changes. Typically, more cost will be spent on high motion and/or detailed MBs and less cost for low motion and/or homogenous MBs, which was verified in the experiments of \cite{TDF-2050}. The reasons for the effectiveness of the RDCost for foreground/background separation can be attributed to the following reasons. First, RDCost reflects the overall coding cost of a MB, which considers the effect of each factor on coding efficiency during video coding, such as prediction mode, MB partition size, motion vectors, residuals, etc. So RDCost can reflect true motion. Second, RDCost is in unit of MB. Only one MB has an RDCost value, as the basic coding unit in H.264/AVC is MB. Thus, RDCost is less affected by noise when compared with MVs \cite{TDF-2000}. Finally, compressed domain algorithms recently focussed attention on the HEVC standard \cite{TDF-2100}\cite{TDF-2110}. \\
\end{itemize}
Table \ref{Overview3} shows an overview of the features in a transform domain. The features in the transform domain are analyzed in details in Section \ref{sec:TransformDomainFeatures}.\\

\subsection{Classification by Intrinsic Properties}
\label{subsec:CIntrinsicProperties}
According to Li et al. \cite{900}, features can be classified by their intrinsic properties into the following categories:
\begin{itemize}
\item  \textbf{Spectral features:}  The intensity or color features are directly available from the images. Spectral features can easily detect changes if the difference in color between the foreground and the background are sufficient. However, spectral features produce \textbf{\textit{1)}} false positive detections particularly when there are illumination changes, and \textbf{\textit{2)}} false negative detections when foreground objects have similar color to the background (camouflage).  Spectral features do not take into account or exploit the neighbourhood relationship of the considered pixel to deal with its poor robustness. Further, stereo features can also be considered as spectral features in the depth domain.\\
\item  \textbf{Spatial features:} Spatial features are edge and texture features. These features help to detect foreground objects that camouflage with the background and suppress shadows. Spatial features are however not applicable to non-stationary background objects at pixel level since the corresponding spatial features vary over time. \\
\item  \textbf{Temporal features:} Temporal features concern the motion between consecutive image frames. One way to obtain temporal features is to estimate the consistency of optical flow over a short duration of time. \\
\end{itemize}

In order to achieve robust background modeling and foreground detection, features in each category are required to deal with a number of challenges commonly encountered in video surveillance as indicated in Bouwmans \cite{JournalCOSREV2014-1}. Table \ref{IPClassificationFeatures} shows an overview of the features classified following their intrinsic properties.\\

\subsection{Classification by Mathematical Concepts}
\label{subsec:CMathematicalConcepts}
Some of the mathematical concepts that underlie during the computation of robust features can present an other useful categorization of features into crisp, statistical and fuzzy types.

\begin{itemize}
\item  \textbf{Crisp Features:} Crisp features are those features which are computed without the use and need of any statistical or fuzzy concepts. It is the case of the color features (RGB, YUV, HSV, etc...), edge features obtained by a filter (Canny, Sobel, Prewitt), motion features obtained through optical flow or temporal operator, and stereo features. \\
\item \textbf{Statistical Features:} Statistical features can be obtained by exploiting some of the statistical properties of the representation of the visual. The first work developed by Satoh et al. \cite{TF-130} proposed a Radial Reach Correlation (RRC)\index{Radial Reach Correlation} feature which has several variants: Bi-polar Radial Reach Correlation (BP-RCC) \cite{TF-132}, Fast Radial Reach Correlation (F-RRC) \cite{TF-135}\cite{TF-135-1}, and Probabilistic Bi-polar Radial Reach Correlation (PrBP-RCC) \cite{TF-136}. In a similar way, Yokoi \cite{TF-5} used Peripheral Ternary Sign Correlation (PTESC). Recently, Yoshinaga et al. \cite{TF-101}\cite{TF-102} proposed the Statistical Local Difference Pattern (SLDP). The aim of these statistical features is to be more robust to illumination changes and dynamic backgrounds. Thus,  SLDP integrates both pixel-based multi-modal model with color feature and spatial-based unimodal model with texture feature in one model feature taking into account the advantages of their respective robustness. \\
\item  \textbf{Fuzzy Features:} Fuzzy features are used to take into account the imprecision and the incertitude in features that represent a visual scene. For example, Chiranjeevi and Sengupta introduced fuzzy 3D Histons \cite{FF-1}, fuzzy correlograms \cite{FF-5}\cite{FF-6} and fuzzy statistical texture features \cite{FF-10}\cite{FA-40}. The aim is to deal with illumination changes and dynamic backgrounds. \\
\end{itemize}
Table \ref{ClassificationFeatures} shows an overview of the features classified following their mathematical concepts.

\subsection{Exhaustive Overview of all the Features}
\label{subsec:ExhaustiveOverviewFeatures}
The reader can refer to Table \ref{Overview1}, Table \ref{Overview2}, Table \ref{Overview3}, Table \ref{Overview4} and Table \ref{Overview5} for an exhaustive overview of all the features. The first column indicates the category model and the second column the name of each method. Their corresponding acronym is indicated in the first parenthesis and the number of papers counted for each method in the second parenthesis. The third column gives the name of the authors and the date of the related publication. Furthermore, prospective features not currently used for background modeling and foreground detection but in other computer vision applications are indicated in each table. \\

\begin{table}
\scalebox{0.75}{
\begin{tabular}{|l|l|l|l|l|} 
\hline
\scriptsize{Spectral Features} &\scriptsize{Spatial Features} &\scriptsize{Temporal Features} 
&\scriptsize{Spectral/Spatial Features} &\scriptsize{Spatial/Temporal Features} \\
\hline
\hline
\scriptsize{\textbf{Color features}}       &\scriptsize{\textbf{Edge Features}}                      &\scriptsize{\textbf{Motion Features}}   &\scriptsize{\textbf{Local Histogram}}     &\scriptsize{\textbf{Texture Features}}                   \\
\scriptsize{RGB, Normalized RGB}           &\scriptsize{Canny, Sobel, Prewitt}                       &\scriptsize{Optical Flow}              
&\scriptsize{LCH,LK-CH}                    &\scriptsize{STLBP, \textbf{\textit{SCS-LBP}}, ST-CS-LBP} \\
\scriptsize{YUV, HSV}                      &\scriptsize{\textbf{Texture Features}}                   &\scriptsize{Temporal Operator}         
&\scriptsize{HOG, LOH, LK-HOG}             &\scriptsize{MV-LBP, SLBP-AM}                             \\
\scriptsize{HSI, Luv}                      &\scriptsize{PISC, PTESC}                                 &\scriptsize{}                          
&\scriptsize{LDH, LHFG}                    &\scriptsize{CS-ST-LTP, ST-SITLP}                         \\
\scriptsize{Improved HLS, Ohta}            &\scriptsize{LBP, ELBP}                                   &\scriptsize{}                
&\scriptsize{\textbf{Texture Features}}    &\scriptsize{CS-SILTP}                                    \\
\scriptsize{YCrCB, Lab, Lab2000HL}         &\scriptsize{CS-LBP,HCS-LBP}                              &\scriptsize{}              
&\scriptsize{SC-LBP, \textbf{\textit{SCS-LBP}}, OC-LBP}     &\scriptsize{}                           \\
\scriptsize{}                              									&\scriptsize{DLBP, ULBP}                 &\scriptsize{}              
&\scriptsize{iLBP}                                          &\scriptsize{}                           \\
\scriptsize{}                                               &\scriptsize{ExtLBP, RI-LBP}             &\scriptsize{}              
&\scriptsize{MC-SITLP}                                      &\scriptsize{}                           \\
\scriptsize{}                                               &\scriptsize{LN-LBP, SALBP}              &\scriptsize{}              
&\scriptsize{\textbf{Frequency Features}}                   &\scriptsize{}                           \\
\scriptsize{}                                      &\scriptsize{WB-LBP}              &\scriptsize{}              
&\scriptsize{FFT, DCT, WT}                         &\scriptsize{}                    \\
\scriptsize{}                                      &\scriptsize{LTP, SILTP}          &\scriptsize{}              
&\scriptsize{HT, GT}                               &\scriptsize{}                    \\
\scriptsize{}                                      &\scriptsize{SCS-LTP}             &\scriptsize{}              
&\scriptsize{\textbf{Video Compressed Features}}   &\scriptsize{}                    \\
\scriptsize{}                                      &\scriptsize{SILS}                &\scriptsize{}              
&\scriptsize{MPEG (MVs, Coefficients, Mixed)}      &\scriptsize{}                    \\
\scriptsize{}                                      &\scriptsize{\textbf{Stereo features}} &\scriptsize{}              
&\scriptsize{H.264/AVC (MBs)}                      &\scriptsize{}                         \\
\scriptsize{}                                      &\scriptsize{Disparity, Depth}          &\scriptsize{}              
&\scriptsize{HEVC (MBs)}                           &\scriptsize{}                          \\
\hline
\end{tabular}}
\caption{Classification by Intrinsic Properties: An Overview.} \centering
\label{IPClassificationFeatures}
\end{table}

\begin{table}
\scalebox{1.0}{
\begin{tabular}{|l|l|l|} 
\hline
\scriptsize{Crisp Features} &\scriptsize{Statistical Features} &\scriptsize{Fuzzy Features} \\
\hline
\hline
\scriptsize{\textbf{Color features}}     &\scriptsize{\textbf{Intensity features}}                  
&\scriptsize{\textbf{Color Coherence Vector}} \\
\scriptsize{RGB, Normalized RGB}         &\scriptsize{Rank Order Statistic \cite{CDFeatureI-1}}  
&\scriptsize{FCC \cite{StF-300}\cite{StF-300}} \\  
\scriptsize{YUV, HSV, HSI}               &\scriptsize{OSID \cite{FeatureI-1}}                                  
&\scriptsize{\textbf{Local Color Histogram}}   \\
\scriptsize{Luv, Improved HLS}           &\scriptsize{\textbf{Color features}}                       
&\scriptsize{LFCH \cite{StF-200}\cite{StF-210}\cite{StF-220}\cite{StF-230}\cite{StF-240}}    \\
\scriptsize{Ohta, YCrCb, Lab, Lab2000HL}                               &\scriptsize{Standard Variance Feature \cite{StF-50}} 
&\scriptsize{LFCDH \cite{StF-250}}  \\
\scriptsize{\textbf{Edge features}}                                    &\scriptsize{Color co-occurences \cite{900}}                    
&\scriptsize{\textbf{Local Gradient Histogram}}        \\
\scriptsize{Canny \cite{EF-1}, Sobel \cite{EF-0}, Prewitt \cite{EF-2}} &\scriptsize{Entropy \cite{StF-30}\cite{StF-31}\cite{StF-32}} 
&\scriptsize{HOFG \cite{StF-135}} \\
\scriptsize{\textbf{Motion Features}}                                  &\scriptsize{IISC \cite{OT-50}} 
&\scriptsize{\textbf{Local Histon}}  \\
\scriptsize{Optical Flow, Temporal Operator \cite{TF-19}}       &\scriptsize{NCC \cite{IF-100}}      
&\scriptsize{3D Fuzzy Histon \cite{FF-1}}    \\
\scriptsize{\textbf{Stereo Features}}                           &\scriptsize{\textbf{Edge features}}                             
&\scriptsize{\textbf{Local Correlogram}}    \\
\scriptsize{Disparity \cite{SF-1}, Depth \cite{SF-100}}         &\scriptsize{Gradient Deviations \cite{MulF-7}}                                 
&\scriptsize{Fuzzy Correlogram \cite{FF-5}}        \\
\scriptsize{LCH \cite{StF-2}, LK-CH \cite{StF-3}}               &\scriptsize{Projection Gradient Statistics \cite{StF-60}}             
&\scriptsize{Multi-channel Kernel FC \cite{FF-6}}  \\
\scriptsize{LCH \cite{StF-2}, LK-CH \cite{StF-3}}               &\scriptsize{\textbf{Texture features}}                            &\scriptsize{\textbf{Local Fuzzy Pattern} \cite{TF-121}}    \\
\scriptsize{LCH \cite{StF-2}, LK-CH \cite{StF-3}}               &\scriptsize{PISC \cite{TF-1}, PETSC \cite{TF-5}}                          
&\scriptsize{\textbf{Fuzzy Statistical Texture}}   \\
\scriptsize{HOG \cite{StF-100}, LOH \cite{StF-6}, LK-HOG \cite{StF-4}} &\scriptsize{RRC \cite{TF-130}, BP-RCC \cite{TF-132}}                        
&\scriptsize{FST \cite{FF-10}\cite{FA-40}} \\
\scriptsize{LDH \cite{StF-10}, LHFG \cite{StF-12}}                       &\scriptsize{F-RRC \cite{TF-135}, PrBP-RCC \cite{TF-136}}                   &\scriptsize{}    \\
\scriptsize{\textbf{Local Histon}}                                       &\scriptsize{ABP-RCC \cite{TF-140}}                   
&\scriptsize{}    \\
\scriptsize{Histon \cite{FF-1}, 3D Histon \cite{FF-1}}                   &\scriptsize{RRF \cite{TF-150}}                   
&\scriptsize{}    \\
\scriptsize{\textbf{Local Correlogram}}                                  &\scriptsize{RPF \cite{TF-160}, MRPF \cite{TF-161}}                        
&\scriptsize{}    \\
\scriptsize{Correlogram \cite{611-10}}                                   &\scriptsize{SRF \cite{StF-40}}      
&\scriptsize{}    \\
\scriptsize{\textbf{Location} \cite{LF-1}}                               &\scriptsize{ST \cite{FA-33}}                        
&\scriptsize{}    \\
\scriptsize{\textbf{Haar like Features} \cite{FA-1}}                     &\scriptsize{}                       
&\scriptsize{}    \\
\hline
\end{tabular}}
\caption{Classification by Mathematical Concepts: An Overview.} \centering
\label{ClassificationFeatures}
\end{table}

\newpage
\begin{table}
\scalebox{0.90}{
\begin{tabular}{|l|l|l|} 
\hline
\scriptsize{Pixel Domain Features} &\scriptsize{Categories} &\scriptsize{Authors - Dates} \\
\hline
\hline
\scriptsize{Intensity features}   &\scriptsize{\textbf{1) Well-Known Intensity Features}}   &\scriptsize{}                 \\
\scriptsize{} &\scriptsize{\textbf{Visible cameras}}                                        &\scriptsize{}                 \\
\scriptsize{} &\scriptsize{Intensity (4)}                 &\scriptsize{Silveira et al. (2005) \cite{IF-1}}                 \\
\scriptsize{} &\scriptsize{\textbf{IR cameras}}                                             &\scriptsize{}                 \\
\scriptsize{} &\scriptsize{Intensity (5)}                 &\scriptsize{Davis and Sharma (2004) \cite{IF-50}}               \\
\cline{2-3}
\scriptsize{}                     &\scriptsize{\textbf{2) Designed Illumination Invariant Features}} &\scriptsize{}                    \\
\scriptsize{} &\scriptsize{Illumination Ratio (1)}                   &\scriptsize{Paruchuri et al. (2011) \cite{IF-140}}               \\
\scriptsize{} &\scriptsize{Brightness (1)}                           &\scriptsize{Wang et al. (2015) \cite{IF-500}}                    \\
\scriptsize{} &\scriptsize{Reflectance (7)}                          &\scriptsize{Toth et al. (2000) \cite{CDFeatureI-303}}            \\
\scriptsize{} &\scriptsize{Surface Spectral Reflectance (SSR) (1)}   &\scriptsize{Sedky et al. (2014) \cite{CDFeatureI-500}}           \\
\scriptsize{} &\scriptsize{Radiance (1)} 														 &\scriptsize{Xie et al. (2004) \cite{CDFeatureI-1}}               \\
\scriptsize{} &\scriptsize{Photometric Variations (1)}               &\scriptsize{Di Stefano et al. (2007) \cite{CDFeatureI-2}}        \\
\cline{2-3}
\scriptsize{}                     ,&\scriptsize{\textbf{3) Statistical Intensity Features}}  &\scriptsize{}                    \\
\scriptsize{} &\scriptsize{Rank Order Statistics (1)}                     &\scriptsize{Xie et al. (2004) \cite{CDFeatureI-1}}  \\
\cline{2-3}
\scriptsize{} &\scriptsize{\textbf{4) Prospective Intensity Features}}    &\scriptsize{}                                       \\
\scriptsize{} &\scriptsize{Ordinal Spatial Intensity Distribution (OSID)} &\scriptsize{Tang et al. (2014) \cite{FeatureI-1}}   \\
\hline
\scriptsize{Color features}   &\scriptsize{\textbf{1) Well-Known Color Spaces}} &\scriptsize{}                           \\
\scriptsize{} &\scriptsize{RGB}                               &\scriptsize{Stauffer and Grimson (1999) \cite{CF-1}}      \\
\scriptsize{} &\scriptsize{Normalized RGB}                    &\scriptsize{Xu et Ellis (2001) \cite{CF-10}}              \\
\scriptsize{} &\scriptsize{YUV}                               &\scriptsize{Harville et al. (2001) \cite{SF-100}}         \\
\scriptsize{} &\scriptsize{HSV}                               &\scriptsize{Sun et al. (2006) \cite{CF-60}}               \\
\scriptsize{} &\scriptsize{HSI}                               &\scriptsize{Wang and Wu (2006) \cite{CF-70}}              \\
\scriptsize{} &\scriptsize{Luv}                               &\scriptsize{Yang and Hsu (2006) \cite{CF-90}}             \\
\scriptsize{} &\scriptsize{Improved HLS}                      &\scriptsize{Setiawan et al. (2006) \cite{CF-100}}         \\
\scriptsize{} &\scriptsize{Ohta}                              &\scriptsize{Zhang and Xu (2006) \cite{FA-10}}             \\
\scriptsize{} &\scriptsize{YCrCb}                             &\scriptsize{Baf et al. (2008) \cite{FA-11}}               \\
\scriptsize{} &\scriptsize{YIQ}                               &\scriptsize{Thangarajah et al. (2015) \cite{CF-130}}      \\
\scriptsize{} &\scriptsize{Lab}                               &\scriptsize{Balcilar et al. (2013) \cite{CF-120}}         \\
\scriptsize{} &\scriptsize{Lab2000HL}                         &\scriptsize{Balcilar et al. (2013) \cite{CF-120}}         \\
\cline{2-3}
\scriptsize{} &\scriptsize{\textbf{2) Designed Shape Color Spaces}}      &\scriptsize{}                                       \\
\scriptsize{} &\scriptsize{Cylinder Color (CY) (5)}                   &\scriptsize{Horprasert  et al. (1999) \cite{CF-300}}   \\
\scriptsize{} &\scriptsize{Hybrid Cone-Cylinder Color (2)}               &\scriptsize{Doshi and Trivedi (2006) \cite{CF-315}} \\
\scriptsize{} &\scriptsize{Arbitrary Cylinder Color (ACY) (2)}     &\scriptsize{Zeng and Jia (2014) \cite{CF-320}}       \\
\scriptsize{} &\scriptsize{Ellipsoidal Color (EC)(1)}							 &\scriptsize{Sun et al. (2011) \cite{CF-330}}         \\
\scriptsize{} &\scriptsize{Box-based Color (1)}                    &\scriptsize{Tu et al. (2008) \cite{CF-340}}          \\
\scriptsize{} &\scriptsize{Cubic Color (1)}                        &\scriptsize{Noh and Jeon (2011) \cite{CF-342}}       \\
\scriptsize{} &\scriptsize{Spherical Color (1)}                    &\scriptsize{Hu et al. (2012) \cite{CF-343}}          \\
\scriptsize{} &\scriptsize{Double-Trapezium Cylinder Color (DTC)}  &\scriptsize{Huang et al. (2015) \cite{CF-350}}       \\
\scriptsize{} &\scriptsize{Conical Color}                          &\scriptsize{-}                                       \\
\cline{2-3}
\scriptsize{} &\scriptsize{\textbf{3) Designed Illumination Invariant Color Features}} &\scriptsize{}                                     \\
\scriptsize{} &\scriptsize{Color Illumination Ratio (2)}                               &\scriptsize{Pilet et al. (2008) \cite{IF-100}}    \\
\scriptsize{} &\scriptsize{RGB-HSV (1)}                      									         &\scriptsize{Takahara et al. (2012) \cite{IF-120}} \\
\scriptsize{} &\scriptsize{RGB-YCrCb (1)}                      									       &\scriptsize{Sajid and Cheung (2014) \cite{IF-120-1}} \\
\scriptsize{} &\scriptsize{Color Illumination Invariant (1)}                           &\scriptsize{Yeh et al. (2012) \cite{CF-500}}         \\
\cline{2-3}
\scriptsize{} &\scriptsize{\textbf{4) Color Filter Array (CFA) Features}}              &\scriptsize{}                                  \\
\scriptsize{} &\scriptsize{CFA patterns (1)}                                           &\scriptsize{Suhr et al. (2011) \cite{CF-2}}    \\
\scriptsize{} &\scriptsize{Bayer CFA patterns (1)}                                     &\scriptsize{Suhr et al. (2011) \cite{CF-2}}    \\
\cline{2-3}
\scriptsize{} &\scriptsize{\textbf{5) Statistics on Color Features}}     &\scriptsize{}                                   \\
\scriptsize{} &\scriptsize{Standard Variance Feature (1)}                &\scriptsize{Zhong et al. (2010) \cite{StF-50}}  \\
\scriptsize{} &\scriptsize{Color co-occurrences (2)}                     &\scriptsize{Li et al. (2004) \cite{900}}        \\
\scriptsize{} &\scriptsize{Entropy (3)}                                  &\scriptsize{Ma and Zhang (2001) \cite{StF-30}}  \\
\scriptsize{} &\scriptsize{IISC (1)}                                     &\scriptsize{Kim and Kim (2016) \cite{OT-50}}    \\
\scriptsize{} &\scriptsize{Normalized Cross Correlation (NCC) (2)}       &\scriptsize{Pilet et al. (2008) \cite{IF-100}}  \\
\cline{2-3}
\scriptsize{} &\scriptsize{\textbf{6) Multiscale Color Features}}       &\scriptsize{}                                    \\
\scriptsize{} &\scriptsize{Multiscale Color Description (1)}            &\scriptsize{Muchtar et al. (2011) \cite{CF-510}} \\
\scriptsize{} &\scriptsize{\textbf{7) Fuzzy Color Features}}            &\scriptsize{}                                    \\
\scriptsize{} &\scriptsize{Fuzzy Color Coherence Vector (1)}            &\scriptsize{Qiao et al. (2014) \cite{StF-300}}   \\
\scriptsize{} &\scriptsize{\textbf{7) Prospective Color Features}}      &\scriptsize{}                                    \\
\scriptsize{} &\scriptsize{Scale-Invariant Feature Transform (SIFT)}    &\scriptsize{-}                                   \\
\scriptsize{} &\scriptsize{Speeded Up Robust Features (SURF) (1)}       &\scriptsize{Shah et al. (2014) \cite{OT-60}}    \\
\scriptsize{} &\scriptsize{Weber Local Descriptor (WLD)}                &\scriptsize{Chen et al. (2009) \cite{OT-70}}     \\
\hline
\scriptsize{Multispectral features}   &\scriptsize{Multi-spectral features (3)}           &\scriptsize{Benezeth et al. (2014) \cite{906}} \\
\hline
\scriptsize{Edge Features}    &\scriptsize{\textbf{1) Crisp Edge}}                         &\scriptsize{}                                    \\
\scriptsize{}    &\scriptsize{Gradient (Magnitude/Direction) (1)}                          &\scriptsize{Javed et al. (2002) \cite{EF-1}}     \\
\scriptsize{}    &\scriptsize{Sobel Edge Detector (7)}                                     &\scriptsize{Jabri et al. (2000) \cite{EF-0}}     \\
\scriptsize{}    &\scriptsize{Prewitt Edge Detector (2)}                                   &\scriptsize{Lindstrom et al. (2006) \cite{EF-2}} \\
\scriptsize{}    &\scriptsize{Local Hybrid Pattern  (LHP) (1)}                             &\scriptsize{Kim et al. (2015) \cite{OT-40}}      \\
\cline{2-3}
\scriptsize{}    &\scriptsize{\textbf{2) Statistical Edge}}                                &\scriptsize{}                                    \\
\scriptsize{}    &\scriptsize{Gradient Deviations (1)}                                     &\scriptsize{Kamkar-Parsi (2005) \cite{MulF-7}}  \\
\scriptsize{}    &\scriptsize{Projection Gradient Statistics (1)}                          &\scriptsize{Zhang (2012) \cite{StF-60}}          \\
\hline
\end{tabular}}
\caption{Features in the Pixel Domain: An Overview (Part 1).} \centering
\label{Overview1}
\end{table}

\begin{table}
\scalebox{0.79}{
\begin{tabular}{|l|l|l|} 
\hline
\scriptsize{Pixel Domain Features} &\scriptsize{Categories} &\scriptsize{Authors - Dates} \\
\hline
\hline
\scriptsize{Motion Features} &\scriptsize{Optical Flow (9)}                        &\scriptsize{Tang et al. (2007) \cite{MF-1}}          \\
\scriptsize{} &\scriptsize{Temporal Operator (4)}                                  &\scriptsize{Zhong et al. (2008) \cite{TF-19}}        \\
\scriptsize{} &\scriptsize{SIFT flow (1)}                                          &\scriptsize{Dou and Li (2014) \cite{MF-19}}          \\
\scriptsize{} &\scriptsize{Magno channel (1)}                                      &\scriptsize{Martins et al. (2016) \cite{OT-10}}      \\
\hline
\scriptsize{Stereo Features} &\scriptsize{\textbf{1) Disparity}}                &\scriptsize{}                                         \\
\scriptsize{}                &\scriptsize{Disparity (5)}                        &\scriptsize{Ivanov  et al. (1997) \cite{SF-1}}        \\
\scriptsize{}                &\scriptsize{Variational Disparity (1)}            &\scriptsize{Javed  et al. (2015) \cite{SF-770}}       \\
\cline{2-3}
\scriptsize{}                &\scriptsize{\textbf{2) Depth}}                    &\scriptsize{}                                         \\
\scriptsize{}                &\scriptsize{Depth (Stereo System) (4)}            &\scriptsize{Harville et al. (2001) \cite{SF-100}}      \\
\scriptsize{}                &\scriptsize{Depth (ToF) (6)}                      &\scriptsize{Silvestre (2007) \cite{SF-200}}            \\
\scriptsize{}                &\scriptsize{Depth (Microsoft Kinect) (13)}        &\scriptsize{Camplani and Salgado (2013) \cite{SF-300}} \\
\hline
\scriptsize{Local Histograms Features}   &\scriptsize{\textbf{1) Local Histograms of Color}}         &\scriptsize{}      \\
\scriptsize{}                            &\scriptsize{\textbf{1.1) Crisp Local Histograms of Color}} &\scriptsize{}      \\
\scriptsize{}                            &\scriptsize{Local Color Histogram (LCH) (1)}  &\scriptsize{Mason and Duric (2001) \cite{StF-2}}      \\
\scriptsize{} &\scriptsize{Local Kernel Color Histograms (LK-CH) (1)}                   &\scriptsize{Noriega  et al. (2006) \cite{StF-3}}      \\
\scriptsize{} &\scriptsize{Estimated Local Kernel Histogram (ELKH) (1)}                 &\scriptsize{Li et al. (2008) \cite{StF-5}}            \\
\scriptsize{} &\scriptsize{Local Color Difference Histograms (LDCH) (1)}                &\scriptsize{Li (2009) \cite{StF-1}}                   \\
\scriptsize{} &\scriptsize{Local Dependency Histograms (LDH) (2)}                       &\scriptsize{Zhang et al. (2008) \cite{StF-10}}        \\
\scriptsize{} &\scriptsize{Spatiotemporal Condition Information (SCI)(1)}               &\scriptsize{Wang et al. (2014) \cite{FSI-3}}          \\
\scriptsize{} &\scriptsize{\textbf{1.2) Fuzzy Local Histogram of Color}}                &\scriptsize{}                                         \\
\scriptsize{} &\scriptsize{Local Fuzzy Color Histograms (LFCH) (6)}  			              &\scriptsize{Kim and Kim (2012) \cite{StF-200}}        \\
\scriptsize{} &\scriptsize{Local Fuzzy Color Difference Histograms (LFCDH) (1)}         &\scriptsize{Panda et al. \cite{StF-250}}              \\
\cline{2-3}
\scriptsize{} &\scriptsize{\textbf{2) Local Histograms of Gradient}}                    &\scriptsize{}                                        \\
\scriptsize{} &\scriptsize{\textbf{2.1) Crisp Local Histograms of Gradient}}            &\scriptsize{}                                        \\
\scriptsize{} &\scriptsize{Local Histogram on Gradient (LGH) (1)}                       &\scriptsize{Mason and Duric (2001) \cite{StF-2}}      \\
\scriptsize{} &\scriptsize{Local Histogram of Oriented Gradient (L-HOG) (3)}            &\scriptsize{Fabian  (2010) \cite{StF-120}}            \\
\scriptsize{} &\scriptsize{Local Adaptive HOG (LA-HOG) (1)}           									&\scriptsize{Hu et al. (2010) \cite{StF-130}}          \\
\scriptsize{} &\scriptsize{Local Orientation Histograms (LOH)(1)}                       &\scriptsize{Jang et al. (2008) \cite{StF-6}}          \\
\scriptsize{} &\scriptsize{Local Kernel Histograms of Oriented Gradients (LK-HOG) (1)}  &\scriptsize{Noriega and Bernier (2006) \cite{StF-4}}  \\
\scriptsize{} &\scriptsize{\textbf{Prospective Histograms}}                             &\scriptsize{}                                         \\
\scriptsize{} &\scriptsize{\textbf{2.2) Fuzzy Local Histograms of Gradient}}            &\scriptsize{}                                         \\
\scriptsize{} &\scriptsize{Local Histograms of Fuzzy Gradient (HOFG) (1)}               &\scriptsize{Salhi et al. (2013] \cite{StF-135}}       \\
\cline{2-3}
\scriptsize{} &\scriptsize{\textbf{3) Local Histograms of Figure/Ground}}               &\scriptsize{}                                         \\
\scriptsize{} &\scriptsize{Local Histogram of Figure/Ground (LHFG) (1)}                 &\scriptsize{Zhong et al. (2009) \cite{StF-12}}        \\
\hline
\scriptsize{Local Histon Features}      &\scriptsize{\textbf{1) Crisp Local Histon}}       &\scriptsize{}      \\
\scriptsize{} &\scriptsize{Histon (1)}                                          &\scriptsize{Chiranjeevi and Sengupta (2012) \cite{FF-1}}      \\
\scriptsize{} &\scriptsize{3D Histon (1)}                                       &\scriptsize{Chiranjeevi and Sengupta (2012) \cite{FF-1}}      \\
\cline{2-3}
\scriptsize{} &\scriptsize{\textbf{2) Fuzzy Local Histon}}                      &\scriptsize{}       \\
\scriptsize{} &\scriptsize{3D Fuzzy Histon (1)}                                 &\scriptsize{Chiranjeevi and Sengupta (2012) \cite{FF-1}}      \\
\hline
\scriptsize{Local Correlogram Features}   &\scriptsize{\textbf{1) Crisp Local Correlogram}} &\scriptsize{}        \\
\scriptsize{}                             &\scriptsize{Correlogram}                         &\scriptsize{Zhao and Tao (2005) \cite{611-10}}       \\
\cline{2-3}
\scriptsize{}                             &\scriptsize{\textbf{2) Fuzzy Local Correlogram}} &\scriptsize{}        \\
\scriptsize{}   &\scriptsize{Fuzzy Correlogram (FC) (1)} &\scriptsize{Chiranjeevi and Sengupta (2011) \cite{FF-5}}                             \\
\scriptsize{}   &\scriptsize{Multi-channel Kernel Fuzzy Correlogram (MKFC) (1)} &\scriptsize{Chiranjeevi and Sengupta (2013) \cite{FF-6}}      \\
\hline
\scriptsize{Location Features}   &\scriptsize{Location Features (6)}                    &\scriptsize{Sheikh and Shah (2005) \cite{LF-1}}       \\
\scriptsize{}  									 &\scriptsize{Invariant Moments (1)}                    &\scriptsize{Marie et al. (2011) \cite{FeatureM-1}}    \\
\hline
\scriptsize{Haar-like Features}  &\scriptsize{Haar-like features (7) }                  &\scriptsize{Klare (2008) \cite{FA-1}}                 \\
\hline
\end{tabular}}
\caption{Features in the Pixel Domain: An Overview (Part 2).} \centering
\label{Overview2}
\end{table}

\begin{table}
\scalebox{0.9}{
\begin{tabular}{|l|l|l|} 
\hline
\scriptsize{Transform Domain Features} &\scriptsize{Categories} &\scriptsize{Authors - Dates} \\
\hline
\hline
\scriptsize{Frequency Features}      &\scriptsize{\textbf{Fourier Transform  (FT)(2)}} &\scriptsize{}        \\
\cline{2-3}
\scriptsize{} &\scriptsize{Fast Fourier Transform (FFT)(1)}                            &\scriptsize{Wren and Porikli (2005) \cite{TDF-1}}        \\
\scriptsize{} &\scriptsize{Discrete Fourier transform (2D-DFT)(1)}                     &\scriptsize{Tsai and Chiu (2008) \cite{TDF-1-1}}         \\
\cline{2-3}
\scriptsize{} &\scriptsize{\textbf{Discrete Cosine Transform (DCT) (9)}}               &\scriptsize{Porikli and Wren (2015) \cite{TDF-2}}        \\
\cline{2-3}
\scriptsize{} &\scriptsize{\textbf{Wavelet Transform (WT) (28)}}                       &\scriptsize{}                                            \\
\scriptsize{} &\scriptsize{Discrete Wavelet Transform (DWT) (3)}                       &\scriptsize{Huang and Hsieh (2003) \cite{WTDF-25}}       \\
\scriptsize{} &\scriptsize{Binary Discrete Wavelet Transform (BDWT) (4)}               &\scriptsize{Gao et al. (2008) \cite{WTDF-1}}             \\
\scriptsize{} &\scriptsize{Modified directional lifting-based 9/7 DWT (MDLDWT)(1)}     &\scriptsize{Hsia and  Guo (2014) \cite{WTDF-9}}          \\
\scriptsize{} &\scriptsize{Orthogonal non-separable Wavelet (OW) (1)}                  &\scriptsize{Gao et al. (2008) \cite{WTDF-1-4}}           \\
\scriptsize{} &\scriptsize{Wavelet multi-scale Transform (2D dyadic WT) (5)}           &\scriptsize{Guan et al. (2008) \cite{WTDF-2}}            \\
\scriptsize{} &\scriptsize{Daubechies Complex Wavelet Transform (DCWT)(9)}             &\scriptsize{Jalal and Singh (2011) \cite{WTDF-4}}        \\
\scriptsize{} &\scriptsize{Multi-Resolution Wavelet Transform (MRWT) (1)}              &\scriptsize{Mendizabal and Salgado (2011) \cite{WTDF-6}} \\
\scriptsize{} &\scriptsize{Undecimated wavelet transform (2D-UWT) (3)}                 &\scriptsize{Antic et al. (2009) \cite{WTDF-30}}          \\
\scriptsize{} &\scriptsize{Three-Dimensional Discrete Wavelet Transform (3D-DWT) (1)}  &\scriptsize{Han et al. (2016) \cite{WTDF-21}}            \\
\cline{2-3}
\scriptsize{} &\scriptsize{\textbf{Curvelet Transform (CT) (1)}}                       &\scriptsize{Khare et al. (2013) \cite{TDF-20}}           \\
\cline{2-3}
\scriptsize{} &\scriptsize{\textbf{Walsh Transform (WalshT) (3)}}                      &\scriptsize{Tezuka and Nishitani (2008) \cite{TDF-30}}   \\
\cline{2-3}
\scriptsize{} &\scriptsize{\textbf{Hadamar Transform (HT) (1)}}                        &\scriptsize{Baltieri et al. (2010) \cite{TDF-50}}        \\
\cline{2-3}
\scriptsize{} &\scriptsize{\textbf{Gabor Transform (GT) (3)}}                          &\scriptsize{Xue et al. (2010) \cite{TDF-100}}            \\
\cline{2-3}
\scriptsize{} &\scriptsize{\textbf{Slant Transform (ST) (2)}}                          &\scriptsize{Haberdar and Shah (2013) \cite{TDF-200}}     \\
\cline{2-3}
\scriptsize{} &\scriptsize{\textbf{Prospective Frequency Features}}                    &\scriptsize{}                                            \\
\scriptsize{} &\scriptsize{Sparse FFT}                                                 &\scriptsize{-}                                           \\
\hline
\scriptsize{Video Compressed Features} &\scriptsize{\textbf{MPEG domain}}              &\scriptsize{}                                         \\
\scriptsize{}                          &\scriptsize{MVs based Features (1)}            &\scriptsize{Babu et al. (2004) \cite{TDF-2000}}       \\
\scriptsize{}                          &\scriptsize{Coefficient based Features (2)}    &\scriptsize{Zeng et al. (2003) \cite{TDF-2010}}       \\
\scriptsize{}                          &\scriptsize{MVs and Coefficient based Features (2)}    &\scriptsize{Porikli (2014) \cite{TDF-2020}}   \\
\cline{2-3}
\scriptsize{}              &\scriptsize{\textbf{H.264/AVC domain}}                     &\scriptsize{}                                        \\
\scriptsize{}              &\scriptsize{MacroBlock based Features (MB) (2)}            &\scriptsize{Dey and Kundu (2013) \cite{TDF-2030}}    \\
\scriptsize{}              &\scriptsize{Enhanced MacroBlock based Features (EMB) (1)}  &\scriptsize{Dey and Kundu (2016) \cite{TDF-2031}}    \\
\cline{2-3}
\scriptsize{}                &\scriptsize{\textbf{HEVC domain}}                  &\scriptsize{}                                       \\
\scriptsize{}                &\scriptsize{MVs based Features (2)}                & \scriptsize{Zhao et al. (2013) \cite{TDF-2100}}    \\
\hline
\scriptsize{Compressive Features} &\scriptsize{\textbf{Compressive measurements}} &\scriptsize{}                                       \\
\scriptsize{}    &\scriptsize{Orthonormal basis (4)}                               &\scriptsize{Cevher et al. (2008) \cite{CS-1}}       \\
\scriptsize{}    &\scriptsize{Linear compressive measurements (5)}                 &\scriptsize{Needell and Tropp (2008) \cite{CS-10}}  \\
\scriptsize{}    &\scriptsize{Cross-validation measurements (ARCS-CV) (2)}         &\scriptsize{Warnell et al. (2012) \cite{CS-30}}     \\
\scriptsize{}    &\scriptsize{Low-resolution measurements (ARCS-LR) (1)}           &\scriptsize{Warnell et al.(2012) \cite{CS-30}}      \\
\scriptsize{}    &\scriptsize{Orthonormal wavelet basis (1)}                       &\scriptsize{Li et al.(2010) \cite{CS-40}}           \\
\scriptsize{}    &\scriptsize{Wavelets transform (5)}                              &\scriptsize{Wang et al. (2015) \cite{CS-90}}        \\
\scriptsize{}    &\scriptsize{Canonical sparsity basis (1)}                        &\scriptsize{Xu and Lu (2011) \cite{CS-60}}          \\
\scriptsize{}    &\scriptsize{Random projections basis (1)}                        &\scriptsize{Shen et al. (2016) \cite{CS-101}}       \\
\scriptsize{}    &\scriptsize{Walsh-Hadamard measurements (1)}                     &\scriptsize{Liu and Pados (2016) \cite{CS-102}}     \\
\scriptsize{}    &\scriptsize{Random Gaussian or Fourier Scrambled matrices (1)}   &\scriptsize{Li and  Qi (2014) \cite{RPCA-40}}       \\
\scriptsize{}    &\scriptsize{Three-dimensional Compressive Sampling (3DCS) (2)}   &\scriptsize{Shu and Ahuja (2011) \cite{LRM-2}}      \\
\scriptsize{}    &\scriptsize{Three-Dimensional Circulant CS (3DCCS) (2)}          &\scriptsize{Kang et al. (2015) \cite{LRM-10}}       \\
\hline
\end{tabular}}
\caption{Features in a Transform Domain: An Overview.} \centering
\label{Overview3}
\end{table}

\newpage
\begin{table*}
\scalebox{1.0}{
\begin{tabular}{|l|l|l|} 
\hline
\scriptsize{Textures}&\scriptsize{Methods-Acronym-Number of papers)}&\scriptsize{Authors - Dates}\\
\hline
\hline
\multirow{1}{*}{\scriptsize{\textbf{1) Crisp Textures (Part 1)}}} & \scriptsize{}               & \scriptsize{}\\
\hline
\multirow{22}{*}{\scriptsize{Local Binary Pattern} (25 variants)}
& \scriptsize{Local Binary Pattern (LBP) (9)}                                          & \scriptsize{Heikkila et al. (2004) \cite{TF-10}}   \\
& \scriptsize{Spatio-temporal Local Binary Pattern (STLBP) (2)}                        & \scriptsize{Zhang et al. (2008) \cite{TF-13}}      \\
& \scriptsize{Epsilon Local Binary Pattern ($\epsilon$-LBP) (2)}                       & \scriptsize{Wang and Pan (2010) \cite{TF-20}}      \\
& \scriptsize{Center-Symmetric Local Binary Patterns (CS-LBP)(3)}                      & \scriptsize{Tan et al. (2010) \cite{TF-33}}        \\
& \scriptsize{Space-Time Center-Symmetric Local Binary Patterns (ST-CS-LBP) (1)}       & \scriptsize{Li et al. (2011) \cite{TF-34}}         \\
& \scriptsize{Spatial Extended Center-Symmetric Local Binary Pattern (SCS-LBP) (1)}    & \scriptsize{Xue et al. (2010) \cite{TF-21}}        \\
& \scriptsize{Hybrid Center-Symmetric Local Binary Pattern (HCS-LBP) (1)}              & \scriptsize{Xue et al. (2011) \cite{TF-23}}        \\
& \scriptsize{Spatial Color Binary Patterns (SCBP) (1)}                                & \scriptsize{Zhou et al. (2011) \cite{TF-24}}       \\
& \scriptsize{Opponent Color Local Binary Patterns (OCLBP) (1)}                        & \scriptsize{Lee et al. (2011) \cite{TF-25}}        \\
& \scriptsize{Double Local Binary Pattern (DLBP) (1)}                                  & \scriptsize{Xu et al. (2009) \cite{TF-60}}         \\
& \scriptsize{Uniform Local Binary Patterns (ULBP) (1)}                                & \scriptsize{Yuan et al. (2011) \cite{TF-26}}       \\
& \scriptsize{Extended Local Binary Patterns (Ext-LBP) (1)}                            & \scriptsize{Yu et al. (2011) \cite{TF-27}}         \\
& \scriptsize{Rotation Invariant Local Binary Patterns (RI-LBP) (1)}                   & \scriptsize{Yu et al. (2011) \cite{TF-27} }        \\
& \scriptsize{Larger Neigborhood Local Binary Patterns (LN-LBP) (1)}                   & \scriptsize{Kertesz (2011) \cite{TF-28}}           \\
& \scriptsize{Motion Vectors Local Binary Patterns (MV-LBP) (3)}                       & \scriptsize{Yang et al. (2012) \cite{TF-29}}       \\
& \scriptsize{Scene Adaptive Local Binary Pattern (SALBP) (1)}                         & \scriptsize{Noh and Jeon (2012) \cite{TF-30}}      \\
& \scriptsize{Stereo Local Binary Pattern based on Appearance and Motion (SLBP-AM) (1)}& \scriptsize{Yin et al. (2013) \cite{TF-31}}        \\
& \scriptsize{Window-Based LBP (WB-LBP) (1)}                                           & \scriptsize{Kumar  et al. (2014) \cite{TF-35}}     \\
& \scriptsize{Intensity Local Binary Pattern (iLBP)(2)}                                & \scriptsize{Vishnyakov et al. (2014) \cite{TF-36}} \\
& \scriptsize{BackGround Local Binary Pattern (BGLBP) (1)}                             & \scriptsize{Davarpanah  et al. (2015) \cite{TF-66}} \\
& \scriptsize{eXtended Center-Symmetric Local Binary Pattern (XCS-LBP) (1)}            & \scriptsize{Silva et al. (2015) \cite{TF-63}}       \\
& \scriptsize{Local SVD Binary Pattern (LSVD-BP)(1)}                                   & \scriptsize{Guo et al. (2016) \cite{TF-67}}         \\
& \scriptsize{Multi-Block Temporal-Analyzing LBP (MB-TALBP) (1)} 											 & \scriptsize{Chen et al. (2016) \cite{TF-68}}        \\
& \scriptsize{Perception-based Local Binary Pattern (P-LBP) (1)} 												& \scriptsize{Chan. (2016) \cite{TF-129-30}}         \\
\cline{2-3}
& \scriptsize{\textbf{Prospective LBP}}                                                & \scriptsize{}                                         \\
& \scriptsize{Multi-scale Region Perpendicular LBP (MRP-LBP) (1)}                      & \scriptsize{Nguyen and Miyata (2015) \cite{TF-1000}}   \\
& \scriptsize{Scale-and Orientation Adaptive LBP (SOA-LBP) (1)}                        & \scriptsize{Hegenbart and  Uhl  (2015) \cite{TF-1010}}  \\
\hline 
\multirow{7}{*}{\scriptsize{Local Ternary Pattern} (7 variants)} 
& \scriptsize{Local Ternary Pattern (LTP) (1)}                                          & \scriptsize{Liao et al. (2010) \cite{TF-70}}    \\
& \scriptsize{Scale Invariant Local Ternary Pattern (SILTP) (1)}                        & \scriptsize{Liao et al. (2010) \cite{TF-70}}    \\
& \scriptsize{Scale-invariant Center-symmetric Local Ternary Pattern(SCS-LTP) (2)}      & \scriptsize{Zhang et al. (2011) \cite{TF-71}}   \\
& \scriptsize{Multi-Channel Scale Invariant Local Ternary Pattern (MC-SILTP) (1)}       & \scriptsize{Ma and Sang (2012) \cite{TF-73}}    \\
& \scriptsize{Center Symmetric Spatio-temporal Local Ternary Pattern (CS-ST-LTP) (2)}   & \scriptsize{Xu (2013)  \cite{TF-74}}            \\
& \scriptsize{Spatio Temporal Scale Invariant Local Ternary Pattern (ST-SILTP) (1)}     & \scriptsize{Ji and Wang (2014) \cite{TF-76}}    \\
& \scriptsize{Center-Symmetric Scale Invariant Local Ternary Pattern (CS-SILTP) (1)}    & \scriptsize{Wu et al. (2014) \cite{TF-77}}      \\
\hline
\multirow{1}{*}{\scriptsize{Local States Pattern}} 
& \scriptsize{Scale Invariant Local States (SILS) (1)}                                 & \scriptsize{Yuk et al. (2011) \cite{TF-90}}           \\
\hline
\multirow{1}{*}{\scriptsize{Local Derivative Pattern}} 
& \scriptsize{SpatioTemporal Center-Symmetric Local Derivative Pattern (STCS-LDP) (1)} & \scriptsize{Jmal et al. (2010) \cite{TF-95}}         \\
\hline
\multirow{2}{*}{\scriptsize{Local Difference Pattern}} 
& \scriptsize{Local Difference Pattern (LDP) (1)}                                      & \scriptsize{Yoshinaga et al. (2010) \cite{TF-100}}    \\
& \scriptsize{Statistical Local Difference Pattern (SLDP) (2)}                         & \scriptsize{Yoshinaga et al. (2011) \cite{TF-101}}    \\
\hline
\multirow{1}{*}{\scriptsize{Local Self Similarity}}
& \scriptsize{Local Self Similarity (LSS) (1)}                                         & \scriptsize{Jodoin et al. (2012) \cite{TF-115}}       \\
\hline
\multirow{1}{*}{\scriptsize{Local Similarity Binary Pattern}} 
& \scriptsize{Local Similarity Binary Pattern (LSBP) (6)}                             & \scriptsize{Bilodeau et al. (2013) \cite{TF-110}}     \\
& \scriptsize{Uniform LSBP (U-LBSP) (1)}                                              & \scriptsize{Yan et al. (2016) \cite{TF-114-10}}       \\
\hline 
\multirow{1}{*}{\scriptsize{Directionnal Rectangular Pattern}} 
& \scriptsize{Directionnal Rectangular Pattern (DRP) (1)}                              & \scriptsize{Zhang et al. (2009) \cite{TF-117}}        \\
\hline 
\multirow{1}{*}{\scriptsize{Local Color Pattern}} 
& \scriptsize{Local Color Pattern (LCP) (3)}                                           & \scriptsize{Chua et al. (2012) \cite{TF-125}}         \\
\hline 
\multirow{1}{*}{\scriptsize{Local Neigborhood Pattern}} 
& \scriptsize{Local Neigborhood Pattern (LNP) (1)}                                    & \scriptsize{Amato et al. (2010) \cite{TF-127}}        \\
\hline 
\multirow{1}{*}{\scriptsize{Local Ratio Pattern}} 
& \scriptsize{Local Ratio Patterns (LRP) (1)}                                    & \scriptsize{Zaharescu and Jamieson (2011) \cite{MulF-40}}    \\
\hline 
\multirow{1}{*}{\scriptsize{Local Ray Pattern}}
& \scriptsize{Local Ray Pattern (LRP)} (2)                                             & \scriptsize{Shimada et al. (2013) \cite{TF-128}}      \\
\hline
\multirow{1}{*}{\scriptsize{Spatio-Temporal Vector}} 
& \scriptsize{Spatio-Temporal Vector (STV) (11)}                                      & \scriptsize{Pokrajac and Latecki (2003) \cite{TF-170}} \\
\hline
\multirow{1}{*}{\scriptsize{Spatio-Temporal Texture}} 
& \scriptsize{Space-Time Patch (ST-Patch) (2)}                                        & \scriptsize{Yumiba et al. (2011) \cite{TF-180}} \\
\hline
\multirow{1}{*}{\scriptsize{Spatio-Temporal Features}} 
& \scriptsize{Spatio-Temporal Features (STF) (3)}                                     & \scriptsize{Nonaka et al. (2012) \cite{TF-190}} \\
\hline 
\end{tabular}}
\caption{Features in the Pixel Domain (Texture Features): An Overview (Part 3).} \centering
\label{Overview4}
\end{table*}

\newpage
\begin{table*}
\scalebox{1.0}{
\begin{tabular}{|l|l|l|} 
\hline
\scriptsize{Textures}&\scriptsize{Methods-Acronym-Number of papers}&\scriptsize{Authors - Dates}\\
\hline
\hline
\multirow{1}{*}{\scriptsize{\textbf{1) Crisp Textures (Part 2)}}}        & \scriptsize{}      & \scriptsize{}\\
\hline
\hline
\multirow{1}{*}{\scriptsize{Texture Pattern Flow}} 
& \scriptsize{Texture Pattern Flow (TPF) (3)}                  & \scriptsize{Zhang et al. (2011) \cite{TF-201}} \\
\hline
\multirow{1}{*}{\scriptsize{Binary Map}} 
& \scriptsize{Binary Map (BM) (1)}                             & \scriptsize{Lai et al. (2013) \cite{MulF-2}} \\
\hline
\multirow{1}{*}{\scriptsize{Textons}} 
& \scriptsize{Textons (T) (2)}                                  & \scriptsize{Spampinato et al. (2014) \cite{TF-210}} \\
\hline
\multirow{1}{*}{\scriptsize{Galaxy Pattern}} 
& \scriptsize{Galaxy Pattern (GP) (2)}                                  & \scriptsize{Liu et al. (2013) \cite{TF-129}} \\
\hline
\multirow{1}{*}{\scriptsize{Bayer-Pattern}} 
& \scriptsize{Bayer-Pattern (BP) (1)}                                   & \scriptsize{Suhr et al. (2011) \cite{CF-2}}  \\
\hline
\multirow{1}{*}{\scriptsize{Structure-Texture Decomposition}} 
& \scriptsize{Structure-Texture Decomposition (STD) (1)}        & \scriptsize{Elharrouss et al. (2015) \cite{TF-220}} \\
\hline
\hline
\multirow{1}{*}{\scriptsize{\textbf{2) Statistical Textures}}}     & \scriptsize{}                           
& \scriptsize{} \\
\hline
\multirow{1}{*}{\scriptsize{Peripheral Increment Sign Correlation}} 
& \scriptsize{Peripheral Increment Sign Correlation (PISC) (1)}    & \scriptsize{Satoh et al. (2004) \cite{TF-1}}\\
\hline
\multirow{1}{*}{\scriptsize{Peripheral TErnary Sign Correlation}} 
& \scriptsize{Peripheral TErnary Sign Correlation (PTESC) (2)}     & \scriptsize{Yokoi (2006) \cite{TF-5}}      \\
\hline
\multirow{5}{*}{\scriptsize{Radial Reach Correlation}} 
& \scriptsize{Radial Reach Correlation (RRC) (7)}                                  & \scriptsize{Satoh et al. (2002) \cite{TF-130}}         \\
& \scriptsize{Bi-Polar Radial Reach Correlation (BP-RCC) (2)}                      & \scriptsize{Satoh  (2005) \cite{TF-132}}               \\
& \scriptsize{Fast Radial Reach Correlation (F-RRC) (2)}                           & \scriptsize{Itoh et al. (2008) \cite{TF-135}}          \\
& \scriptsize{Probabilistic Bi-Polar Radial Reach Correlation (PrBP-RCC) (1)}      & \scriptsize{Yokoi (2009) \cite{TF-136}}                \\
& \scriptsize{Adaptive Bi-Polar Radial Reach Correlation (ABP-RCC) (1)}            & \scriptsize{Miyamori et al. (2011) \cite{TF-140}}      \\
\hline
\multirow{1}{*}{\scriptsize{Radial Reach Filter}} 
& \scriptsize{Radial Reach Filter (RRF) (5)}                                       & \scriptsize{Satoh et al. (2002) \cite{TF-150}}         \\
\hline
\multirow{2}{*}{\scriptsize{Radial Proportion Filter}} 
& \scriptsize{Radial Proportion Filter (RPF) (1)}                                  & \scriptsize{Miyamori et al. (2012) \cite{TF-160}}      \\
& \scriptsize{Multi Radial Proportion Filter (MRPF) (1)}                           & \scriptsize{Miyamori et al. (2012) \cite{TF-161}}      \\
\hline
\multirow{1}{*}{\scriptsize{Statistical Reach Feature}} 
& \scriptsize{Statistical Reach Feature (SRF) (2)}                                 & \scriptsize{Iwata et al. (2009) \cite{StF-40}}         \\
\hline
\multirow{1}{*}{\scriptsize{Statistical Texture}} 
& \scriptsize{Statistical Texture (ST) (2)}                                     & \scriptsize{Chiranjeevi and Sengupta (2014) \cite{FA-33}}  \\
\hline
\hline
\multirow{1}{*}{\scriptsize{\textbf{3) Fuzzy Textures}}}                              
& \scriptsize{}                                                                    & \scriptsize{} \\
\hline
\multirow{1}{*}{\scriptsize{Local Fuzzy Pattern}}  
& \scriptsize{Local Fuzzy Pattern (LFP) (2)}                                       & \scriptsize{Ouyang et al. (2012) \cite{TF-121}}         \\
\hline
\hline
\multirow{1}{*}{\scriptsize{\textbf{4) Fuzzy Statistical Textures}}}                  
& \scriptsize{}                                                                    & \scriptsize{} \\
\hline
\multirow{1}{*}{Fuzzy Statistical Texture}  
& \scriptsize{Fuzzy Statistical Texture (FST) (2)}                                 & \scriptsize{Chiranjeevi and Sengupta (2012) \cite{FF-10}}  \\
\hline 
\end{tabular}}
\caption{Features in the Pixel Domain (Texture Features): An Overview (Part 5).} \centering
\label{Overview5}
\end{table*}

\newpage
\subsection{Features and Their Usage}
\label{subsec:FUsage}
Different features have different properties that enable them to handle critical situations such as illumination changes, motion changes and structure background changes, differently. Therefore there is a need to characterize features so that benchmarking becomes possible. In this context, reliability has been considered Furthermore, each feature can be characterized by its reliability (Section \ref{subsec:FReliability}). If more than one feature is required to be used, then there is a need for fusion schemes that can aggregate the results of each one (Section \ref{subsec:FFusion}). Finally, feature selection can be used to optimize the discrimination between the background and foreground classes (Section \ref{sec:FeatureSelection}).

\subsubsection{Feature Reliability}
\label{subsec:FReliability}
Although efforts have been made to perform reliable foreground detection, in reality, results are not 100\% reliable. This is mainly because, applications are often considered in controlled environments, where closed world assumptions could be applied. However, in general, one has to deal with a real-world scenario, which means that the background or foreground may significantly change dynamically making it literally impossible to predict possible changes. For example, the light conditions may suddenly fluctuate in some parts of the image, video compression or transmission artefacts may cause noise, a wind may cause a stationary camera to tremble, and so on. The fundamental problem lies in building an appropriate and robust background modeling that is capable of accurate detect moving objects without compromising model against background changes, and thus fail to detect all changes. \\
\indent One potential solution is in monitoring the reliability of the features by analyzing their general properties. For example Latecki et al. \cite{FR-1} considered statistical properties of feature value distributions as well as temporal properties as a source of tracking feature reliability. The proposed strategy is to estimate the deviations of the feature to the distribution and if found reliable then compute feature properties, else detect as being unreliable. As computed features will never be 100\% reliable, it is interesting to compute reliability measures. This way decisions will only be made when features are sufficiently reliable. This means that in addition to feature computation, an instantaneous evaluation of their reliability should also be made, and then adapt the decision in accordance to the detected level of reliability. For example, if the goal of the application is to monitor motion activity, and to signal an alarm if the activity is high, the system is allowed to make reliable decisions only if there exist a subset of the computed motion activity features that is sufficiently reliable. The monitoring of features reliability and adjusting the system behaviour accordingly, seems to be the best mechanism to deploy autonomous video surveillance systems. \\
\indent To determine whether a particular feature is reliable, Latecki et al. \cite{FR-1} assumed that the feature bears more information if its distribution differs more significantly from a normal (Gaussian) distribution. The assumption is that the feature becomes unreliable if an addition random noise is superimposed, which would lead the distribution of such noisy feature to become more Gaussian like. Hence, by measuring to what extent a feature distribution differs from a Gaussian distribution, it would be possible to get an notion of how useful the feature is and importantly detect if such usefulness drops. Latecki et al. \cite{FR-1} proposed an entropy-based technique for feature reliability assessment. The proposed parametric negentropy estimation, inspired from information theory, can be efficiently used to evaluate the usability of a motion measure employed for the detection of moving objects. This approach is useful for one-dimensional features and works under assumption that the noise, which corrupts the observed feature, is additive Gaussian. Some future extensions of this work could be to generalize it for the multidimensional case and for the feature reliability detection with non-Gaussian and non-Additive noise. \\
\indent In an an other study, Latecki et al. \cite{FR-2}\cite{FR-3}\cite{FR-4} described a simple temporal method to determine the reliability of motion features. The input motion feature has binary values for each $8\times8$ block with $1$ for "motion detected" and $0$ for "no motion detected". Let $f(n)$ be the number of 1s in the frame number n, i.e., $f(n)$ is the number of detected moving blocks as function of frame number. The finite difference approximation of first derivative of $f$ is used to monitor the reliability of the motion feature. If the jump in values of $f$ is above a certain threshold for a given time interval, the binary feature is unreliable in this interval. The threshold necessary to detect the unreliable features is not static and is determined by a dynamic thresholding algorithm. \\
\indent Harville et al. in \cite{SF-10} detected invalid depth and chroma components in the manner as follows: \textbf{\textit{1)}} The chroma (U and V) components become unstable when the luminance is low. So, Harville et al. \cite{SF-10} defined a chroma validity test based on luminance and set a predicate that operated it on a Gaussian model by applying it to the luminance mean of the distribution. When one of the test is failed, the chroma components of the current observation or the Gaussian distribution are not used, and \textbf{\textit{2)}} The depth computation relies on finding small area correspondences between image pairs, and therefore does not produce reliable results in regions of little visual texture and in regions, that are visible in one image but not the other. Most stereo depth implementations attempt to detect such cases and label them with one or more special values. Harville et al. \cite{SF-10} relied on the stereo depth system to detect invalid depth data and define a depth validity predicate. When the test is false, the depth is not used.\\

\indent According to the literature in this area, feature reliability has been less investigated and only the works of Latecki et al. \cite{FR-1}\cite{FR-2}\cite{FR-3}\cite{FR-4} specifically addressed this problem. Thus, the determination of the features' reliability and then when to use them is still an open problem and may be one of the main future developments in this field.

\subsubsection{Feature Fusion}
\label{subsec:FFusion}
It has been discussed that the use of multiple features could bring complementary advantages to the modeling and detection techniques. However, the feature agregation procedure requires reliable operators that can efficiently fuse features. There are several operators which can be used for feature agregation. These operators can either be of basic (logical AND, logical OR), statistical or fuzzy types (Sugeno integral \cite{FA-10}, Choquet integral \cite{FA-12}, interval valued Choquet integral \cite{FA-40}) as follows:

\begin{enumerate}
\item \textbf{Basic operators:} Logical operators such as OR and AND are the simplest ways to combine different results. AND is more useful when the aim is to suppress false positive detections which appear in one mask and not in the other. But it presents the disadvantage to suppress true positives which appear in one mask and not in the other one. On the other hand, OR is more suitable when the aim is to suppress false negative detections which appear in one mask and not in the other one. But it presents the disadvantage to add false positive which appear in one mask and not in the other one. Other basic operators include mean, median, minimum and  maximum operators, as well as some generalizations like the Ordered Weighted Average (OWA) having the minimum and the maximum as particular cases. \\

\item  \textbf{Statistical operators:} The simplest way to statistically combine features consists of a basic product formulation of the likelihoods \cite{TF-71}\cite{900} but it has the limitation that a single close to zero probability in one of the sensors may lead to the cancellation of the overall combination. In order to avoid the zero probability problem, which could lead to critical misclassification errors, Logarithmic Opinion Pool can be used as proposed by Gallego and Pardas \cite{SF-500}. Thus, by taking logarithms, a weighted average of the log-likelihoods can be obtained. The weighting factors is central to the correct working of the sensor fusion system. The weighting factors can be according to the reliability that each one of the sensors presents. An an other possible way is to exploit the similarity between foreground and background classes for each one of the sensors, assuming that \textbf{\textit{(1)}} high similarity implies that both classes are modeling the same space in a camouflage situation, and thus, the decision is not reliable, and \textbf{\textit{(2)}} small similarity implies classes separated enough to achieve a correct decision. In this idea, Gallego and Pardas \cite{SF-500} computed the Hellinger distance between the pdf's that model the $i^{th}$ pixel of the $j^{th}$-sensor color or depth for the foreground and background classes, respectively. Thus, this distance detects the degree of similarity between foreground and background models that each one of the sensors present. Moreover, the Hellinger distance presents two main characteristics that are very interesting for the application of background/foreground separation: Unlike the Bhattaharyya distance, or the Kullback-Leibler divergence, which give a similarity distance bounded between $[0,\infty)$, the Hellinger distance gives a normalized distance among models bounded between $[0,1]$. Furthermore, unlike the Kulback-Leibler divergence, the Hellinger distance is symmetric. Finally, the weighting factors are defined as a function of the Hellinger distance. Thus, sensors that present a higher degree of similarity between foreground and background classes have a close-to-zero weight, thus avoiding misclassification errors in case of color or depth camouflage problems. \\

\item \textbf{Fuzzy operators:} The family of fuzzy integrals is a generalization of the weighted average technique using the Choquet integral, as well as the minimum and the maximum using the Sugeno integral. The advantage of fuzzy integrals is that they take into account the importance of the coalition of any subset of criteria. A brief summary of the basic concepts around fuzzy integrals (Sugeno and Choquet) is described below:\\ 
\begin{itemize}
\item \textbf{Sugeno and Choquet integrals:} Let $\mu$ be a fuzzy measure on a finite set $X$ of criteria and $h:X\rightarrow\left[0,1\right]$ be a fuzzy subset of $X$. \\
\begin{de} The Sugeno integral of $h$ with respect to $\mu$ is defined by:
\begin{equation}
S_{\mu}=Max\left(Min\left(h\left(x_{\sigma\left(i\right)}\right),\mu\left(A_{\sigma\left(i\right)}\right) \right)\right)
\label{EquationSugeno}
\end{equation}
where $\sigma$ is a permutation of the indices such that \\$h_{\sigma\left(1\right)}\leq\ldots\leq h_{\sigma\left(n\right)}$ and $A_{\sigma\left(i\right)}=\left\{\sigma\left(1\right),\ldots,\sigma\left(n\right)\right\}$
\end{de} 
\begin{de} The Choquet integral of $h$ with respect to $\mu$ is defined by:
\begin{equation}
C_{\mu}=\sum_{i=0}^{n}h\left(x_{\sigma\left(i\right)}\right)\left(\mu\left(A_{\sigma\left(i\right)}\right)-\mu\left(A_{\sigma\left(i+1\right)}\right)\right)
\label{EquationChoquet}
\end{equation}
with the same notations as above.
\end{de} 
An interesting interpretation of the fuzzy integrals arises in the context of source fusion. The measure $\mu$ can be viewed as a factor that describes the relevance of the sources of information where $h$ denotes the values that the criteria has reported. The fuzzy integrals then aggregates nonlinearly the outcomes of all criteria. The Choquet integral is adapted for cardinal aggregation while Sugeno integral is more suitable for ordinal aggregation. Additionaly, Sugeno integral calculates only minimum and maximum weightage and  the Choquet integral has the same functionality as Sugeno integral but it also uses additional operations like arithmetic mean and Ordered Weighted Averaging (OWA). Thus, Choquet integral is more suitable for background/foreground separation. \\
\item \textbf{Fuzzy measures:} While fusing different criteria or sources, fuzzy measures take on an interesting interpretation. A pixel can be evaluated based on a criteria or sources providing information about the state of the pixel whether pixel corresponds to background or foreground. The more the criteria provides information about the pixel, the more relevant the decision of pixel's state. Let $X=\left\{x_{1},x_{2}, x_{3}\right\}$, with each criterion, a fuzzy measure is associated, $\mu\left(x_{1}\right)=\mu\left(\left\{x_{1}\right\}\right)$, $\mu\left(x_{2}\right)=\mu\left(\left\{x_{2}\right\}\right)$ and $\mu\left(x_{3}\right)=\mu\left(\left\{x_{3}\right\}\right)$ such that the higher the $\mu\left(x_{i}\right)$, the more important the corresponding criterion in the decision.To compute the fuzzy measure of the union of any two disjoint sets whose fuzzy measures are given, an operational version proposed by Sugeno which called $\lambda$-\textit{fuzzy measure} can be used. To avoid excessive notation, let denote this measure by $\mu_{\lambda}$-\textit{fuzzy measure}, where $\lambda$ is a paramater of the fuzzy measure used to describe an interaction between the criteria that are combined. Its value can be determined through the boundary condition, i.e. $\mu\left(X\right)=\mu\left(\left\{x_{1},x_{2}, x_{3}\right\}\right)=1$.
The fuzzy density values over a given set $K\subset X$ is computed as: \\
\begin{equation}
\mu_{\lambda}\left(K\right)=\frac{1}{\lambda}\left[\prod_{x_{i}\in K} \left(1+\lambda \mu_{\lambda}\left(x_{i}\right)\right)-1\right]
\label{EquationFuzzyDensity}
\end{equation} \\
\item \textbf{Interval-valued fuzzy integrals:} Although discrete (or real-valued) Sugeno and Choquet integrals defined in Equation \ref{EquationSugeno} and Equation \ref{EquationChoquet} can be used as a decision making operator to fuse the information from multiple sources, it does not consider the uncertainty. But in practice information sources have wide range of possible values (i.e., high uncertainty) and hence cannot be represented by a single number. To solve this problem, interval-valued fuzzy sets (IVFSs) \cite{FA-40-1}\cite{FA-40-2}\cite{FA-40-3} can be used to model the uncertainty in these values. Thus, the value are represented as an interval. In this context, an aggregation operator is needed to integrate the information sources, represented by IVFSs. Thus, modifications of the discrete integrals called interval-valued integrals are used as in Chiranjeevi and Sengupta \cite{FA-40}. \\
\end{itemize}
\end{enumerate}

\indent Dempster-Shafer theory can be also used in feature fusion as in the work of Munteanu et al. \cite{FA-70}. According to the literature in this area, these different feature fusion schemes have been applied when multiple features are used as can be seen in Section \ref{sec:MultipleFeatures} and Table \ref{FFOverview}. 

\subsubsection{Feature Selection}
\label{subsec:FSelection}
As seen in Section \ref{subsection:FPD}, there is not a unique feature that performs better than any other feature independently of the background and foreground properties because each feature has its strenghtness and weakness against each challenge. Thus, a way to take advantage of the properties of each feature is to perform feature selection. The aim is to use the best feature or the best combination of features on a per-pixel \cite{FS-1}\cite{FS-2}\cite{FS-3}\cite{FS-10}\cite{FS-30} or per-block \cite{TF-25} basis. Thus, ensemble learning methods such as the boosting classifier are suitable for feature selection. Boosting algorithms usually generate a weighted linear combination of some weak classifiers that perform only a little better than random guess. So, weak classifiers can be learned from the feature values at a pixel and combined to perform better than the others alone. This combination produces a strong classifier. Thus, this method can effectively select different features at each pixel to distinguish foreground objects from the background. \\
\indent Extensions of this conventional algorithm are available in the form of on-line boosting algorithms \cite{FS-1}\cite{FS-2}\cite{FS-3} which use several classifier pools, and each pool contains several weak classifiers. Once an input image is given, each classifier pool selects the best classifier for the given image. The selected classifiers form a strong classifier group, and the final classification is performed using those strong classifiers. At the same time, each classifier pool selects the worst classifier as well. The worst classifier is replaced with a randomly selected classifier so that a better classifier can be included in the classifier pool. Instead of selecting the best classifier from each classifier pool as the previous method does, an improvement according to \cite{TF-25} selects several good classifiers from each pool. While the previous method replaces the worst classifier in each pool, instead this improvement replaces several bad classifiers. \\
\indent According to the literature in this area, feature selection has been less investigated in backgroud modeling and foreground detection methods with only 9 papers. Practically, only five approaches have so far been used in the literature: \textbf{\textit{(1)}} Adaboost \cite{FS-1-1} used with the classifier-based background model \cite{FS-1}\cite{FS-2}\cite{FS-3}\cite{TF-25}, \textbf{\textit{(2)}} Realboost \cite{FS-10-1} used with the KDE model \cite{FS-10}, \textbf{\textit{(3)}} dynamic feature selection \cite{FS-40} with OR-PCA model \cite{FS-41}, \textbf{\textit{(4)}} generic feature selection \cite{FS-50} with the ViBe model \cite{FS-51}, and \textit{5)} One-class SVM \cite{FS-30}. These different approaches and their characteristics are analyzed in Section \ref{sec:FeatureSelection}.

\subsubsection{Feature Relevance}
\label{subsec:FRelevance}
To choose the most discriminative features in a multiple features or feature selection scheme, feature relevance may be address. The one work which concerns feature relevance is the work of Molina-Giraldo et al. \cite{FR-10}\cite{FR-11}. The feature relevance analysis is made through a Principal Component Analysis (PCA), searching for directions with greater variance to project the data. Thus, the relevance of the original features is quantified with weighting factors. Finally, Molina-Giraldo et al. \cite{FR-10}\cite{FR-11} developed a background subtraction method based a multi-kernel learning in which the weight are selected from the feature relevance analysis. Experimental results \cite{FR-10}\cite{FR-11} on the I2R dataset \cite{900} show that the proposed Weighted Gaussian Kernel Video Segmentation (WGKVS) model outperforms SOBS \cite{CF-203-1}.

\subsubsection{Features and Challenges}
\label{subsec:FeaturesChallenges}
In this section, we grouped all the advantages and disavantages of the different features in terms of robustness against the different challenges met in video and detailed in Bouwmans \cite{JournalCOSREV2014-1}, and they can be summarized as follows: \\
\begin{itemize}
\item \textbf{Color features:} Although intensity and color features are often very discriminative features and allow basic foreground detection, they are not robust in challenges such as illumination changes, foreground aperture, camouflage in color and shadows. However, intensity can be used in complementarity of color to deal with different color problems such as dark foreground and light foreground. Furthermore, this combination solves saturation problems and minimum intensity problems \cite{MulF-32}, and reduces the number of false negatives, false positives and increase true positives. But, the intensity as colors can not work with intense shadows and highlight that often occur in indoor and outdoor scenes, and in presence of gradual or sudden illumination changes \cite{IF-300}. Then, different strategies can be found in literature to alleviate the limitations of the basic color spaces: \textbf{(1)} the use of well-known color spaces which separate the luminance and the chrominance information such as HSV and YCrCb, \textbf{(2)} the use of designed shape color space models such as the cylinder color model \cite{CF-300}\cite{CF-301}\cite{CF-310}\cite{CF-320}, the hybrid cone-cylinder \cite{CF-315}\cite{CF-316}, the ellipsoidal color model \cite{CF-330}, the box-based color model \cite{CF-340}, and the double-trapezium cylinder model \cite{CF-350}, \textbf{(3)} the use of characteristics in addition of the intensity or color value (mean, variance, minimum, maximum, etc..) (See Section \ref{sec:MultipleCharacteristics}), \textbf{(4)} the use of designed illumination invariant intensity or color features obtained by normalization \cite{IF-100}\cite{IF-110}\cite{IF-120}\cite{IF-140}, \textbf{(5)} the use of illumination compensation methods \cite{IF-160}\cite{IF-200}\cite{IF-220}\cite{IF-230}\cite{IF-310}\cite{IF-320}\cite{IF-330}\cite{IF-340}\cite{IF-350}\cite{CDFeatureI-4}, and \textbf{(6)} the addition of other features (See Section \ref{sec:MultipleFeatures}). Normalization based features sacrifice discriminability while texture features cannot operate on texture-less regions. Both types of features produce large missing regions in the foreground mask. \\
\item \textbf{Edge features:} Edge features are obtained with edge detectors which operate on the difference between neighboring pixels, hence an edge detector should be reasonably insensitive to global shifts in the mean level, i.e. to global illumination changes. Therefore it is interesting to run background/foreground separation algorithms on the output from edge detectors, hopefully reducing the effects of rapid illumination changes. So, the edge could  handles the local illumination changes but also the ghost leaved when waking foreground objects begin to move. However, edge features are not sufficiently good to segment the foreground objects isolatedly. Indeed, edge features can sometimes handle dark and light camouflage problems and it is less sensitive to global illumination changes than color feature \cite{MulF-30}. Nevertheless, problems like noise, false negative edges due to local illumination prob-lems, foreground aperture and camouflage do not allow  an accurate foreground detection. Furthermore, due to the fact that it is sometimes difficult to segment the foreground object borders, it is not possible to fill the objects, and solve the foreground aperture problem. Since it is not possible to handle dark and light camouflage problems only by using edges due to the foreground aperture difficulty, the brightness of colour model is used to solve this problem and help to fill the foreground objects.\\
\item \textbf{Texture features:} Texture features allow to be robust in presence of shadows and gradual illumination changes, and sometimes in dynamic backgrounds. Texture features can produce false detections due to textures induced by local illumination effects like in cast shadows. Furthermore, an algorithm based only on texture may cause detection errors in regions of blank texture and heterogeneous texture. \\
\item \textbf{Motion features:} Motion features can handle irrelevant background motion and clutter such as waving trees and waves. \\
\item \textbf{Stereo features:} Stereo features allow the model to deal with the camouflage in color but not in depth. \\
\end{itemize}
Thus, multiple features appproaches with two, three or a set of features obtained from a bag-of features or by feature selection are suitable to address multiple challenges in the same video (See Section \ref{sec:MultipleFeatures}). A representive work developed by Li et al. \cite{900} consists in a sets of features built following the type of background (static or dynamic) as follows:
\begin{itemize}
\item \textbf{Features for static background pixels:} For modeling pixels belonging to a stationary background object, the stable and most significant features are its color and local structure (gradient). As the gradient is less sensitive to illumination changes, the two types of feature vectors are integrated under the Bayes framework in the basic product formulation of the likelihoods. \\
\item \textbf{Features for dynamic background pixels:} For modeling dynamic background pixels associated with nonstationary objects, color co-occurrences are used as their dynamic features. This is because the color co-occurrence between consecutive frames has been found to be suitable to describe the dynamic features associated with nonstationary background objects, such as moving tree branches or a flickering screen. \\
\end{itemize}

\subsubsection{Features and Strategies}
\label{subsec:FeaturesStrategies}
There are several strategies in literature such as multi-scales strategies, multi-levels strategies, multi-resolutions strategies, multi-layers strategies, hierarchical strategies, and coarse-to-fine strategies (See Section \ref{subsec:CSize}). Practically, different features can be used following the scale, the level or the resolution. For example, a feature can be used at the block level (such as Haar-like features in \cite{TDF-3100}), and other features can be used at the pixel level (such as RGB in \cite{TDF-3100}). Thus, these strategies employed multiple features schemes. Please see Table \ref{MFOverview-1}, Table \ref{MFOverview-2}, and Table \ref{MFOverview-3} for a quick overview. \\

\subsubsection{Features and Similarities}
\label{subsec:FeaturesSimilarities}
The foreground mask is usually obtained by thresholding with a fixed, statistical or fuzzy threshold the difference between the value of the feature in the background model and the current frame. The value of a feature can be a scalar (intensity value, etc..), a vector (2D spatial vector, 3D spatiotemporal vector, etc...) or a histogram (correlogram, etc..). Practically, comparison of features can be made by using similarities obtained with \textbf{1)} a crisp, statistical or fuzzy distance for scalar cases, \textbf{2)} a ratio for scalar cases, \textbf{3)} linear dependence measure for vector cases, and  \textbf{4)} a interesection measure for histogram (correlogram) case.  The choice of the suitable similarity is guided by the properties and the distribution of the concerned features. Furthermore, spatial and temporal features such as LBP and LTP need also measures for their computing as follows: \textbf{1)} a measure for the distance in the spatial neighborhood, and \textbf{2)} a measure for the distance in the temporal neighborhood. Thus, for spatial and temporal features like texture, it needs to choose three distances. We list below the different measures and similarities found in the literature (See Table \ref{OverviewSimilarities} for a quick overview):\\

\textbf{A) Similarities for scalar case:}  Scalar value is the most common case in the literature and the similarities used can be classified as follows: \\
\begin{itemize}
\item \textbf{Difference:} The difference computed in a pixel-wise manner between the feature in the background model and the current frame is the most measure used. Thus, the difference is obtained by a distance and then a threshold is used to classifiy the pixel as background or foreground: \\
\begin{equation}
distance(B(x,y)-I(x,y)) < T
\end{equation}
where $B(x,y)$ and $I(x,y)$ are the values of the feature in the background image and in the current image, respectively. $distance(,)$ is a distance function. Several distance functions have been used in the literature and they can be classified as follows: \\
\begin{enumerate} 
\item \textbf{Crisp distance:} The most common distance function used for intensity/color values is the absolute distance \cite{CDFeatureI-300}\cite{CDFeatureI-303}. Aach et al. \cite{CDFeatureI-402} used a total least squares distance measure. In an other work, Yadav and Sing quasi-euclidian distance.  To compare Spatiotemporal Condition Information (SCI), Wang et al. \cite{FSI-3} designed a specific measure called neighborhood weighted spatiotemporal condition information (NWSCI). Using compressive features \cite{CF-1000}, Yang et al. \cite{CF-1001} developed a (Pixel-to-Model) P2M distance. \\
\item \textbf{Statistical distance:}  St-Charles and  Bilodeau \cite{TF-111} employed the Hamming distance to compare LSBPs. Mukherjee  et al. \cite{StF-131} developed a distance measure based on support weight to compare RGB features.\\
\item \textbf{Order-Consistency Measure:}  Xie et al. \cite{CDFeatureI-1} used an explicit model for the camera response function, the camera noise model, and illumination prior. Assuming a monotone and nonlinear camera response function, Xie et al. \cite{CDFeatureI-1} show that the sign of the difference between two pixel measurements is maintained across global illumination changes. Noise statistics are used to transform each frameinto a confidence frame where each pixel is replaced by a probability that it is likely to keep its sign with respect to the most different pixel in its neighborhood. Hence, an order consistency measure is defined as a distance between two distributions. Xie et al. \cite{CDFeatureI-1} used the Bhattacharyya distance due to  its properties to the Bayes error. Finally, an Illumination Invariant Change Detector via order consistency (IICD-OC) is developed. Experimental results \cite{CDFeatureI-1} on videos taken by an omni-directional camera show the robustness of IICD-OC against illumination changes. But, the ordinal measure required a reordering of blocks and it is computationally expensive. To solve this problem, Singh et al. \cite{CDFeatureI-3} explicitly modeled noise under which rank-consistency is tested, and used a probabilistic generative model under which frame blocks are generated. The order-consistency is posed  as a hypothesis validation problem using fast significance testing based on PAV. In a further work, Parameswaran et al. \cite{CDFeatureI-4} used the same order-consistency measure in an illumination compensation approach.\\
\end{enumerate} 
\item \textbf{Ratio:} The ratio computed in a pixel-wise manner between the feature in the background model and the current frame is also used in several works. Thus, the ratio is obtained by a division and then a threshold is used to classifiy the pixel as background or foreground:
\begin{equation}
ratio(B(x,y),I(x,y)) < T
\end{equation}
where $B(x,y)$ and $I(x,y)$ are the values of the feature in the background image and in the current image, respectively. For example, Baf et al. \cite{FA-12}\cite{FA-13}\cite{FA-15} computed the ratio of color components and LBP that are further aggregated with the Choquet integral. In a further work, Baf et al. \cite{FA-14} used the ratio of the IR intensity and LBP. In a similar approach, Ding et al. \cite{FA-20}\cite{FA-21} computed the ratio between the difference and the number of gray level for the color components and LBP. In a further work, Ding et al. \cite{FA-22} developed a specific ratio measure to compare gradients. This measure is a ratio between a product and a sum of the gradient in the background and the current images. Azab et al. \cite{FA-30} used a similar scheme than Ding et al. \cite{FA-20}\cite{FA-21} but with statistical values. In a change detection approach,  Aach et al. \cite{CDFeatureI-301} and Aach an Kaup \cite{CDFeatureI-302} used a ratio of probabilities. \\
\end{itemize}

\textbf{B) Similarities for vector case:} Linear dependence measures and colinearity measures are the most used similarity measure to compare vector. First, Durucan and Ebrahimi \cite{CDFeatureI-100} proposed to use a linear dependence measure (colinearity measure) to test the depence/independence properties of each vector that represented information on the neighborhood region of the concerned pixel. The idea is that illumination changes do no change the colinearity of the vector. Thus, a Linear Dependence Detector (LLD) which is invariant to transformations was defined for change detection, and applied to the vector composed by the reflectance for change detection. Experimental results \cite{CDFeatureI-100} show that LDD outperforms the Statistical Change Detector (SCD) developed by Aach et al. \cite{CDFeatureI-300} in presence of noise as well as global illumination changes and local shadows and reflection. LLD presents the advantage to detect semantic objects but object interiors maybe not well detected and it requires high computational complexity. To solve these problems, Durucan and Ebrahimi \cite{CDFeatureI-101} used a more rigorous test based on the Wronskian determinant. The corresponding detector called Wronskian Detector (WD) outperforms both SCD and LDD. In a further work, Durucan and Ebrahimi \cite{CDFeatureI-102}\cite{CDFeatureI-103} used an other test based on the Gramian determinant which provides low computational cost and is easy to implement. Thus, Durucan and Ebrahimi \cite{CDFeatureI-102}\cite{CDFeatureI-103} proposed a Gramian Detector (GD) which can be applied on color images in the RGB color space. In practice, other than illumination changes, change detection is also influenced by the noises of cameras and reflections but the previous linear algebra detectors have intrinsic weakness in case of noises \cite{CDFeatureI-110}.  
To solve this problem, Gao et al. \cite{CDFeatureI-110} developed a Linear Approximation Detector (LAD). Experimental results on the PETS 2001 dataset show that LAD show more robustness against noise than LDD and WD. In an other work, Ming et al. \cite{CDFeatureI-200}\cite{CDFeatureI-201} proposed a local linear dependence based Cauchy Statistical Model (LLD-CSM). Experimental results \cite{CDFeatureI-200}\cite{CDFeatureI-201} demonstrate that LLD-CSM can tolerate the local or global slow or sudden illumination changes, noise due to small motion in the background. \\

\textbf{C) Similarities for histogram case:} Intersection measures are the most used similarity measure to compare histograms or correlograms. First, Mason and Duric \cite{StF-2} used an intersection measure to compare Local Color Histograms (LCH). In an other work, Fabian \cite{StF-120} first used simple metric to compare Histograms of Oriented Gradients (HOG) in a discrete metric space. But, this metric made no difference between two different bins. Due solve to this problem, Fabian \cite{StF-120} proposed a complex metric. Several works \cite{FF-1}\cite{StF-3}\cite{StF-5} \cite{StF-10}\cite{StF-11}\cite{StF-12}\cite{StF-130} used the Bhattacharyya distance as can be seen in Table \ref{OverviewSimilarities}. The chi-squared measure is used to compare local gradient histograms in Mason and Duric\cite{StF-2}. The Bhattacharyya distance is the most statistical used distance and was employed for the following histograms: 3D HRI \cite{FF-1}, Local Kernel Color Histograms \cite{StF-3}, ELKH \cite{StF-5}, LDH \cite{StF-10}\cite{StF-11}, LH-FGs \cite{StF-12} and adaptive HOG \cite{StF-130}. In an other approach, Mukherjee  et al. \cite{StF-131} developed a HoG distance.\\

\begin{table*}
\scalebox{0.70}{
\begin{tabular}{|l|l|l|} 
\hline
\scriptsize{Similarities} &\scriptsize{}                                &\scriptsize{Features} \\
\hline
\hline
\scriptsize{\textbf{Scalar Case}} &\scriptsize{}                         &\scriptsize{} \\
\hline
\scriptsize{Difference}  &\scriptsize{\textbf{1) Crisp Measure}}        &\scriptsize{} \\
\scriptsize{}            &\scriptsize{Absolute Distance}                 &\scriptsize{Reflectance \cite{CDFeatureI-303}, Intensity, Color} \\
\scriptsize{}            &\scriptsize{Total Least Squares Distance}      &\scriptsize{Reflectance \cite{CDFeatureI-402}} \\
\scriptsize{}            &\scriptsize{Quasi-Euclidian Distance}          &\scriptsize{RGB \cite{IF-75-100}} \\
\scriptsize{}            &\scriptsize{Cosine Measure}                    &\scriptsize{Gradient \cite{FA-10}} \\
\scriptsize{}            &\scriptsize{Exponential Cosine Measure}        &\scriptsize{Gradient \cite{FA-10}} \\
\scriptsize{}            &\scriptsize{Angular Deviation}                 &\scriptsize{Intensity Vector \cite{MulF-7}} \\
\scriptsize{}            &\scriptsize{P2M distance}                      &\scriptsize{Compressive Feature \cite{CF-1000}\cite{CF-1001}} \\
\scriptsize{}            &\scriptsize{Neighborhood Weighted Spatiotemporal Condition Information (NWSCI)} &\scriptsize{SCI \cite{FSI-3}} \\
\cline{2-3}
\scriptsize{}     &\scriptsize{\textbf{2) Statistical Measure}}         &\scriptsize{}     \\
\scriptsize{}     &\scriptsize{Mahalanobis distance}                    &\scriptsize{}     \\
\scriptsize{}     &\scriptsize{Hamming distance}                        &\scriptsize{LSBP \cite{TF-111}} \\
\scriptsize{}     &\scriptsize{Distance Measure based on Support Weight}   &\scriptsize{RGB \cite{StF-131}} \\
\cline{2-3}
\scriptsize{}     &\scriptsize{\textbf{3) Fuzzy Measure}}             &\scriptsize{}     \\
\scriptsize{}     &\scriptsize{Interval-valued fuzzy similarity}      &\scriptsize{FST Features \cite{FA-40}}     \\
\hline
\scriptsize{Ratio}  &\scriptsize{\textbf{1) Division with the direct values}}  &\scriptsize{IR Intensity \cite{FA-14}}                \\
\scriptsize{} &\scriptsize{}                                                   &\scriptsize{RGB \cite{FA-13}, YCrCb \cite{FA-13}}     \\
\scriptsize{} &\scriptsize{}                                                   &\scriptsize{LBP \cite{FA-13}, ULBP \cite{FA-34}}      \\
\scriptsize{} &\scriptsize{}                                                   &\scriptsize{Intensity Vector Magnitude \cite{MulF-7}} \\
\cline{2-3}
\scriptsize{} &\scriptsize{\textbf{2) Division with differences}}      &\scriptsize{\textit{Crisp values:}} \\
\scriptsize{} &\scriptsize{}                                           &\scriptsize{RGB \cite{FA-20}, YCrCb \cite{FA-20}, HSI \cite{FA-20}}   \\
\scriptsize{} &\scriptsize{}                                           &\scriptsize{LBP \cite{FA-20}}                                         \\
\scriptsize{} &\scriptsize{}                                           &\scriptsize{\textit{Statistical values:}}       \\
\scriptsize{} &\scriptsize{}                                           &\scriptsize{RGB \cite{FA-30}, LBP \cite{FA-30}} \\
\cline{2-3}
\scriptsize{} &\scriptsize{\textbf{3) Division with other values}}       &\scriptsize{\textit{Product/Difference:}}                  \\
\scriptsize{} &\scriptsize{}                                             &\scriptsize{Gradient \cite{FA-22}, LBP \cite{FA-20}}       \\
\scriptsize{} &\scriptsize{}                                             &\scriptsize{LBP \cite{FA-20}}                              \\
\scriptsize{} &\scriptsize{}                                             &\scriptsize{\textit{Probabilities:} \cite{CDFeatureI-302}} \\
\hline
\hline
\scriptsize{\textbf{Vector Case}} &\scriptsize{}                                                       &\scriptsize{} \\
\hline
\scriptsize{Linear Dependence Measure} &\scriptsize{\textbf{1) Linear Dependence Measure}}             &\scriptsize{} \\
\scriptsize{}    &\scriptsize{LDD \cite{CDFeatureI-100}}                                               &\scriptsize{Reflectance (Vector)} \\
\scriptsize{}    &\scriptsize{Wronskian Detector (WD) \cite{CDFeatureI-101}}                           &\scriptsize{Reflectance (Vector)} \\
\scriptsize{}    &\scriptsize{Gramian Detector (GD) \cite{CDFeatureI-102}\cite{CDFeatureI-103}}        &\scriptsize{Color (Vector)} \\
\cline{2-3}
\scriptsize{}    &\scriptsize{\textbf{3) Linear Approximation Measure}}                                &\scriptsize{} \\
\scriptsize{}    &\scriptsize{Linear Approximation Detector (LAD) \cite{CDFeatureI-110}}               &\scriptsize{Reflectance (Vector)} \\
\cline{2-3}
\scriptsize{}    &\scriptsize{\textbf{4) Local Linear Dependence Measure}}                                            &\scriptsize{} \\
\scriptsize{}    &\scriptsize{Local Linear Dependence Measure (LLD-CSM) \cite{CDFeatureI-200}\cite{CDFeatureI-201}}   &\scriptsize{YCbCr (Vector)} \\
\cline{2-3}
\scriptsize{}    &\scriptsize{\textbf{5) Spectral Distance Measure}}              &\scriptsize{} \\
\scriptsize{}    &\scriptsize{Spectral Distance Measure \cite{906}}             	&\scriptsize{Spectral Feature (Vector)} \\
\hline
\scriptsize{Order Consistency Measure} &\scriptsize{\textbf{Order-Consistency Measure}}               &\scriptsize{} \\
\scriptsize{}    &\scriptsize{Bhattacharyya distance \cite{CDFeatureI-1}}                             &\scriptsize{Reflectance} \\
\scriptsize{}    &\scriptsize{PAV \cite{CDFeatureI-3}}                                                &\scriptsize{Color}       \\
\scriptsize{}    &\scriptsize{qPAV \cite{CDFeatureI-3}}                                               &\scriptsize{Color}      \\
\hline
\hline
\scriptsize{\textbf{Histogram Case}} &\scriptsize{}                   &\scriptsize{} \\
\hline
\scriptsize{}    &\scriptsize{\textbf{1) Intersection Measure}}       &\scriptsize{LCH \cite{StF-2}}    \\
\scriptsize{}    &\scriptsize{Simple metric (discrete metric space)}  &\scriptsize{HOG \cite{StF-120}}  \\
\scriptsize{}    &\scriptsize{Complex metric (discrete metric space)} &\scriptsize{HOG \cite{StF-120}}  \\
\scriptsize{}    &\scriptsize{Normalized Histogram Intersection}      &\scriptsize{LFCH \cite{StF-200}} \\

\cline{2-3}
\scriptsize{}    &\scriptsize{\textbf{2) Proximity Measure}}          &\scriptsize{LBP \cite{FA-30}}    \\
\cline{2-3}
\scriptsize{}    &\scriptsize{\textbf{3) Bhattacharyya distance}}     &\scriptsize{3D HRI \cite{FF-1}, Local Kernel Color Histograms \cite{StF-3}} \\
\scriptsize{}    &\scriptsize{}                            &\scriptsize{ELKH \cite{StF-5}, LDH \cite{StF-10}\cite{StF-11}, LH-FGs \cite{StF-12}} \\
\scriptsize{}    &\scriptsize{}                            &\scriptsize{adaptive HOG \cite{StF-130}} \\
\cline{2-3}
\scriptsize{}    &\scriptsize{\textbf{4) Chi-Squared Measure}}                    &\scriptsize{LGH \cite{StF-2}}  \\
\cline{2-3}
\scriptsize{}    &\scriptsize{\textbf{5) HoG distance}}                           &\scriptsize{HOG \cite{StF-131}} \\
\hline \end{tabular}}
\caption{Similarities Measures: An Overview. The first column indicates the type of the measure for each category of feature in terms of value (scalar, vector and histogram). The second column gives the name of the measures and the third column indicates the corresponding features on which the measure is applied.}
\label{OverviewSimilarities}
\end{table*}

In the following sections, we present and analyze the different features currently used in background modeling and foreground detection in terms of robustness against the challenges met in videos taken by a fixed cameras.\\

\newpage
\section{Intensity Features}
\label{sec:IntensityFeatures}
Intensity features are the most basic features that can be provide by gray-level cameras \cite{IF-1}\cite{IF-2}\cite{IF-3}\cite{IF-4} or infra-red (IR) cameras \cite{IF-10}\cite{IF-11}\cite{IF-12}\cite{IF-60}\cite{IF-70}\cite{IF-71}\cite{IF-75}\cite{IF-75-1}\cite{IF-76}\\cite{TF-179-2}. Practically, several works combined both of them by using a fusion scheme strategy \cite{IF-13}\cite{IF-14}\cite{IF-15}\cite{IF-16}\cite{IF-30}\cite{IF-31}\cite{IF-31-1}\cite{IF-32}\cite{IF-33}\cite{IF-34}\cite{IF-40}\cite{IF-50}
\cite{IF-51} to combine their respective advantages. In other works, several designed illumination invariant features based on intensity have been developed as follows:\\
\begin{itemize}
\item \textbf{Reflectance:} Several reflectance models have been developed to describe the reflectance due to normal, forescatter and backscatter  distributions such as the Lambertian model, the Phong shading model, the dichromatic reflection model \cite{CDFeatureI-500-1} and the Ward model \cite{IF-31-100}. Toth et al. \cite{CDFeatureI-303} were the first authors who used the reflectance in change detection and they proposed to obtain the illumination and the reflectance components with a homomorphic filtering \cite{CDFeatureI-310} based on the \textit{Lambertian model}. Then, only the reflectance components are compared by using a sum of absolute differences within a sliding window for foreground detection between consecutive frames. Toth and Aach \cite{CDFeatureI-305} used this approach to detect and further distinguish between humans, vehicles, and background clutters. In further works, Toth et al. \cite{CDFeatureI-304}\cite{CDFeatureI-401} improved this method by using a Bayesian framework. In an other way, Aach et al. \cite{CDFeatureI-402} used a total least squares distance measure instead of sum of absolute differences. Aach and Condurache \cite{CDFeatureI-410} used a threshold transform by significance invariance. \\
\item \textbf{Surface Spectral Reflectance:} Sedky et al. \cite{CDFeatureI-500} developed a method called 
Spectral-360 which adopted a physics-based model called the \textit{dichromatic reflection model} \cite{CDFeatureI-500-1}. This approach is different from  the  previous  work,  in that it relies on models, which can represent wide classes of  surface materials. Then, the feature used is the Surface Spectral Reflectance (SSR). Thus, the background model based on the mean SSR as  well  as  its maximum  correlation   and   its minimum  correlation, is built during a training step and updated continuously. Experimental results \cite{CDFeatureI-500} on the ChangeDetection.net 2014 dataset show that Spectral-360 outperforms MOG \cite{CF-1} and KDE \cite{204} in terms of F1-measure. In IR camera, Nadimi and Bhanu \cite{IF-30} obtained the reflectance throught the \textit{dichromatic reflection model} too, and they combined visible and IR intensity feature via a physics-based fusion. In further works, Nadimi and Bhanu \cite{IF-31}\cite{IF-31-1} used an evolutionary-based fusion method. \\
\item \textbf{Radiance:} Based on the \textit{Phong shading model}, Xie et al. \cite{CDFeatureI-1} used the radiance obtained by considering both ambient and diffuse reflection, and thus take into account an explicit model for the camera response function, the camera  noise model,
and illumination prior. Then, Xie et al. \cite{CDFeatureI-1} developed an illumination-invariant foreground detection
via order consistency. \\
\item \textbf{Photometric Variations:} Di Stefano et al. \cite{CDFeatureI-2} adopted a visual correspondence measure called 
Matching Function (MF) \cite{CDFeatureI-2}. This measure matches corresponding blocks of two images by checking an ordering constraint. Since photometric variations tend to respect the ordering of intensities in a neighborhood of pixels, MF allows to handle sudden and strong illumination variations between the background scene and the current frame. Practically, MF matched the high contrast regions corresponding to the intensity edges of two blocks, since only high intensity differences can provide high contributions to the correlation score. In a second step, Di Stefano et al. \cite{CDFeatureI-2} used a tonal alignment technique. Hence,
the background is tonally aligned to the current frame by applying a
histogram specification method \cite{CDFeatureI-2-100}, and then Di Stefano et al. \cite{CDFeatureI-2} obtained a background image where the photometric distortions have been removed. Finally, a pixelwise difference between the background image and the current frame extracts the foreground regions. \\ 
\end{itemize}

\section{Color Features}
\label{sec:ColorFeatures}
Color features provide \textit{\textbf{spectral information}}, and are the natural features as they are directly available from the sensor or the camera. Although they are often used for facilitating easy discrimination between the background and the foreground, color features are generally not robust against illumination conditions and shadows cast by the moving objects. Furthermore, similar colors between background and foreground lead to the well-known problem of camouflage in color. 

\subsection{Features in Well-Known Color Spaces}
Several color features in different color space have been proposed in the literature and are described as follows: \\
\begin{itemize}
\item \textbf{RGB color space:} The RGB color space is the most popularly used feature due to their direct availability from the sensor or the camera. Red, Green, and Blue channels of each pixel are usually measured with 8-bits resolution, where $0$ is no color (black) and $255$ is the maximum color (white), therefore, a total of 24-bit true color definition. But the RGB color space has several limitations: \textbf{\textit{1)}} it is well known that the RGB color space does not reflect the true similarities among colors, \textbf{\textit{2)}} depending on the scene one color component could be more informative than the others, so it should be given more importance than others, \textbf{\textit{3)}} the three components are dependent on each other which increase its sensitivity to illumination changes. For example, global illumination changes shift the mean level of the entire RGB image, possibly with shifts of different magnitude for each color component, and \textbf{\textit{4)}} as the three channel components are correlated, there is a need to compute inter-correlation terms in the covariance matrix which shall be incorporated into existing background models such as in the MOG model \cite{CF-1}. Stauffer and Grimson \cite{CF-1} demonstrated that by not computing these inter-correlations terms, computational speed improves with increased false detections. \\
\item \textbf{Normalized RGB (rgb) color space:} The normalized RGB space is derived from the traditional RGB color space to be illumination invariant. Xu and Ellis \cite{CF-10} used the normalized RGB to allow the MOG to be robust to fast illumination changes in an outdoor environment lit by sunlight and shadowed by clouds.  \\
\item \textbf{YUV color space:} The YUV space separates luminance and chroma and so it is more suitable for improving the robustness of the model against illumination changes. For example, Wren et al. \cite{CF-50} used the normalised components, U/Y and V/Y to remove shadows in a relatively static indoor scene. Using the MOG model, Harville et al. \cite{SF-100} defined a chroma validity test based on the luminance Y as the chroma (U and V) components become unstable when the luminance is low. When the test is not verified, the chroma components of the current observation are not used and so are its current Gaussian distributions. Furthermore, the detection in luminance was combined with the detection in depth improve robustness to color camouflages.  \\
\item  \textbf{HSV color space:} The HSV color space is used to improve the discrimination between shadow and object, classifying shadows as those pixels having approximately the same hue and saturation values compared to the background, but lower luminosity. For example, Sun et al. \cite{CF-60} used the Hue-Saturation-Value (HSV) color space, because the likelihood term in the MOG model shows stronger contrast in HSV space rather than the RGB space, especially for objects that share similar appearance to the background (Camouflage in color). In an other study, Kanprachar and Tangkawanit \cite{CF-201} compared the RGB and HSV color spaces. Experimentally, HSV color space was very suitable under low illumination intensity conditions. Kanprachar and Tangkawanit \cite{CF-201} showed that not all three parameters in HSV color system are useful for the detection. Saturation (S) and Value (V) are the two key parameters to be used. \\
\item  \textbf{HSI color space:} HSI color space is closer to human interpretation of colors in the sense that brightness, or intensity, is separated from the base color. HSI uses polar coordinates. In the original MOG model, shadows are extracted as part of object mask when using the RGB color space. To address this problem, Wang and Wu \cite{CF-70} used the HSI color space which tends to be shadow-removable. However, the obtained results are not satisfactory due to the fragmented segmentation obtained by using hue and saturation. In order to achieve both "shadow-rejection" and "segmentation stability over time", Wang and Wu \cite{CF-70} employed the MOG on chroma (hue and saturation) and luma (intensity) separately. The fused results obtained by combining chroma and luma is prepared using two criteria. This scheme reserves the advantage of using chroma (i.e. avoiding shadow) and that of luma (i.e. stability of segmentation). The foreground mask is refined using an Hidden Markov Model (HMM) for improved performance. \\
\item  \textbf{Luv color space:} Yang and Hsu \cite{CF-90} used the Luv components assuming independence in the computation of covariance matrix required in the MOG model. Then, Yang and Hsu \cite{CF-90} built an hybrid feature space with spatial and color features to obtain a 6-dimensional hybrid feature vector for each pixel. A mean-shift procedure classified each
hybrid feature vector to its corresponding local maximum along the gradient direction. Thus, a set of neighbouring pixels associated with the same local maximum (i.e. mode) is highly similar in this hybrid feature space. Yang and Hsu \cite{CF-90} then assign pixel-wise background likelihood  for each pixel using the MOG likelihood, and further obtain a smoothed version of MOG in terms of spatial and color coherency. \\
\item \textbf{Improved HLS color space:} Setiawan et al. \cite{CF-100} proposed to use the IHLS color space which has the following advantage against the RGB color space. That is to identify shadow region from object by utilizing luminance and saturation-weighted hue information directly, without any calculation of chrominance and luminance. By exploiting this color space in the MOG model, Setiawan et al. \cite{CF-100} obtained good sensitivity to color changes and shadow. \\
\item \textbf{Ohta color space:} The axes of the Ohta space are the three largest eigenvectors of the RGB space, found from the principal components analysis of a large selection of natural images. This color space is a linear transformation of RGB. Using the mean model, Zhang and Xu \cite{FA-10}\cite{FA-11} used the Ohta color space. The three orthogonal color features of Ohta color space are important components for representing color information. Good results in the case of illumination changes and shadows in outdoor scenes are achieved by using only the first two components which are combined with the texture feature (LBP).\\
\item \textbf{YCrCb color space:} YCbCr uses Cartesian coordinates. El Baf et al. \cite{FA-12}\cite{FA-13} used the YCrCb color space combined with the texture feature (LBP) to be robust to illumination changes and shadows. Experimental results in \cite{FA-12}\cite{FA-13} showed that YCrCb color space is more robust in these cases than the Ohta and HSV color spaces.\\
\item \textbf{Lab/Lab2000HL color space:} Lab color space is a color space which indicates proper changes in the direction of human color perception. Its components are the lightness of the color and two color opponent dimensions. Lab2000HL color space, which is an improved version of Lab color space, was introduced and is thought to perform a better modeling of human perception. Particularly, Lab2000HL color space has linear hue band. So, Balcilar et al. \cite{CF-120} investigated the performance of the Lab2000HL color space. The average precision value of Lab2000HL is the greatest in all videos in comparison to all other color spaces. In terms of the computational costs for each color space (YCrCb, Luv, Lab,Lab2000HL), RGB color space leads the lowest. The reason is that it does not require any transformation since the information gathered from the camera sensors is directly in RGB. Lab2000HL color space, on the other hand, has the most computational cost, since a computationally intensive procedure is required to apply first the Lab transformation, and then the computation of transformation value with respect to the transition map using interpolation. The conclusion is that the Lab2000HL color space increases foreground detection rate significantly, in spite of its high computational cost. Balcilar et al. \cite{CF-121} improved the performance obtained by the Lab2000HL color space with a spatial and temporal smoothing scheme.\\
\end{itemize}

Several comparisons on the color spaces can be found in the literature \cite{CF-110}\cite{CF-200} 
\cite{CF-201}\cite{CF-202}\cite{CF-203}\cite{CF-204}. The most complete comparison made by Kristensen et al. \cite{CF-110} compared the following color spaces using the MOG model: RGB, HSI, YCbCr, rgb, $C_1C_2C_3$ and $l_1l_2l_3$. rgb, $C_1C_2C_3$ and $l_1l_2l_3$ are all invariant to changes in brightness, a color space that should decrease the sensitivity to shadows. This investigation showed that the HSI, rgb, $C_1C_2C_3$, $l_1l_2l_3$, and $m_1m_2m_3$ are noisy but less sensitive to shadows than RGB and YCbCr. YCbCr is less noisy than RGB, due to its more independent color channels.  Even though both HSI and $l_1l_2l_3$ are sensitive to changes close to the gray scale, the result is much worse for $l_1l_2l_3$. The light invariant color spaces do not detect as much shadows as the other color spaces, at a cost of missed detection of bright areas. With the $m_1m_2m_3$ color space the segmentation algorithm becomes more of an edge detector, since it is based on two neighbouring pixels. Overall, the most suitable color space for the segmentation algorithm is YCbCr. It is least sensitive to noise, due to numerical stability and more independent color channels. No information is lost when it is transformed from RGB in comparison to other normalized color spaces in which brightness information is usually lost. Finally, YCbCr is affected by shadows and compared to RGB, is too insensitive in some cases but Kristensen et al. \cite{CF-110} presented compensation methods for these two cases. \\

\indent In the case of self-organizing neural network model \cite{CF-203-1}, Lopez et al. \cite{CF-203} compared five different color spaces : RGB, Lab, Luv, HSV and YCrCb. Furthermore, they proposed a color component weighting selection process to take into account the different importance of each component in a color space. A set of 22 different configurations was then evaluated on the I2R dataset \cite{900}. The performance of color spaces has been different in each sequence but YCrCb is the best choice in most cases, and it remains as an interesting option. YCrCb gives the best results when setting the weight to $1$ or close to this value for Y, which indicates that the most important channel for a robust detection is the one that corresponds to lightness, while the remaining channels often add nothing to the result. \\

\indent For moving shadow elimination, Shan et al. \cite{204} determined the optimal color space in which to remove shadow among the following set of
color spaces: RGB, HSV, YCrCb, XYZ, L*a*b*, c1c2c3, $l_1l_2l_3,$ and normalized RGB. Practically, there are three key points with moving shadow detection and elimination \cite{204}: \textbf{(1)}It exists different classes of shadow due to the variety of scenes and different properties of shadow, \textbf{(2)} shadows in each color spaces are differents, which provide different foreground detection results, and \textbf{(3)} there might be not one but different optimal color space following the prominent properties of the shadow. Shan et al. \cite{204} focused  in detecting two kinds of shadows: shadow's illumination and shadow's reflection. Then, shadows are reclassified as invisible and visible. Experimental results \cite{204} show that every space cannot be suitable for all kinds of shadows. HSV, c1c2c3, normalized RGB spaces are appropriate for visible shadow, and YCbCr,
L*a*b* spaces are appropriate for invisible shadow. Even if all types of shadows can be removed in one color space with special methods, different application can selected suitable color space, which can provide twice the result with half effort. \\

\subsection{Features in Designed Shape Color Spaces}
In literature of background modeling and foreground detection, several color space models were designed with a particular shape for the test volume around a background pixel, whose space defines the cluster associated with that pixel. In other words, the test pixels which is inside this volume are associated with the corresponding background. The shape of the volume determines its robustness to highlight and shadows and the different shapes can be classified as follows:
\begin{itemize}
\item \textbf{Cylinder Color Model:} Horprasert et al. \cite{CF-300}\cite{CF-301} separated the brighness and the chromacity components to deal with shading and shadows. Thus a brighness distortion and a color distortion are computed from the RGB components. This color model is called the cylinder color model,and allows to discriminate a RGB color pixel into shadow, highlight, background, and foreground under static background. In the case of the codebook model, Kim et al. \cite{CF-310} used also a cylinder color model which separated the color and brightness component to cope with the problem of illumination changes such as shading and highlights. Thus, a color distortion feature is computed from the current RGB components and the RGB components stored in the codebook. For brightness changes, Kim et al \cite{CF-310} computed a brightness distortion computed from statistics on the intensity to allow the brightness to vary in a certain range that limits the shadow level and highlight level. In an other work, Guo and Hsu \cite{CFS-1}\cite{CFS-2} improved this cylinder color model by using an adaptive threshold instead of a fixed threshold to increase its ability to distinguish highlight and shadow. \\
\item \textbf{Arbitrary Cylinder Color Model:}  However, the CY model is valid only if the spectrum components of the light source change in the same proportion. In fact, it is not true in many practical scenes, where the variations of each spectrum component of the light source would not be in proportion. In these cases, the CY  model is inaccurate and much less efficient. To  solve  this  problem, Zeng and Jia \cite{CF-320}\cite{CF-321} used an Arbitrary CYlinder based color (ACY) model. The ACY model uses cylinders whose axes  needs not going through the origin, so that the CY model is extended to more general cases. Futhermore, the ACY  model reduces the false classification rate of  CY  model  by  more  than  50\% without loss of real-time performance. Practically, the  CY  modelis only a special case of the ACY model. \\
\item \textbf{Hybrid Cone-Cylinder Color Model:} In the cylinder color model, almost every low-intensity test pixel are assigned to the cylinder of a background pixel of very small intensity. Thus, as two similarly grouped pixels increase intensity, they have less chance of being in the same cluster, even though their respective chrominance remains the same. As a more suitable shape, a cone corrects these problems and more precisely covers the color space but a pure conical highlight detector attracts too many pixel values within its space and thus increases the false negative rate. For sensitive detections, the highlight volume should be limited to a cylinder. Thus, Doshi and Trivedi \cite{CF-315}\cite{CF-316} presented a Hybrid Cone-Cylinder Codebook (HC3) model for a 24-7 long-term surveillance system. Experimental results \cite{CF-315} on videos taken by a fixed camera and an omni-directional camera show that the hybrid cone-cylinder color model in HSV outperforms the cylinder color model in RGB. \\
\item \textbf{Ellipsoidal Color Model:} In the CY model, the  two  opposite  margins  of  the  cylinder  cell  are  not  sufficient  to  model  the  background  pixel  with  changed illumination. Indeed, it is not optimal in case of illumination changes to use cylindrical cells of the same size with the same threshold. Furthermore, false negative detections may occur due to the dark  background  pixels  of  which  the  cylinder  codewords  are  located near the origin. The closer to the origin the cylinder  codeword  is,  the  larger  possibility  that  the  corresponding  dark  background pixels may be easily miss-classified as a foreground pixel because of the small illumination changes. To handle this problem, Sun et al. \cite{CF-330} described the color distribution of each background pixel by ellipsoidal shape codewords based on multiple 3D Gaussian distributions.  The variation of every stationary background pixel is limited so that they may be described by only one ellipsoidal cell. Experimental results \cite{CF-330} show that the ellipsoidal color model outperforms the cylinder color model.\\
\item \textbf{Box-based color model:} To reduce computation time, Tu et al. \cite{CF-340} used a box-based color model which
using box-based subspace to represent a codeword. The three edge-lengths values of the box are usually set to the same in the RGB color space, so the box is a cubical box. This model reduces computation complexity. Letting a 24-bit color image sequences, its color mode consisting of
RGB uses 8 bits to represent red, blue and green separately. Thus, the color space can be considered as a large cubical box $256 \times 256$. Then the input pixel value is encoded into the center of cluster when its value lies within a cubical box. So, the pixel value is represented by the cluster center. \\
\item \textbf{Double-Trapezium Cylinder Color Model:}  Huang et al. \cite{CF-350} proposed a Double-Trapezium Cylinder Color Model on YUV, called DTCC-YUV. First, Huang et al. \cite{CF-350} located the shadow detection area in the lower part of the model based on the low luminance of shadows in comparison with the background. Then, a trapezium cylinder is built as the structure of the shadow detection area by using the variable chrominance of shadows. Second, Huang et al. \cite{CF-350} located the highlight detection area in the upper part of the model based on the high luminance of highlights in comparison with the background.  An inverted trapezium cylinder is built as the structure of the highlight detection area based on the variable chrominance of highlights. Finally, the main background area is built with a cylinder structure in the middle part of the model and then DTCC-YUV model is completel built. Experimental results \cite{CF-350} on the PETS 2001 dataset show that the DTCC-YUV model outperforms the cylinder color model \cite{CF-300} and the hybrid cone-cylinder codebook model \cite{CF-315}.  \\
\end{itemize}

\subsection{Designed Illumination Invariant Color Features}
Design of illumination invariant color features was investigated to be robust with illumination changes because the well-known color spaces show limitations in this case. So, Pilet et al. \cite{IF-100} adopted a color illumination ratio as feature to deal with sudden illumination changes. This illumination ratio does not depend of the surface albedo. It depends of the surface orientation and on the illumination environment. Therefore, Pilet et al. \cite{IF-100} used the MOG model with this color illumination ratio instead of RGB with a spatial feature which is the Normalized-Cross Correlation (NCC). Experimental results \cite{IF-100} show that MOG with the color illumination ratio is more robust to illumination changes than the improved MOG of Zivkovic \cite{912} with RGB. In a further work, Wang and Yagi \cite{IF-110} used the same ratio but with a more efficient learning model which improved the robustness against illumination changes. In a multi-backgrounds approach, Takahara et al. \cite{IF-120} used both RGB and HSV features be robust againts various illumination changes while Sajid and Cheung \cite{IF-120-1} used both RGB and YCrCb features. In an other work, Yeh et al. \cite{CF-500} developed a color illumination-invariant. \\

\subsection{Color Filter Array (CFA) Features}
A color image is composed of three channels per pixel which are usually acquired by using three spatially aligned sensors. But, this method present two main disadvantages: 1) it increases the camera  size, and cost, and 2) it requires a complicated pixel registration procedure. So, most digital color cameras employ a single image sensor with a Color Filter Array (CFA) in front. Eeach pixel measures only one color and spatially neighboring  pixels which correspond to different colors are used to estimate unmeasured colors. Among all the CFA patterns, the Bayer CFA pattern is the most widely used pattern \cite{CF-2}. Practically, the Bayer CFA pattern is a $2 \times 2$ pattern which has two green components 
in diagonal locations, and red and blue components in the other locations. An image based on this pattern is called a Bayer-pattern image. The  interpolation process which allow to obtaina a full-color image is called "demosaicing". Suhr et al. \cite{CF-2} used the Bayer pattern instead of RGB in the original MOG \cite{CF-1}. This method present almost the same accuracy as the original MoG using RGB features while maintaining computational resources in terms of time and memory similar to the original MoG using grayscale feature. \\

\subsection{Statistics on Color Features}
\indent Statistics on color feature can be defined and used as features such as:
\begin{itemize}
\item \textbf{Standard Variance:}  Zhong et al. \cite{StF-50} divided each image into patches by representing each image patch as a standard variance feature computed on the intensity. Then, assuming that standard variance feature fits a MOG distribution, Zhong et al. \cite{StF-50} used the MOG model \cite{CF-1}. The advantages of using the standard variance feature as co-occurrence statistics features for dynamic background modeling are the following: \textbf{\textit{1)}} It explicitly considers correlation between pixels in the spatial vicinity. Indeed, an image patch's center pixel in current frame would be a neighbouring pixel in the next frame due to the small movements of objects in dynamic scenes. The center pixel's intensity will change non-periodically. However, the image patch's standard variance feature is unchanged due to the spatial co-occurrence correlations between the center pixel and its neighbouring pixels. \textbf{\textit{2)}} Image noise is largely filtered out with the average filter during the computation of standard variance feature; \textbf{\textit{3)}} The standard variance feature is invariant to mean changes such as identical shifting of intensities. This is very suitable in the case of illumination changes;  \textbf{\textit{4)}} The standard variance feature results in a low dimensional scalar representation of each image patch. This avoids expensive computation during the background modeling phase. Experimental results  \cite{StF-50} on several dynamic backgrounds show that the MOG model is more robust with this feature than RGB features.\\ 
\item \textbf{Color co-occurrences:} Li et al. \cite{900} used in feature selection scheme the color co-occurrences which are more significant features for dynamic background pixels (See Section \ref{subsec:FeaturesChallenges}). In an other work, Adam et al. \cite{OT-1} used co-occurrences of intensities in the spatio-temporal neighborhood of a pixel for dynamic background modeling in an ocean scenes. There are three approaches to used intensities and color co-occurences \cite{OT-1}: \textit{1)} The intensities are collected into a vector. Then similarity between two spatio-
temporal neighborhoods is measured by the norm of the difference vector or by correlation between the two vectors. This approach is too strict for dynamic environments because we cannot expect a good correlation between two space-time volumes of a waving tree or an ocean for example, \textit{2)} Based only on the histogram of the intensities, the two spatio-temporal neighborhoods are compared with their histograms. Instead of the first approach, the order of the samples is completely irrelevant. Furthermore, it ignores important spatio-temporal relationships between pixels, and \textit{3)} a compromise between the two previous approaches which robusly combines each corresponding fragments' similarity scores to obtain an overall similarity measure between the spatio-temporal neighborhoods. \\
\item \textbf{Entropy:} Entropy is a measure of uncertainty. The pixel value changes over time from frame to frame due to the following reasons: (1) changes in the background as in the case of illumination changes and dynamic backgrounds; (2) Motion objects which make the pixel value change from background to object or from object to background. Thus, in  Ma and Zhang \cite{StF-30}, the change of pixel value is considered as the state transition of pixel, e.g. in a 256 level gray image, each pixel has 256 states. Pixel state's change brought by noises would be in a
small range, but those brought by motion will be large. So the diversity of state at each pixel can be used to characterize the intensity of motion at its position. With this assumption, the probability distribution of each pixel's state is observed along temporal axis. Ma and Zhang \cite{StF-30}
used a temporal histogram to present state distribution in a sliding window. The probability density function of pixel's state is
obtained by histogram normalization. Once the histogram is obtained, the corresponding probability density function for each pixel is
computed as follows: \\
\begin{equation}
P_{i,j,q}=\frac{H_{i,j,q}}{N}
\end{equation}
where $H_{i,j,q}$ denotes the spatial-temporal histogram i.e. $w \times w \times L$ pixels are accumulated to form the histogram
of pixel (i,j), $q$ denotes the bins of the histogram, the total number of bins is $Q$, $N$ is the total number of pixels in the histogram and $\sum_{q=1}^{Q} P_{i,j,})=1$. Once the pdf of the pixel is known, the state diversity level of this pixel is computed using entropy definition as follows: \\
\begin{equation}
E_{i,j}=-\sum_{q=1}^{Q} P_{i,j,q} log(P_{i,j,q})
\end{equation}
where $E_{i,j}$ is called the spatial-temporal entropy of pixel (i,j). $E_{i,j}$ is quantized into 256 gray levels to form
an energy image, named as Spatial Temporal Entropy Image (STEI). In STEI, the lighter the pixel is, the
higher its energy is, and the more intensive its motion is. But, STEI cannot differentiate the high entropy caused by motion and spatial structure of the image. Thus, Jing et al. \cite{StF-31} proposed a method based on difference image. The entropy images formed this way is denoted
as Difference-based Spatial Temporal Entropy Image (DSTEI). Experimental results on the PETS 2001 dataset show that DSTEI shows more robustness than STEI in presence of gradual illumination changes and dynamic backgrounds but false detections occur in the case of shadows or sudden illumination changes. In an other way, Chang and Cheng \cite{StF-32} used  the  entropy image with an  adaptive  state-labeling  technique.  Similar  to  
the STEI and DSTEI methods, Chang and Cheng \cite{StF-32}  construct  a  spatio-temporal  sliding  window  for  each  pixel (i,j) to  
acquire the  corresponding spatio-temporal histogram but the  histogram is based on the distribution  of  pixels' state  labels instead on  the  distribution  of  pixels' intensity. A  statelabel  by  a  simple  three-frame  differencing  rule is assigned to each pixel,  and  the  
state label  might be changed  adaptively when the algorithm proceeds over time. This algorithm leads to lower computational complexity compared to to STEI and DSTEI. The detection has a more precise outline and the algorithm is more robust and less sensitive to the changes of parameters. \\
\item \textbf{Illumination-Invariant Structural Complexity (IISC):} Based on the idea that the orthogonal decomposition is effective for separately
handling structural features from the illumination effects in a small local region, Kim and Kim \cite{OT-50} used the singular value decomposition (SVD). Thus, SVD coefficients normalized by the largest singular value provide the illumination-invariant feature space. Due to the fact that pixel values in the small local region are mainly determined by the illumination, its variation is well revealed in the largest singular value in the SVD-based scheme. Then, Kim and Kim \cite{OT-50} defined the unit brightness level in which all the singular values are divided by the
largest one. The illumination-invariant structural information are revealed by such normalized singular values. Therefore, the sum of those values is employed as feature called Illumination-Invariant Structural Complexity (IISC). The background model used the traditional on-line updating scheme, i.e the running average. Experimental results \cite{OT-50} on the PETS2001 dataset and OTCBVS dataset show that IISC has the ability to discriminate structural changes due to moving objects from those due to illumination effects. \\
\end{itemize}

\subsection{Discussion on Intensity and Color Features}
According to the literature, it appears that \textbf{1)} Spectral-360 \cite{CDFeatureI-500} based on the dichromatic reflection model is the best reflectance approach for intensity features,  \textbf{2)} YCrCb color space seems to be the most suitable space for category of color features in well-known color spaces as demonstrated in case of the MOG  model and the self-organizing neural network model \cite{CF-203} even if it is affected by shadows and it is too insensitive in some cases. But, compensation methods \cite{CF-110} can be employed for these two cases, \textbf{3)} for features in designed shape color spaces, the double-trapezium cylinder color model \cite{CF-350} in YUV color space appears as the most suitable shape model for the codebook model,  \textbf{4)} color illumination ratio \cite{IF-100} provides a robust ilumination-invariant color features, and \textbf{5)} When a single image sensor with a Color Filter Array (CFA) is used , Bayer CFA pattern \cite{CF-2} allows the MOG model to keep the same accuracy as the RGB features while maintaining computational resources in terms of time and memory similar to the grayscale feature. Furthermore,  statistics on color features are useful to be more robust in presence of dynamic backgrounds and illumination changes. \\

\indent All the intensity and color features can be combined with edge, texture, motion or stereo features to take advantage of these features. In literature, edge features are one the first features used in addition of color features and are detailed in the following section. \\

\section{Edge Features}
\label{sec:EdgeFeatures}

\subsection{Intensity Edge Features}
Edge based on intensity features give \textit{\textbf{spatial informations}}, and are computed using a gradient approach such as Canny, Sobel  \cite{EF-0}\cite{EF-60}\cite{FA-30}\cite{FS-20}\cite{900}\cite{MulF-30} or Prewitt \cite{EF-2}\cite{EnF-5} edge detector. The gradients can be calculated from the gray level image or in each components of a color space. The edge features are generally used in addition to intensity or color features  \cite{EF-0}\cite{EF-1}\cite{EF-2} or alone \cite{EF-100} \cite{EF-130}\cite{EF-131}\cite{EF-132}\cite{EF-133}\cite{EF-134}\cite{EF-135}\cite{EF-136}\cite{EF-140} as follows: \\ 

\begin{itemize}
\item \textbf{In addition with intensity or color features:} First, Jabri et al. \cite{EF-0} used in addition to the intensity features the intensity gradient obtained by the Sobel edge detector. Large changes in either intensity or in edges are fused. However, the involvement of the intensity model retains the sensitivity to sudden changes in illumination. In Javed et al. \cite{EF-1}, the edge and color information obtained from pixel level is integrated at the region level. The basic idea is that any foreground region that corresponds to an actual object will have high values of gradient based background difference at its boundaries. It requires significant changes in both the intensity and intensity gradient. The use of a gradient feature removes many false alarms due to small illumination changes. However, intensity gradients arising from large illumination changes can still generate false detections. As edge detection consists of first filtering the image using a suitable approximation of derivatives followed by a thresholding, it gives an image where the pixel values come from a binary distribution. This image is difficult to model using Gaussian mixtures. Skipping the thresholding gives pixel values that come from a continuous distribution. However, since a majority of the picture usually contains no edges such a distribution will be extremely skewed, and thus still difficult to model using Gaussian mixtures. To solve this problem, Lindstr\"{o}m et al. \cite{EF-2} proposed to use a Prewitt edge detector without the thresholding independently to each color component followed by a log-transformation, to reduce skewness, gives a color edge image with pixel values that can be modelled using Gaussian mixtures. Experimental results \cite{EF-2} shows better performance against illumination changes for the log-transformed detection using the Prewitt edge detector. In an other work, Kim et al. \cite{OT-40} used edge and non-edge (texture) features in a hybrid background model
to generate the background model. Thus, theses features are encoded into a coding scheme called Local Hybrid Pattern (LHP). LHP selectively models edges and non-edges features of each pixel. Then, each pixel is modeled with an adaptive code dictionary to take into account the background dynamism. In the background maintenance, stable codes are added in the model while unstable ones are discarded. The incoming codes that deviate from the dictionary are classified as edge or inner region. Experimental results  \cite{OT-40} on the ChangeDetection.net dataset show that this Adaptive Dictionary Model (ADM) with LHP features outperforms the original MOG \cite{CF-1}, the original LBP \cite{TF-11} and SALBP \cite{TF-30}. ADM achieves similar results than ViBe \cite{FS-51} and PBAS \cite{231}. The reader can see how the color features and edge features are fused in these different approaches in Section \ref{sec:MultipleFeatures}. \\
\item \textbf{Edges alone:} First, Kim and Hwang \cite{EF-120} proposed to use only edges to model the background, and thus this approach used binary information as existence of edge on each pixel. But, regions in consecutive frames may not have exactly the same edge position, and have shape and length changes due to presence of noise. This strategy may generate many false alarms in the foreground mask due to edge distortion from consecutive frames. To solve the edge-distortion problem, edge-segment-based methods have emerged to take advantage of the edge existence and its shape information \cite{EF-121}. An edge-segment approach consists of the concatenation of adjacent edges, and it inherits the problems of edges: shape and position changes. Thus, basic comparison of edge-segments produces similar results as edge-pixel-based approaches. To solve this problem, statistical edge-segment-based methods extract movement of edge-segments including edge distortion \cite{EF-130}\cite{EF-131}\cite{EF-132}\cite{EF-133}\cite{EF-134}\cite{EF-135}\cite{EF-136}. Thus, these methods solve the edge-variation
problem by accumulating edge existence from a training sequence \cite{OT-40}. Pratically, each accumulated region represents an edge-
segment distribution. Each region refines their statisti-cal properties after each frame to provide a stable background model. Since edge-based and edge-segment-based methods detect foreground as edges, these methods depend on a post-processing to extract the regions defined by the detected edges. Moreover, these methods have problems updating their background model to adapt the background. \\
\end{itemize}

\subsection{Subpixel Edge-Maps Features}
A key limitation of intensity gradient features is that they do not take spatial interactions into account. Alternatively, edge maps tag those background pixels which maximize local gradient in a neighborhood of pixels. This tagging increases selectivity which in turn reduces both the number of pixels which would have been discarded from the background model and the pixels would have been erroneously labeled as foreground \cite{EF-50}\cite{EF-51}. Discretization errors in pixel-based edge maps generate unnecessary broad background models: a background edge halfway between the pixels will require both pixels modeled as background, thus unnecessarily "blurring" the background model, which in turn reduces sensitivity to detecting regions. Instead of intensity gradient features, Jain et al. \cite{EF-100} proposed to use subpixel edge-maps by modeling the position and the orientation of subpixel edges which disambiguates between edges of the same orientation but at different positions and vice versa. Subpixel edge-maps has high precision and accuracy with invariance to illumination changes and suitable for small translations. An other advantage is that it requires fewer frames to build the background model even in case of slow moving objects and bootstrapping when edges from a region can share the same pixel as well as the same orientation. \\

\subsection{Statistics on Gradient Features}
Statistics on gradient features are also used as follows:
\begin{itemize}
\item \textbf{Gradient deviations:} Kamkar-Parsi \cite{MulF-7} proposed to model the probability of appearance and disappearance of edges due to  
moving objects in the scene. This probability model is similar to the order consistency criteria described in \cite{CDFeatureI-1} applied to gradient deviations intead of intensity values. \\
\item \textbf{Projection Gradient Statistics:} Zhang \cite{StF-60} used projection gradient statistics as features. In the projection statistics curves, the largest change of the statistical value is caused by the moving objects. Thus, the gradient of the curve can reflect the statistical curve change clearly but also the position of the moving objects. First, the background is identified by projection statistics and projection gradient statistics of sub-images. Second, the background is reconstructed according to the results of each sub-image. Experimental results \cite{StF-60} show that this approach has a good anti-interference to low intensity vehicles in highway traffic scenes, and the processing time is less as well. \\
\end{itemize}

\subsection{Discussion on Edge Features}
The influence of the edge detector (Canny, Sobel, Prewitt) on robustness have not be studied in literature. Edges based approaches used in addition with intensity or color features are the most investigated approaches and allow to combine the advantage of the two features. Forthe approahces based on edges alone, only two main works emerged that are the works of based on edge segments\cite{EF-130}\cite{EF-131}\cite{EF-132}\cite{EF-133}\cite{EF-134}\cite{EF-135}\cite{EF-136} and the work based on subpixel edge \cite{EF-100}. These approaches appear to be relevant too and merit to be more investigated. Statistics on gradient features are useful to be more robust in presence of low intensities. \\

\section{Texture Features}
\label{sec:TextureFeatures}
Texture features give \textit{\textbf{spatial informations}}, and are the most investigated features in the field as can be seen in Table \ref{Overview3}. The Local Binary Pattern (LBP) \cite{TF-10}, and the Local Ternary Pattern (LTP) \cite{TF-70} are the most used with numerous variants. Furthermore, some improvements are developed in literature or can be developed to allow texture features to integrate both:
\begin{enumerate}
\item \textbf{Spectral and temporal informations} such as Spatial-Color Binary Patterns (SCBP) \cite{TF-24}, Opponent Color LBP (OCLBP) \cite{TF-25}, Uniform LBP (ULBP) \cite{TF-26}, Scene Adaptive LBP (SALBP) \cite{TF-30}, Intensity LBP (iLBP)\cite{TF-36}, eXtended Center-Symmetric \cite{TF-63}  and Multi-Channel Scale Invariant LTP (MC-SILTP) \cite{TF-73}.\\
\item \textbf{Spatial and temporal informations} such as Spatial-Temporal LBP \cite{TF-12}\cite{TF-13}, Space-Time Center-Symmetric LBP (ST-CS-LBP) \cite{TF-34}, Spatial Center-Symmetric LBP (SCS-LBP) \cite{TF-21}, Motion Vectors Local Binary Patterns (MV-LBP) \cite{TF-29}\cite{TF-61}\cite{TF-62}, Stereo LBP based on Appearance and Motion (SLBP-AM) \cite{TF-31}, Space-Time Center-Symmetric Local Binary Patterns (ST-CS-LBP) \cite{TF-34}, Center Symmetric Spatio-temporal LTP (CS-ST-LTP) \cite{TF-74}, Spatio Temporal Scale Invariant LTP (ST-SILTP) \cite{TF-76}, Texture Pattern Flow (TPF) \cite{TF-201}\cite{TF-202} and dynamic texture patterns \cite{TF-31}\cite{OT-30}\cite{OT-31}\cite{TF-125-1}\cite{OT-33}\cite{OT-34}. \\
\item \textbf{Spectral, spatial and temporal informations:} There are no improvements in this category. Thus, investigation can be made in this direction. \\
\end{enumerate}
Other approaches \cite{TF-20}\cite{TF-22}\cite{TF-32}\cite{TF-33}\cite{TF-23}\cite{TF-35}\cite{TF-90}\cite{TF-110} reduce memories and computation cost of texture features. Moreover, other improvements was made by the use of statistical or fuzzy concepts. We classified them as follows : 
\begin{enumerate}
\item \textbf{Crisp features:} such as local patterns \cite{TF-10}\cite{TF-70}\cite{TF-90}\cite{TF-115}\cite{TF-110}\cite{TF-117}\cite{TF-125}\cite{TF-127}\cite{TF-128} and spatio-temporal patterns \cite{TF-170}\cite{TF-174}\cite{TF-175}\cite{TF-180}\cite{TF-181}. \\
\item \textbf{Statistical texture features:} such as peripheral patterns \cite{TF-1}\cite{TF-5}\cite{TF-6} and reach patterns \cite{TF-130}\cite{TF-131}\cite{TF-132}\cite{TF-135}\cite{TF-135-1}\cite{TF-136}\cite{TF-140}.\\
\item \textbf{Fuzzy texture features:} Local Fuzzy Pattern (LFP) \cite{TF-120}\cite{TF-121} and Fuzzy Statistical Texture (FST)\cite{FF-10}\cite{FA-40}. \\
\end{enumerate}
In the following sub-sections, we brieftly describe each texture feature in each category and the reader can see how the texture features are fused with other features in Section \ref{sec:MultipleFeatures}.

\subsection{Local Patterns}
\label{subsec:LocalPatterns}

\subsubsection{Local Binary Pattern (LBP)}  
Heikkila et al. \cite{TF-10} proposed to use a texture feature called Local Binary Pattern (LBP) computed from the intensity feature. Based on this texture feature, a block-wise LBP histogram based approach (LBP-B) \cite{TF-10} and pixel-wise LBP histogram based one (LBP-P) \cite{TF-11} were developed. These methods can deal with gradual and sudden illumination changes due to the robustness of the texture feature to illumination variations. But these methods only use one learning rate that is a trade-off between different rates of change in background. When a high learning rate is used, the model updates quickly and slow-moving objects are incorporated into the background model which causes in a high false negative rate. When a low learning rate is used, these methods cannot handle sudden changes in background which causes in a high false positive rate. To address this problem, Goyal and Singhai \cite{TF-65} proposed a LBP texture-based algorithm which uses adaptive learning rate to deal with different rates of change in background. In an other approach, intead of computing LBP from the intensities, Satpathy et al. \cite{TF-14} proposed to apply LBP on the background and current edge images obtained by a Difference of Gaussians (DoG) edge detectors. Li et al. \cite{TF-15} combined the LBP difference between the background and the current images, the single Gaussian \cite{200} and the codebook \cite{302} for foreground detection to combine complementary advantages of each method. In a further work, Wu et al. \cite{TF-16}\cite{TF-17} proposed a layered background modeling. First, every block on the first layer is modeled via texture based on local binary pattern (LBP) operators. Then, the modeling granularity is deflated onto the second layer to model via codebook. Layered match is done from top down when a new video frame enters. Experimental results \cite{TF-16}\cite{TF-17} show that this approach efficiently avoids the false negative detection rate in the pixel-based background modeling when the object color is similar to the background, and also stops the false positives occurring on the contour areas of the moving objects due to the block model. In an other work, Zhong et al.\cite{TF-18}  used intensity, LBP code and the norm of the first order derivative of the intensity with respect to $x$ and $y$ in a covariance based model to deal with dynamic backgrounds. In a further work, Zhong et al. \cite{TF-19} developed  a background subtraction algorithm, which takes both texture and motion information into account. Texture information is represented by local binary pattern (LBP), which is tolerant of illumination changes and is computational simplicity. Assuming that there is significant structure in the correlations between observations across time, an operator to extract motion information is used. Then, each pixel is modeled as a group of texture pattern histograms and motion pattern histograms respectively. Finally, the texture pattern-based and motion pattern-based background model are combined using a weighted rule. A mixture factor $\gamma$ is used to control the influence of the texture pattern-based background model and the motion pattern-based background model. Practically, it was taken to $0.5$. This combination is more robust to dynamic backgrounds. These methods applied the original LBP in different ways compared to Heikkila et al. \cite{TF-10}. However, there are numerous variants of LBP and some of them have been developped for background/foreground separation. They can be classified as follows:  \\
\begin{itemize}
\item \textbf{Spatial-Temporal LBP (STLBP):} A spatial background model, like  LBP \cite{TF-10} (and RRF \cite{TF-152}), has the following problem. Their method uses spatial invariant features to monotonic changes in a local area. When pixel values change in a part of the local area, spatial invariant features are no longer good. Such a situation often occurs especially in outdoor scene changing of weather conditions. To address this problem, Shimada and Taniguchi \cite{TF-12} proposed an invariant feature using both spatial invariance and temporal invariance called Spatial-Temporal LBP (STLBP) suitable for outdoor scene in which the illumination condition can change gradually. In a similar way, Zhang et al. \cite{TF-13} extended the ordinary LBP from spatial domain to spatio-temporal domain, and proposed a new online dynamic texture extraction operator, named spatio-temporal local binary patterns (STLBP). Then, Zhang et al. \cite{TF-13} developed a dynamic background modeling and subtraction based on STLBP histograms which combine spatial texture and temporal motion information together. However, the computational load was increased and comparative results with the original LBP \cite{TF-10} were not presented. \\
\item \textbf{$\epsilon$-LBP:} The main limitation of LBP is that both memories and computation costs increase greatly with the increasing of the images resolution. To solve it, Wang and Pan \cite{TF-20} proposed a fast background subtraction method based on the novel LBP called $\epsilon$-LBP. Furthermore, it overcomes two drawbacks brought by the LBP operator, i.e. the neighboring pixels are conditional independent under the center pixel, and it is weakly to measure the difference between the center pixel and its neighborhood. Compared with the LBP, $\epsilon$-LBP improves greatly memories and computation efficiency by a simple measurement, which is linearly proportional of the images resolution. But, $\epsilon$-LBP  needs a threshold which is empirically selected as a global constant. Thus, this method only performs well when the illumination variation is global. To address this problem, Wang et al. \cite{TF-22} improved th $\epsilon$-LBP by adding local adaptive property. The threshold is adaptively selected for each pair of two neighboring pixels. With two evaluation criterions, i.e. the description stability and the discriminative ability, a simple yet effective approach is presented to adaptively estimate the threshold by classifying all the pixels into two groups, i.e. the edge pixels and the texture pixels. In background modeling procedure, a naive Bayesian technique is adopted to effectively model the probability distribution of local patterns in the pixel level. The utilization of single $\epsilon$-LBP (pixel level) improves the robustness to the illumination variation and reduces the computation cost compared with LBP. \\
\item \textbf{Center Symmetric Local Binary Patterns (CS-LBP):}  Tan et al. \cite{TF-33} proposed to use the Center Symmetric Local Binary Patterns (CS-LBP) \cite{F-1}, which effectively decreases the influence of noise, illumination changes and shadows. This method firstly computes the CSLBP feature images, then it calculates integral histogram as basic feature of background model, and finally establishes a background model containing $K$ models. Finally, it classifies pixels as background or foreground and updates the background model by comparing the similarity of histogram within $L$ square neighborhood. This methods increases processing speed effectively by the use of one integral histogram and two integral histograms, respectively. To decrease the computation time, Wu and Zhu \cite{TF-32} proposed to compute the CS-LBP eigenvalue in the current image and to subtract it with the eigenvalue of the same pixel in the background image. The difference is then thresholded to classify pixel as background or foreground. This method could meet the requirements of real-time detection. In a further work, Zhang et al. \cite{TF-64} used an adaptive strategy based on CS-LBP. Then, foreground detection is used for detecting moving object by using a confidence factor to determine whether the current pixel is a background or foreground pixel. The confidence factor of the current pixel is computed in terms of the difference values of its neighbourhood pixels. Experimental results made with PETS 2009, BMC 2012 \cite{903} and SABS \cite{902} datasets demonstrate that this approach can robustly detect moving object under various scenes. An other approach proposed by Li et al. \cite{TF-34} consists in a space-time symmetrical ST-CS-LBP operator which integrates time prediction and texture information. A ST-CS-LBP histogram is built and merges the advantages of time-domain statistics and spatial distribution. Results show more robustness to long and short luminance changes than the CSLBP. \\
In an other way, Xue et al. \cite{TF-21} proposed to use a Spatial Center-Symmetric Local Binary Pattern (SCS-LBP). This operator not only has the property of illumination invariance, but also produces short histograms and be more robust to noise. So, Xue et al. \cite{TF-21} extended the CS-LBP operator from spatial domain to spatial-temporal domain and proposed a texture operator named SCS-LBP which extracts spatial and temporal information simultaneously. Then, combining the SCS-LBP operator with an improved temporal information estimation scheme, Xue et al. \cite{TF-21} obtained a background modeling approach which reach high accurate detection in dynamic scenes while reducing the computational complexity compared to the LBP based method. The LBP operator produces long feature set since it only adopts the first-order gradient information between center pixel and its neighbors. Xue et al. \cite{TF-23} developed a second-order center-symmetric local derivative pattern (CS-LDP) operator which extracts more detail local information than CS-LBP. Then, Xue et al. \cite{TF-23} concatenated the CS-LBP histogram and CS-LDP histogram to get a new hybrid feature. This LBP pattern is called Hybrid Center-Symmetric LBP (HCS-LBP) Experiments on challenging sequences indicate that this method can produce comparable results while using less computation time (25\%) compared to the LBP based method. \\
\item \textbf{Spatial-Color Binary Patterns (SCBP):} Zhou et al. \cite{TF-24}  proposed an spatial-color feature extraction operator named spatial-color binary patterns (SCBP). It extracted spatial texture and color information. In addition, a refine module is designed to refine the contour of moving objects. For each pixel, first, a histogram of SCBP is extracted from the circular region, and then a model consist of several histograms is built. For a new observed frame, each pixel is labeled either background or foreground according to the matching degree between its SCBP histogram and its model, then the label is refined and finally the model of this pixel is updated. \\
\item \textbf{Opponent Color Local Binary Pattern (OC-LBP):} Lee et al. \cite{TF-25} developed a novel training process for classifiers which use block based Opponent Color Local Binary Pattern (OC-LBP). So, the pixel based OC-LBP \cite{TF-65} was extended to to block-level.  \\
\item \textbf{Double Local Binary Pattern (DLBP):} Xu et al. \cite{TF-60} developed a Double Local Binary Pattern (DLBP). \\
\item \textbf{Uniform Local Binary Pattern (ULBP):}  Yuan et al. \cite{TF-26} proposed in a $K$ histograms model a combination of color feature, i.e hue, and an uniform local binary pattern (ULBP) texture to be robust to shadows. The results from each features are combined with a weighted average. ULBP with the hue outperforms LBP \cite{TF-10} and DLBP \cite{TF-60}.  \\
\item \textbf{Extended and Rotation Invariant LBP (Ext-LBP):} Yue et al. \cite{TF-27} developed an extended LBP (Ext-LBP) and a Rotation Invariant LBP (RI-LBP). Ext-LBP is obtained by expanding the adjacent area of original LBP. Combined RI-LBP with MOG effectively removes shadow influences. \\
\item \textbf{Larger neighborhood LBP (LN-LBP):} Kertesz  \cite{TF-28} used a larger neighborhood than the original LBP (LN-LBP). It is calculated inside a $5\times5$ neighborhood, therefore, it is not defined two pixels wide on the image borders. The algorithm specified a correction value to the LN-LBP calculation in order to handle the flat color areas where the color values almost do not change. Furthermore, Kertesz \cite{TF-28} used a Markov Random Field (MRF) as a higher level classification of the LN-LBP histograms into foreground and background.  LN-LBP with MRF is quite invariant for the resolution and provides better performance at higher resolutions than the original LBP \cite{TF-11}. \\
\item \textbf{Motion Vectors Local Binary Patterns (MV-LBP):} Yang et al. \cite{TF-29}, Wang et al. \cite{TF-61} and Wang et al. \cite{TF-62} proposed a moving object detection method towards H.264 compressed surveillance videos. First, the motion vectors (MV) are accumulated and filtered to achieve reliable motion information. Second, considering the spatial and temporal correlations among adjacent
blocks, spatio-temporal Local Binary Pattern (LBP) features of MVs are extracted to obtain coarse and initial object regions. Finally, a coarse-to-fine segmentation algorithm of boundary modification is conducted based on the DCT coefficients. \\
\item \textbf{Scene Adaptive Local Binary Pattern (SALBP) :} Noh and Jeon \cite{TF-30} developed a texture operator namely, Scene Adaptive Local Binary Pattern (SALBP) that provides more consistent and accurate texture-code generation by applying scene adaptive multiple thresholds. A background subtraction framework employing diverse cues (pixel texture, pixel color and region appearance) is presented. The SALBP information of the scene is clustered by the conventional codebook model \cite{302}, and utilized to detect initial foreground regions. Background statistics of the color cues are also modeled by the codebook model and employed to refine the texture-based detection results by integrating color and texture characteristics. \\
\item \textbf{Stereo Local Binary Pattern based on Appearance and Motion (SLBP-AM) :} Yin et al. \cite{TF-31} proposed a Stereo Local Binary Pattern based on Appearance and Motion (SLBP-AM) descriptor. The motion of pixels is represented as dynamic texture in ellipsoidal domain. Then, Yin et al. \cite{TF-31} combined texture histograms in the XY, XT, YT planes in the ellipsoid. SLBP-AM is more robust to slight disturbance, but also adapts quickly to the large-scale and sudden changes. \\
\item \textbf{Window-based LBP (WB-LBP):} The histogram computation and construction of LBP is a very time-consuming and complex process. Kumar et al. \cite{TF-35} proposed to reduce the complexity to a large extent by using a window-based LBP (WB-LBP) subtraction method. Moreover, the efficacy of the WB-LBP in terms of correct classification is quite satisfactory as compared to the other LBP-based methods. \\
\item \textbf{Intensity Local Binary Patterns (iLBP):} The original LBP \cite{TF-10} has a main drawback. Indeed, it ignores the intensity information when comparing LBP descriptors. Because of this, there could be a paradoxical situation which generates wrong pixel comparison result when intensity values of pixels differ drastically, but their LBP descriptors are identical. To solve it, Vishnyakov et al. \cite{TF-36} proposed a intensity Local Binary Patterns (iLBP) descriptor and built a fast background model on its basis. This feature allows stabilizing the value of the descriptor and constructing a background model that is robust to lighting conditions changes in the scene and is applicable for the real time multi-camera setup. In a further work, Vishnyakov et al. \cite{TF-36-1} used iLBP in a background model based onregression diffusion maps. This approach allows objects that move with different speed or even stop for a short while to be uniformly detected.\\
\item \textbf{BackGround Local Binary Pattern (BGLBP):}  Davarpanah et al. \cite{TF-66} developed a BackGround LBP (BGLBP) which has been designed to inherit the positive properties of Direction Local Binary Pattern (D-LBP), CS-LBP \cite{TF-33}, ULBP \cite{TF-26}, and RI-LBP \cite{TF-27}. Experimentals results on the I2R dataset \cite{900} show that BGLBP outperforms several variants of LBP such as the original LBP, ULBP and RI-LBP in presence of illumination changes and dynamic backgrounds. \\
\item \textbf{eXtended Center-Symmetric Local Binary Pattern (XCS-LBP):} Silva et al. \cite{TF-63} presented an extension of CS-LBP called XCS-LBP (eXtended CS-LBP) by comparing the gray values of pairs of centersymmetric pixels so that the produced histogram are short as well, but considering the central pixel also. This combination makes the resulting descriptor less sensitive to noise. Experimental results on the BMC dataset \cite{903} show that XCS-LBP outperforms LBP, CS-LBP and CS-LDP. For computation time, XCS-LBP has slightly better time performance than both CS-LBP and CS-LDP. \\
\item \textbf{Local SVD Binary Pattern (LSVD-BP):} Guo et al. \cite{TF-67} developed the LSBP feature descriptor which has the ability to gain the potential structures of local regions. LSBP also inhibits the effect of illumination changes especially cast shadows and noise. Experimental results on the ChangeDetection.net dataset \cite{901} show that LSBP allows robustness in the "Shadow" and "Thermal" categories while LSBP seems to perform at a level comparable to PBAS \cite{231} the "Dynamic Backgrounds" and "Baseline". \\
\end{itemize}
In summary, three main properties are required for a designed LBP version for the application of background modeling and foreground detection \cite{TF-66}: \textit{1)} It should be fast to compute, \textit{2)} the number of bins in the plotted histogram should be the least, and \textit{3)} It should be computed based on whole pixels' values belonging to each block. Thus, the original LBP \cite{TF-10}\cite{TF-11} and the LBP based on gradient \cite{TF-14} are not optimal. Multi Block LBP calculates a LBP value for each block instead of each pixel separately. ULBP \cite{TF-26} is not rotation invariant. Finally, there are many bins in the CS-LBP model. Regarding all of these limitations, BGLBP and XCS-LBP seem to be the best LBP variants for background modeling and foreground detection. \\

\subsubsection{Local Ternary Pattern (LTP)}
\label{subsubsec:LocalTernaryPattern}

\begin{itemize}
\item \textbf{Local Ternary Pattern (LTP):} First, Tan and Triggs  \cite{TF-70-1} proposed to include an additional buffering
state to solve the instability problem of LBP, and introduced the local ternary pattern (LTP) as a robust extension of LBP. Liao et al. \cite{TF-70} proposed to use the Local Ternary Pattern (LTP) operator \cite{TF-70-1} for background/foreground separation. Pratically, LTP is more robust by introducing a small tolerative range. However, the descriptor is extended from LBP by simply adding a small offset value for comparison, which is not invariant under scale transform of intensity values by a multiplying constant. For example, LTP descriptor can not keep its invariance against scale transform when all local pixel values are multiplied by $2$. \\
\item \textbf{Scale Invariant Local Ternary Pattern (SILTP):} For background modeling and foreground detection, the intensity scale invariant property of a local comparison operator is very important, because illumination variations, either global or local, often cause sudden changes of gray scale intensities of neighboring pixels simultaneously, which would approximately be a scale transform with a constant factor. Therefore, Liao et al. \cite{TF-70} proposed to extend LTP to Scale Invariant Local Ternary Pattern (SILTP). Furthermore, Liao et al. \cite{TF-70} improved Heikkila et al's region histogram based method by modeling pixel process with a single local pattern instead. However, local patterns are not ordinal numerical values, thus can not be modeled directly into traditional density functions. Therefore,  Liao et al. \cite{TF-70} developed a Pattern Kernel Density Estimation (PKDE) technique to effectively model probability distributions of such patterns for handling complex dynamic backgrounds and a multiscale fusion scheme to consider the spatial scale information. The SILTP operator was presented by $8$ bits \cite{TF-70}. This encoding strategy only used $4$ pixels of its 8 neighborhoods which might result in loss of the texture information. If all $8$ neighboring pixels were used, the length of encoding was 16 bits which led to computational complexity increasing. \\
\item \textbf{Scale Invariant Center-symmetric Local Ternary Pattern (SCS-LTP):} Zhang et al. \cite{TF-71} proposed a texture operator named Scale Invariant Center-symmetric Local Ternary Pattern (SCS-LTP), and a corresponding Pattern Adaptive Kernel Density Estimation technique for its probability estimation. Zhang et al. \cite{TF-71} used a simplified Gaussian Mixture Models for intensity feature. Then, the results from texture and intensity are combined in a multi-scale fusion scheme based on the basic product formulation of the likelihoods. Zhang et al. \cite{TF-72} used a similar scheme but based on Gaussian Mixture Models for both texture (SCS-LTP) and color features (Normalized RGB). The results from texture and color  are combined in a multi-scale fusion scheme based on the weighted average. The weights are a function of a parameter balancing texture
distance and color distance which is empirically set to $0.7$. There are two advantages of the SCLTP operator in foreground/background separation. The first one is that the SCLTP operator is robust not only to noise but also to illumination by introducing a scale factor like SILTP. The second one is that the SCLTP operator can represent more texture information with less bits. The SCLTP operator can use $8$ bits to express a pixel with all its $8$ neighboring pixels because of only comparing center-symmetric pairs of pixels. Experimental results  \cite{TF-71}\cite{TF-72} on several complex real world videos with illumination variation, soft shadows and dynamic backgrounds (I2R dataset \cite{900}) show that SCS-LTP outperfoms sligthly SILTP \cite{TF-70}. \\
\item \textbf{Multi-Channel Scale Invariant Local Ternary Pattern (MC-SILTP):} Ma and Sang \cite{TF-73} proposed to extends the SILTP to feature space and to operate on the three channels of RGB images rather than the only channel of gray images to get the texture pattern. This texture descriptor  is called Multi-Channel Scale Invariant Local Ternary Pattern (MC-SILTP). \\
\item \textbf{Center Symmetric Spatio-temporal Local Ternary Pattern (CS-ST-LTP):} Xu \cite{TF-74} and Lin et al. \cite{TF-75} decomposed the scene background into a number of regular cells, within which a batch of video bricks (e.g.pixels) are extracted as observations. In order to better represent video bricks and enhance the robustness to illumination variations, Xu \cite{TF-74} proposed a brick-based descriptor, namely Center Symmetric Spatio-temporal Local Ternary Pattern (CS-ST-LTP), instead of using pixel intensity. CS-ST-LTP is inspired by the 2D local pattern descriptor and adapted to characterize video brick.. \\
\item \textbf{Spatio Temporal Scale Invariant Local Ternary Pattern (ST-SILTP):} Considering the temporal persistence of texture sequences, Ji and Wang \cite{TF-76} extended the SILTP \cite{TF-70} to the spatiotemporal domain called ST-SILTP. Second, Ji and Wang \cite{TF-76} presented an adaptive fusion approach of color and texture to compensate for their respective defects. Furthermore, since a pixel of foreground or not depends on not only itself but also its neighborhood, a lateral inhibition filter model incorporated the neighborhood information into calculating the pixel's confidence score. Practically, a pixel is classified as background or foreground bt using its probability computed from the similarities in each feature, and from
confidence of color and texture components. For color or texture feature of a background mode, if its corresponding confidence is a larger value, it means that it plays a more important role in making the decision about the label of the given pixel, i.e., foreground or background. In the computation of updating the background model, the weight and confidences of color and text features are constantly updated in the online learning way. Thus, in estimating the probability of a pixel as background, the contribution of color and texture components can be adaptively adjusted and both kinds of information are fully fused together to segment foreground object. Experimental results \cite{TF-76} on I2R dataset \cite{900} show that ST-SILTP gives silhouettes with less holes than the original SILTP \cite{TF-70}.\\
\item \textbf{Center-Symmetric Scale Invariant Local Ternary Pattern (CS-SILTP):} Wu et al. \cite{TF-77} extended the SILTP descriptor by introducing a the Center-Symmetric Scale Invariant Local Ternary Pattern (CS-SILTP)
descriptor, by exploring spatial and temporal relationships of neighborhood. \\
\end{itemize}

\subsubsection{Local States Pattern (LSP)} 
\label{subsubsec:LocalStatePattern}
Yuk et al. \cite{TF-90} introduced Scale Invariant Local States (SILS) as texture features for modeling a background pixel, and a
pattern-less probabilistic measurement (PLPM) which is derived to estimate the probability of a pixel being background from its SILS. An adaptive background modeling framework was also proposed for learning and representing a multi-modal background model. Practically, considering $N$ neighbors in the texture pattern and $K$ background models, the memory space and operations required for the SILS based method  are only $O(3NK)$ comparing to SILTP based method \cite{TF-70} which requires $O(3NK)$. Experimental results \cite{TF-90} on the I2R dataset \cite{900} show that SILS based method runs nearly $3$ times faster than SILTP based method \cite{TF-70} with the same accuarcy.

\subsubsection{Local Difference Pattern (LDP)} 
\label{subsubsec:LocalDifferencePattern}
Yoshinaga et al. \cite{TF-100} proposed a probabilistic background model which combined pixel-based multi-modal model with color features and spatial-based uni-modal model with texture features by considering the illumination fluctuation in localized regions. Then, Yoshinaga et al. \cite{TF-100} used several pairs of a focused pixel and its peripheral pixels, i.e., its surrounding pixels, in a localized region, and modeled the distribution of the difference between pixel values of each pair with a mixture of Gaussians. These pixel value differences in the localized region is called Local Difference Pattern (LDP). LDP presents several advantages: \textbf{(1)} There are little changes in a LDP in presence of sudden illumination changes because the pixel values in a localized region similarly increase and decrease, and \textbf{(2)} LDP can also deal with periodic changes of pixel values because MOG represents multiple hypotheses of the background. Thus, LDP used both properties of pixel-based and spatial-based features, without decreasing the accuracy. In further works, Yoshinaga et al. \cite{TF-101}\cite{TF-102} called this feature Statistical Local Difference Pattern (SLDP). Experimental results on the BMC dataset \cite{903} show that SLDP outperforms adaptive RRF \cite{TF-152} in presence of illumination changes and dynamic backgrounds.

\subsubsection{Local Self Similarity (LSS)} 
\label{subsubsec:LocalSelfSimilarity}
LBP \cite{TF-10} and SPF \cite{TF-190} performed reasonably well, but the features that are used can only capture change in texture, not change in intensity. Furthermore, because these features are computed based on comparisons with a center pixel, change cannot be detected if the intensity of the center pixel remains larger (or smaller) than each neighboring pixel after a change in a scene. Recently, Jodoin et al \cite{TF-115} proposed to use the local self-similarity (LSS) descriptor, but because the LSS descriptor is calculated on a large region, there are borders of falsely detected pixels around the detected foreground. Some morphological operations were applied to improve the detected foreground by removing extra pixels, but holes or spaces between legs cannot be recovered that way. Furthermore, LSS is slow to compute on complete image. \\

\subsubsection{Local Similarity Binary Pattern (LSBP)}  
To solve the shortcomings of both LBP and LSS, Bilodeau et al. \cite{TF-110} proposed a descriptor called Local Similarity Binary Pattern (LSBP) which is binary and fast to compute, works on small regions, and captures both change in texture and change in intensity. Practically, a binary vector is built from the comparisons of pixel intensities centered around a point of interest over a small predetermined pattern. Unlike
LBP and LSS which compute the difference between two values, the LBSP approach returns whether they are similar or not via absolute difference. Its temporal aspect and sensitivity to illumination variation is due to its ability to use the central pixel intensity of a previous frame for new comparisons. But, one of the disadvantages of LBSP is that it is not a spatiotemporal descriptor because it does not keep both feature information and pixel intensity information jointly up to date. To address this problem, St-Charles et al. \cite{TF-111}
presented a spatiotemporal LSBP in the framework of an adaptive background subtraction method called LOcal Binary Similarity segmenTER (LOBSTER). This modification allows the spatiotemporal LSBP to be more suitable in noisy or blurred regions and more robust to high illumination variations than the original LSBP. In further works, St-Charles et al. used this spatiotemporal LSBP in addition with color features with a flexible background subtraction algorithm called Self-Balanced SENsitivity SEgmenter (SuBSENSE) \cite{TF-112}\cite{TF-114}, and a background subtraction algorithm
that analyses the periodicity of local representations called Pixel-based Adaptive Word Consensus Segmenter (PAWCS) \cite{TF-113}\cite{TF-114-1}. \\

\subsubsection{Directionnal Rectangular Pattern (DRP)}  Zhang et al. \cite{TF-117}  developed a Directional Rectangular Pattern (DRP) based complex background modeling method to detect the moving objects in a video sequence. Different from LBP encoding the binary result of first-order derivative between the central point and its neighborhoods, Directional Rectangular Pattern encoded the binary result of first and second order derivative direction in all neighborhoods among a rectangular region. To model the distribution of the DRP micro-patterns, Zhang et al. \cite{TF-117} used DRP integral histograms. The local gray-level feature based Gaussian Mixture Model (GMM) is exploited to calculate an adaptive threshold for the histogram similarity measure to decide which part/pixel is background or moving object. Experimental results \cite{TF-117} show the effectiveness of DRP by comparing with LBP. \\

\subsubsection{Local Color Pattern (LCP)} 
All LBP-based algorithms are often invariant to local illumination changes but they are unable to detect uniform foreground objects in large uniform background except at the objects'edges. To solve this problem, Chua et al. \cite{TF-125} proposed a robust texture-color based background modeling. The texture feature is the LBP histogram and the color feature called Local Color Pattern (LCP) is  is formed by concatenating the quantized hue, luminance, and saturation histograms, summed over a structuring element. For the initialization, LBP and LCP histograms are computed and
stored as background models for the first $N$ image frames. Since the background model is represented by LBP and LCP histograms,
the final similarity is obtained by computing a weighted rule with a weight $\tau$ that controls the importance of texture and color features. Higher $\tau$ indicates that texture is more important than color features. $\tau$ is determined adaptively. The ability to adapt the weight of color and texture information makes the algorithm very suitable for video surveillance applications especially with dynamic scenes.  Reckley et al. \cite{TF-126} used LCP features in a sensor selection scheme in which spatiotemporal signatures of moving objects are integrated from different sensing modalities into a video segmentation method in order to improve object detection and tracking in complex scenes such as dynamic backgrounds with moving water and high reflections. Experimental results \cite{TF-126} on two complex datasets demonstrate that this technique significantly improves the accuracy and utility of the original LCP \cite{TF-125}\cite{TF-125-1}, and largely outperforms CS-LBP \cite{TF-33}. \\

\subsubsection{Local Neigborhood Patterns (LNP)} 
Amato et al. \cite{TF-127} used two discriminative features based on angular and modular patterns, which are formed by similarity measurement between two sets of RGB color vectors: one belonging to the background image and the other to the current image. This Local Neigborhood Patterns (LNP) improved foreground detection in the presence of moving shadows. \\

\subsubsection{Local Ratio Pattern (LRP)}
In LBP and LTP, the very coarse binning of local intensity ratios may result in substantial feature noise when the
true ratios fall close to a bin boundary. To solve this problem, Zaharescu and Jamieson \cite{MulF-40} proposed the Local RationPattern (LRP) in a multi-scale multi-feature codebook-based background subtraction method. LRP is similar to LBP and LTP, but extended to
$4$ bits, in order to divide the range of possible intensity ratios into $16$ bins, instead of $2$ or $3$. This finer-grained encoding allows the feature to respond more proportionately to a change in the underlying image. Then, a similarity between two LRPs is defined with a ratio. Furthermore, Zaharescu and Jamieson \cite{MulF-40} addressed the LRP reliability by evaluating the confidence of a match between two LRP features with their flatness (which are their similarity score to a perfectly flat LRP).  \\

\subsubsection{Local Ray Pattern (LRP)}  
Light field camera was originally proposed for image-based rendering for computer graphics. But, the light field camera has been applied to solve a difficult computer vision and pattern recognition problem too \cite{TF-128-100}. In this context, Shimada et al. \cite{TF-128}\cite{TF-128-1} proposed a feature representation, called Local Ray Pattern (LRP) to evaluate the spatial consistency of light rays. The combination of LRP and GMM-based background modeling realizes object detection on the infocus plane. Experimental results \cite{TF-128}\cite{TF-128-1} demonstrate the effectiveness and applicability for video surveillance. \\

\subsection{Spatio-temporal Patterns}
\label{subsec:STPatterns}

\subsubsection{Spatio-Temporal Vectors}
Pokrajac and Latecki \cite{TF-170} proposed to use as features Spatio-Temporal Vectors (STV), that are 3D blocks vectors. So, Pokrajac and Latecki \cite{TF-170} decomposed a given video into spatiotemporal blocks ($8\times8\times3$ blocks). Then, a dimensionality reduction technique is applied to obtain a compact representation of color or gray level values of each block as vector of just a few numbers. The block vectors provide a joint representation of texture and motion patterns. Then, the MOG model is used on the spatiotemporal blocks. So, the MOG model \cite{CF-1} is applied on the pixel and region levels with a single level texture representation, that is the 3D block, whereas it has been applied on pixel level in the original MOG. This method implicitly assumes the feasibility of computing projection matrix from blocks that adequately represent the texture. This approach provides good results on videos with comparatively high stationarity in background. Furthermore, improvements are possible if the projection matrix is computed dynamically. However, the techniques for adaptive estimation of projection coefficients in time are not studied. \\

\indent Texture at a given pixel is very likely to highly vary when a moving object is passing through this location. Therefore, Latecki et al. \cite{TF-171}\cite{TF-174}\cite{TF-175} proposed to use a local variation of the texture vectors. To robustly measure this variation, Latecki et al. \cite{TF-171}\cite{TF-174}\cite{TF-175} measure it in a limited and as short as possible window of time, since at a given pixel a moving object can quickly appear or disappear. So, the local variation is defined as the largest eigenvalue of spatio-temporal texture vectors in a small time window. It is computed by applying PCA to the covariance matrix of the SP texture vectors within a small temporal window. This way, Latecki et al. \cite{TF-171} indirectly determined the magnitude of texture variability in the direction of its maximal change. Thus, Latecki et al. \cite{TF-171}\cite{TF-174}\cite{TF-175} used PCA twice, first time to compute the sp texture vectors, and the second time to compute the variation of a set of texture vectors in a given time window. The decision whether a moving object or a stationary background is identified at a given spatiotemporal location is then made by dynamic thresholding of the obtained eigenvalues. \\

\indent Zeljkovic et al. \cite{TF-172} and Pokrajac et al. \cite{TF-173} studied the resilience of moving objects detection algorithm based on the spatiotemporal blocks on additive Gaussian noise. Experimental results \cite{TF-172}\cite{TF-173} on the PETS 2001 dataset show that the STV is robust to strong additive Gaussian noise. \\

\indent Pokrajac et al. \cite{TF-176} evaluated the STV on monochrome and multispectral IR videos. Experimental results show that the STV can provide low false positive error rates and successful identification of the front edge of the moving object. \\

\indent  Latecki et al. \cite{TF-177} extended the original method of STV \cite{TF-170} by replacing the dynamic threshold with dynamic
distribution learning and outlier detection significantly improving the performance of the original approach. \\

\indent Miezianko and Pokrajac \cite{TF-178} proposed a local background dissimilarity measurement  based on wavelet decomposition of
localized texture maps. Dynamic threshold of the normalized dissimilarity measurement identifies changed local background blocks, and spatial
clustering isolates the regions of interest.  The use of STV provides accurate and illumination invariant separation of foreground and background textures. The rough dissimilarity measurement of collected background texture maps and the dynamic thresholding locate regions of
interest in the background that exhibit significant changes. The clustering of the blocks in localized spatial regions establish the boundaries of the background at different times. The texture maps are created with gray level values to increase the computational speed of the wavelet decomposition and background change detection. \\

\indent Miezianko and Pokrajac \cite{TF-179} developed an effective method for extracting changed backgrounds
for regions observed by multiple overlapping cameras. A multi-layers background is constructed, and spatiotemporal texture blocks are used to detect motion. Detected motion delineated temporally cohesive non-moving regions to extract hyperspherical clusters and detect changed backgrounds. The proposed method allows for detecting background changes in crowded environments exhibiting large motion flows and significant occlusion. \\

\subsubsection{Spatio-Temporal Textures}
Yumiba et al. \cite{TF-180}\cite{TF-181} proposed a spatio-temporal texture called Space-Time Patch (ST-Patch) which describes motion in addition to appearance for detecting moving objects in presence of dynamic changes. This approach can cover global changes by using appearance information as conventional spatial textures. In addition, it can cover local changes by using motion information. Yumiba et al. \cite{TF-180} applied the MOG model with the ST-Patch. Experimental results \cite{TF-180}\cite{TF-181} show that ST-Patch allows the MOG to be more robust on videos with dynamic backgrounds and illumination changes.

\subsubsection{Spatio-temporal Features}
In a multi-level approach, Tanaka et al. \cite{TF-190}\cite{TF-192} and Nonaka et al. \cite{TF-191} proposed a integrated background modeling based on spatio-temporal features. First, foreground objects are detected based on the pixel-level background modeling which is a KDE model with the location (x,y) and the RGB color components as features and based on the region-level background modeling with Radial
Reach Correlation (RRC) as feature. In the further step, the two foreground masks are combined. That is, pixels which are classified as foregrounds by both of the pixel-level and region-level are classified as foregrounds and other pixels are classified as backgrounds. Then, a frame-level background modeling  which is based on brightness normalization of a model background image Fukui et al. \cite{TF-195}\cite{TF-196}, is used to be
robust against sudden illumination changes. Experimental results on the CD.net 2012 dataset show that this multi-level approach with the considered spatio-temporal features outperforms the original RRC \cite{TF-152}, the orginal MOG \cite{CF-1} and the improved KDE \cite{TF-197} in presence of dynamic backgrounds and illumination changes. 

\subsection{Statistical Texture Features}

\subsubsection{Peripheral Patterns}
\label{subsubsec:PeripheralPatterns}

\begin{itemize}
\item \textbf{Peripheral Increment Sign Correlation (PISC):} Satoh et al. \cite{TF-1} proposed Peripheral Increment Sign Correlation (PISC) feature that encodes the intensity differences between a target pixel and its peripheral pixels as a 0/1 binary code similar to the case of Local Binary Pattern (LBP) proposed by Heikkila et al. \cite{TF-10}. However, this leads to increase false positives because the code is reversed easily with slight intensity changes in regions with small intensity differences, for example in plain regions. Plain regions often occupy large spatial region within images, which makes stabilizing on them very important. \\
\item \textbf{Peripheral TErnary Sign Correlation (PTESC):} In order to alleviate the stabilization issue of PISC, Yokoi \cite{TF-5} proposed a Peripheral TErnary Sign Correlation (PTESC) features that stabilized the encoding by using -1/0/1 ternary code. \\
\end{itemize}

Though these texture-based methods are robust against illumination changes, they cannot cope with regions that have poor texture. Plain foreground objects before plain background with different intensity from foreground cannot be detected by these methods because both foreground and background have the same plain texture. For this, Yokoi \cite{TF-6} proposed a texture descriptor combining Peripheral TErnary Sign Correlation (PTESC) \cite{TF-5} and Bi-Polar Radial Reach Correlation (BPRRC) \cite{TF-132}\cite{TF-133}. The idea is to take advantages of the both descriptors. Indeed, Peripheral TErnary Sign Correlation (PTESC)  is robust against illumination changes by using -1/0/1 ternary code for encoding the intensity difference between pixels in texture, and Bi-polar Radial Reach Correlation (BPRRC) \cite{TF-132}\cite{TF-133} yields high detectability in a region with little texture.

\subsubsection{Reach Patterns}
\label{subsubsec:ReachPatterns}

\begin{itemize}
\item \textbf{Radial Reach Correlation (RRC):}  Satoh et al. \cite{TF-130}\cite{TF-131} defined a texture feature called Radial Reach Correlation (RRC) (also called Radial Reach Filer \cite{TF-134}\cite{TF-150}\cite{TF-151}\cite{TF-152}\cite{TF-153}\cite{TF-154} )which achieves robust detection of moving objects while remaining insensitive to moving object intensity distribution and
intensity changes in scene objects. RRC evaluates foregroundness based on local texture described in the magnitude relation between the center pixel and its neighbor pixel. In principle, this magnitude relation is not affected by the changes of illumination and, thus, RRC is more robust than features which  only  use  distribution  information  of the center pixel values. However, it cannot handle the changes of the textural information caused by the small background fluctuation such as swaying tree leaves \cite{TF-139-1}. In further works, Satoh and Sakaue \cite{TF-132}\cite{TF-133} proposed Bi-Polar Radial Reach Correlation (BP-RRC) which expands RCC in cases where the image texture is feeble and the intensity distribution is biased. Thus, Satoh and Sakaue \cite{TF-132}\cite{TF-133} used a mechanism that simultaneously defines and utilizes intensity differences (positive or negative) relative to a focal pixel. Experimental results \cite{TF-132}\cite{TF-133} with the RCC showed that, even in cases where image texture is weak or intensity distribution is biased, BP-RCC offers superior detection performance and stability. BP-RRC is robust against gradual illumination changes by using texture model but its present several weakness: \textit{1)} it is not robust against background movements because of its rigid texture model \cite{TF-136},  \textit{2)} Since BP-RRC uses the texture information, it does not detect offset/gain change, and  \textit{3)} it cannot respond to the texture change due to illumination change under multiple light sources. To address the first weakness, Yokoi \cite{TF-136} proposed the Probabilistic BP-RRC (PrBP-RRC) which  preserves BP-RRC's robustness against illumination changes and adds the robustness against background movements. PrBPRRC introduces a probabilistic model for background texture and learns a probabilistic background with inputs including background movements and presence of moving objects. Experimental results \cite{TF-136} on ATON dataset \cite{9070} and PETS 2007 dataset show that PrBP-RRC outperforms BP-RRC. To solve the second problem and the third problem, Miyamori et al. \cite{TF-140} proposed the Adaptive Bi-Polar Radial Reach Correlation Mixture Model (ABP-RRC), which generates and selects several BP-RRC accommodative to background changes. Thereby, Miyamori et al. \cite{TF-140} developed a stable background/foreground separation method, even when the texture changes occurred which various illumination change causes.  To reduce the computation time of RRC, Itoh et al. \cite{TF-135}\cite{TF-135-1} developed the Fast Radial Reach Correlation (F-RRC). As a result, Itoh et al.\cite{TF-135}\cite{TF-135-1} reduce the RRC calculation process to about $1/4$ in comparison with normal RRC. In an other approach, Tanaka et al. \cite{TF-138}\cite{TF-139}\cite{TF-139-1} improved the original RRC \cite{TF-130}\cite{TF-131} which is further used and combined in pixel-level and region level method. \\ 
\item \textbf{Radial Proportion Filter (RPF):} Miyamori et al. \cite{TF-160} defined the Radial Proportion Filter (RPF). In a further work, Miyamori et al. \cite{TF-161} developed a Multi-RPF (MRPF). \\
\item \textbf{Statistical Reach Feature (SRF):} Iwata et al. \cite{StF-40} defined a set of statistical pair-wise features, derived by intensity comparison in a local neighborhood. The classification for each pixel $P$ is achieved by a set of selected $N$ points $Q_n$ ($1\leq n \leq N $), which are called reference points. Thus, SRF can be described as the selection of $Q_n$. Three factors impact the search of $Q_n$ : \textit{1)} the absolute value of the intensity difference between $P$ and $Q_n$ must exceed a given threshold $T$ which has an important role because it allows the background model to tolerate noise, \textit{2)} $Q_n$ needs to meet a statistical requirement, that is its intensity remains $T$ units smaller (or larger) than that of $P$ in most images, and  \textit{3)} $Q_n$ is searched from the starting point $P$ to the edge of image in $N$ (with $N=8$) directions. Then, two types of $Q_n$ can be selected. SRF defines the sign between $P$ and $Q_n$, which satisfies $I_{t}(P)-I_{t}(Q_{n}) \geq T$ in most images, as $SRF(P,Q)=1$. It also defines the sign between $P$ and $Q_n$, which satisfies $I_{t}(P)-I_{t}(Q_{n}) \leq T$  in most images, as $SRF(P,Q)=-1$. Then, comparing the signs in the background model with that in the current image allows to classify $P$ as a background or foreground pixel, depending on whether the sign between $P$ and $Q_n$ has changed. Due to the special properties of point pairs, SRF works well in object detection, but SRF present three weakness: \textit{1)} SRF may not search a sufficient number of $Q_n$ because SRF searches for $Q_n$ in only eight directions, implying that most pixels are not taken into account. This may lead to an insufficient number of $Q_n$ that causes drawbacks in the foreground detection step, \textit{2)} the searching way of $Q_n$ is not optimal because SRF searches$Q_n$ in the order of space instead of the order of intensity difference, which leads to an incomplete detection. The larger the intensity difference is, the less sensitive the model becomes. The intensity difference between P and $Q_n$ searched by SRF tends to be too large. This leads to incomplete detection, and \textit{3)} the one-sided criterion in the SRF detection step results in false detections. A one-sided criterion means that the two signs of SRF are combined into one binary decision. In the case of presence of a moving object with much brighter color, pixel are misclassified as background.
To solve these problems, Zhao et al. \cite{StF-41} improved SRF both in theory and algorithm. \\
\end{itemize}

\subsection{Fuzzy Texture Features}

\begin{itemize}
\item \textbf{Local Fuzzy Pattern (LFP):} Ouyang and Chen \cite{TF-121} proposed a texture descriptor called the Local Fuzzy Pattern (LFP) histogram. The LFP histogram of each pixel is calculated for each new frame and is compared to its corresponding values in the background model for classifying the pixel into foreground objects. Experimental results \cite{TF-121} show that LFP possesses better adaption capability and tolerance for dynamic conditions such as shadows and illumination variation in comparison to LBP. Liu et al. \cite{TF-120} developed a similar LFP.\\
\item \textbf{Fuzzy Statistical Texture (FST):} Chiranjeevi and Sengupta \cite{FF-10} proposed to apply a fuzzy membership transformation on the
co-occurrence vector. The idea is to derive a fuzzy transformed co-occurrence vector with shared membership values in a reduced dimensionality vector space, and called Fuzzy Co-occurrence Vector (FCV). FCV can handled better the dynamic backgrounds than the crisp Co-occurrence Vector (CV). Then, Chiranjeevi and Sengupta \cite{FF-10} defined a normalized FCV (NFCV) from which fuzzy statistical texture features are derived. Practically, the background model is initialized with a feature vector, composed of intensity, energy, texture mean and local homogeneity, obtained from the
first frame of the video sequence. Then, FST are fused with intensity by using the Choquet integral. This approach is called IFST. For a fair comparison, Chiranjeevi and Sengupta \cite{FF-10} implemented the approaches with Statistical Features (FT), called IST. Experimental results \cite{FF-10} show that IFST  handled better dynamic backgrounds than IST, the Choquet integral with (RGB) features \cite{FA-15}, and the Choquet integral with (R,G,LBP) features \cite{FA-13}. In a further work, Chiranjeevi and Sengupta \cite{FA-40} used the same FST combined with intensity but aggregated with he intervalued-valued fuzzy Choquet integral instead of the real-valued Choquet integral used in Chiranjeevi and Sengupta \cite{FF-10}.\\
\end{itemize}

\subsection{Others Texture Features}
\subsubsection{Texture Pattern Flow}
Both LBP and SILTP-based approaches do not consider the temporal variations in patterns which contains inherent motion information derived from moving objects in the video. In order to address this problem, Zhang et al. \cite{TF-201} developed a spatial-temporal feature, dubbed as the Texture Pattern Flow (TPF), to compute inherent motion information. TPF encodes texture and motion information in a local region from both spatial and temporal aspects. TPF features are based on the gray-level image in order to increase robustness against complex and dynamic backgrounds.  The integral histogram of the TPF is computed within a local region around the pixel and is employed to extract statistically discriminative features. After estimating the TPF integral histogram features, the background model is built and dynamically adapted to complex scenarios as new frames begin to be processed. A similarity measurement between the current image and the background model is made using a Kernel Similarity Modeling (KSM) approach which integrates the TPF integral histogram features. Zhang et al. \cite{TF-202} proposed an improvement to this method using an adaptive threshold. \\

\subsubsection{Textons}
Texture Pattern Flow (TPF) \cite{TF-201} is based on the assumption that the movement of the foreground is always towards a certain direction. While this assumption can mostly work with human movements in constrained environments, it may not hold for animals such as a fish as their movement is fairly erratic with frequent direction changes \cite{NETTIES2006}\cite{SIGMAP2007}\cite{IWSSIP2007}\cite{TF-210}. However, other texture feature called textons have been extensively adopted in texture analysis, and can be defined as sets of patterns shared over an image \cite{TF-210-1}. Practically, several texton images have been proposed in the literature \cite{TF-210-2}. For fish detection in underwater scenes , Spampinato \cite{TF-210} adopted $7$ textons to capture even the slightest texture variations within any considered rectangular region of size $w \times w$ centered on the considered pixel. As a global texture feature describing each pixel $(x,y)$ and its neighbors in the region, the energy of the texton is computed. Therefore, each pixel is represented with a feature vector which contains the location $(x,y)$, the three color components (RGB) and the energy of the textons. The joint domain-range model consists in the corresponding 6-dimensional space, on which the pdfs of the background and foreground models are built. This is performed by means of KDE. Experimental results \cite{TF-210} on the Fish4Knowledge dataset \cite{907} show that the texton feature allow the KDE model to be more robust than the orginal KDE with RGB \cite{204}, SILTP \cite{TF-70}, VKS with rgb \cite{LF-10} and VKS with Lab and SILTP. \cite{LF-10}. In an other work, Panda and Meher \cite{TF-211} used a Texton Co-occurrence Matrix (TCM) feature with the original MOG model \cite{CF-1} to deal with dynamic backgrounds. TCM is widely used in the  field of image retrieval.  The TCM feature integrates both the color, texture, and shape features and is computed in a neighbourhood region of each  pixel. Thus, TCM feature implicitly used image features and the spatial relationship between the pixels. Experimental results \cite{TF-211} on the I2R dataset show that the MOG with the TCM feature show better robustness than original OG with RGB, covariance-based background subtraction \cite{TF-18}, moments-based background subtraction \cite{FeatureM-1}, the original LBP \cite{TF-10} and intensity and texture-based background subtraction \cite{MulF-9-1} in presence of dynamic backgrounds. \\

\subsubsection{Galaxy Pattern}
Liu et al. \cite{TF-129} applied a binary descriptor called galaxy pattern \cite{TF-129-2} due to its robustness in presence of illumination changes and its efficiency for computation time compared to the state-of-the-art descriptors. Liu et al. \cite{TF-129} used background instances of galaxy patterns computed on non-overlapped regions from observed backgrounds to model the background. Experimental results \cite{TF-129} show that galaxy patterns allow the model to more robust to illumination changes and dynamic backgrounds than MOG \cite{CF-1}, Codebook \cite{302} and ViBe \cite{FS-51}. In a further work, Yang et al. \cite{TF-129-1} improved this approach with a fine level detection method to identify the label of each pixel. \\

\subsection{Discussion on Texture Features}
For local binary patterns, BGLBP \cite{TF-66}and XCS-LBP \cite{TF-63} appear to be the best LBP variants for background modeling and foreground detection while ST-SILTP and CS-SILTP seem to be the best LTP variants. SILS \cite{TF-90}, LSS \cite{TF-115}, LSBP \cite{TF-110} and LRP \cite{MulF-40} present a robust alternative to LBP. Furthermore, patterns can be extended to spatio-temporal patterns, statistical patterns and fuzzy textures patterns to provide more robustness in presence of dynamic backgrounds and illumination changes. Moreover, designed texture features are proposed as textons \cite{TF-210} to deal with specific challenges met in aquatic environments. \\

\section{Stereo Features}
\label{sec:StereoFeatures}
Stereo features encapsulate \textit{\textbf{spectral information}} in the depth domain, and are obtained through a specific sensor which can provided a quantitative estimate of disparity \cite{SF-1} or depth \cite{SF-100}. Traditional stereo systems can provide both color and depth information \cite{SF-1}\cite{SF-100}. However, it requires the setup of two cameras. Practically, stereo depth computation relies on finding small area correspondences between image pairs, and therefore does not produce reliable results in regions with little visual texture and in regions, often near depth discontinuities in the scene, that are visible in one image but not the other. Most stereo depth implementations attempt to deal with such cases of limitations and label them with one or more special values. Furthermore, noise and subtle lighting changes can cause the depth measurement at a pixel to erroneous, even if nothing actually changes in the scene. In this context, shadows often provide the texture needed to extract valid depth in regions where measurements are usually invalid. \\

\indent On the other hand, Time of Flight Cameras (ToF) which is a range imaging camera system obtains distance based on the known speed of light and by measuring the time-of-flight of a light signal between the camera and the object for each point of the image. Apart from their advantages of high frame rates and ability to capture the scene all at once, ToF based cameras have generally the disadvantage of low resolution.  Several other limitations of ToF cameras are detailed in \cite{SF-204}. Nevertheless, the 2D/3D cameras provide additional depth features when compared to ordinary video, which makes it possible to deal with color camouflage issues during background modeling \cite{SF-200}\cite{SF-201}\cite{SF-202}. \\

\indent Recently, low cost RGB-D cameras such as the Microsoft's Kinect or the Asus's Xtion Pro are completely transformed several scientific visual signal processing applications. For example, Camplani et al.\cite{SF-300}\cite{SF-301}\cite{SF-302}\cite{SF-303}\cite{SF-305} used a Microsoft Kinect for objects detection through foreground/background segmentation. An other background subtraction algorithm based on RGB-D camera was developed by Fernandez-Sanchez et al. \cite{SF-400}\cite{SF-401}. RGB-D cameras based on structured light scanner (i.e., Microsoft Kinect) are not suitable for outdoor environments, due to the range limitation and errors introduced by interference with the sunlight. Stereo features provided by RGB-D cameras presents other challenging problems: \\
\begin{enumerate}
\item \textbf{Non-measured depth (NMD) pixels:} NMD pixels are mainly caused due to \textbf{\textit{1)}} occlusions (typically around object boundaries), \textbf{\textit{2)}} scattering of particular surfaces, \textbf{\textit{3)}} concave surfaces, \textbf{\textit{4)}} multiple reflections, \textbf{\textit{5)}} out-of-range points (very distant points), and \textbf{\textit{6)}} randomly, in homogeneous image regions. \\
\item \textbf{Noisy and irregular object boundaries:} Depth measurements at object boundaries are also heavily affected by noise. Sharp depth transitions produce misleading reflection patterns that result in rough and inaccurate depth measurements that are far from being correctly aligned to the actual object boundaries. \\
\item \textbf{Time dependent measurement noise:}  Depth measurements are also affected by instability over time and space. On one side, measurements taken for a static object that correspond to the same image pixel vary with time. On the other side, different depth values are obtained for spatially neighbouring pixels that correspond to points situated at the same distance from camera. The impact of this error varies with the distance. \\
\item \textbf{Distance dependent measurement noise:} The theoretical dispersion of depth measurements varies with the distance, following a quadratic law. This variation pattern has been confirmed in the study by Camplani et al. \cite{SF-301}. \\
\item \textbf{Camouflage in depth:} Similar depth between background and foreground lead to the problem of camouflage in depth. \\
\end{enumerate}
Other limitations of the Microsoft Kinect based depth features are detailed in \cite{SF-600}. The stereo features allows dealing with camouflage in color and hence are generally used in conjunction with color features. However, the impact of using depth features independently from other features has been studied in \cite{SF-304}\cite{SF-600}. In the following paragraphs, we review the different exiting approaches and the reader can see how the stereo features are fused with other features in Section \ref{sec:MultipleFeatures}.

\subsection{Disparity}
Ivanov et al. \cite{SF-1}\cite{SF-2}\cite{SF-3} was among the first authors who proposed a background subtraction method based on disparity verification. The use of disparity as a feature within background modeling allows coping to changes in illumination. The set-up for the extraction of the disparity map included three cameras called primary, left and right auxiliary cameras. A  basic background disparity verification algorithm is applied on each two cameras views (primary, right) and (primary, left) for each pixel in the primary image:
\begin{itemize}
\item Using the disparity map find the current pixel in the auxiliary image, which corresponds to the primary current pixel.
\item If the two pixels have the same color, label the current pixel as background.
\item If the pixels have different colors, then the current pixel either belongs to the foreground class, or corresponds to occlusion/shadow (a region of the primary image which is not seen in the auxiliary camera view due to the presence of the actual object). 
\end{itemize}
By cross-verifying each pixel across three camera views, Ivanov et al. \cite{SF-1}\cite{SF-2}\cite{SF-3} can distinguish the foreground object from occlusion/shadows. Practically, this method required the off-line construction of disparity fields mapping the background images that contained no foreground objects. At runtime, foreground detection is made by checking background image to each of the additional auxiliary color intensity values at corresponding pixels. When more than two cameras are available, more robust foreground detection could be achieved as in the case of occlusion/shadows. Because this method only assumed fixed background geometry, illumination variation at runtime can be handled. Since no disparity search was performed, the algorithm could be implemented in real-time on conventional hardware. Experimental results \cite{SF-1}\cite{SF-2}\cite{SF-3} show that this method extracted robustly silhouettes even under illumination changes in indoor scenes. \\
\indent Eveland et al. \cite{SF-10} proposed the use of a disparity feature which was coded from $0$ to $63$, with higher values being brighter and closer to the camera, respectively. A single Gaussian \cite{CF-50} background model is applied on the disparity feature to achieve reliable foreground detection in indoor video scenes. Results have been applied to a real-time tracking system. \\
\indent Gordon et al. \cite{SF-11} modelled the background using a multidimensional mixture of Gaussians model with the (R,G,B,D) features. A significant advantage of incorporating both color and depth features within the background model is that, Gordon et al. \cite{SF-11} could correctly estimate depth and color of the background when the background is available in a fewer number of initialization frames. For pixels which possessed significantly invalid range of values in depth, Gordon et al. \cite{SF-11} relied on using only the color features. Finally, Gordon et al. \cite{SF-11} used a disjunction of the results coming from each feature to obtain the final foreground detection. A pixel classified as foreground based on either color or depth is taken to be foreground in the final foreground detection.

\subsection{Depth}
Depth-based detections result in compact silhouettes, are not often affected by illumination changes or shadows but is likely to show imprecise and noisy contours and, unclassified pixels due to NMD pixels.\\

\subsubsection{Depth from Stereo Systems}
Harville et al. \cite{SF-100}\cite{SF-101}\cite{SF-102} modelled the background using a mixture of Gaussians with the (Y,U,V, D) features. Color and depth features are considered independent and the same updating strategy of the original MOG \cite{CF-1} is used to update the distribution parameters. The matching strategy of the original MOG \cite{CF-1} is adapted to deal with combination of color (Y,U,V) and depth (D) features. At low luminance, the chroma components U and V  become unstable and hence chroma is disregarded when comparing current observations. Similarly, when the depth becomes invalid, it is ignored. One the contrary, if in case a reliable distribution match is found for the depth component, the color-based matching criterion is relaxed thus reducing the color camouflage error. Similarly, in case that the stereo matching algorithm becomes unreliable, the color-based matching criterion is set to be harder to avoid problems such as shadows or local illumination changes. As in Gordon et al. \cite{SF-101}, a pixel that is detected as a foreground based on either color or depth is classified as foreground in the final foreground detection. \\

\indent For more reliability when combining color with depth, Song et al. \cite{SF-103}, instead of appending depth information into a color vector, designed two probabilistic background models corresponding to color and depth based on MOG \cite{CF-1}. Then, the combined probability and hence the foreground detection, is marginalized as the product of each individual probabilities. In addition, Song et al. \cite{SF-103} also incorporate a scheme that handles noise in depth images to improve the accuracy and robustness of foreground detection. 

\subsubsection{Depth from ToF Cameras}
\indent Silvestre \cite{SF-200} proposed to used both the grey-scale (I) and the depth information provided by a ToF camera (SwissRanger). The background was modeled by the MOG \cite{CF-1} applied on (I,D) features. Then, the distance between the samples has been considered in a two-dimensional grayscale-depth space. Thus, the depth and the grey-scale information are considered dependent. In the same way, Silvestre \cite{SF-200} adapted the KDE \cite{204}. The probability to belong to the foreground is then given to each pixel according to the difference between the current and background pixel depth and brightness. \\

\indent Langmann et al. \cite{SF-201}\cite{SF-202} modeled the background using a mixture of Gaussians with the (Y,Cr,Cb, D, a) features where $a$ is an amplitude modulation value. Thus, the 2D/3D camera produced a full size color image, low resolution depth and amplitude modulation images which are resized to match to color images by the nearest neighbor method. The matching function assumed that observations in the color, depth and amplitude modulation dimensions are in practice not independent. Indeed, a foreground object has most likely only a different depth, and also at least a slightly different color. Other reasons concern the limitations of ToF cameras as the infrared reflectance of an object has an influence on the depth measurement. Therefore, a linkage between the dimensions reduces the noise level in the foreground mask, the amount of misclassification due to
shadows and block artifacts which occur when only depth measurements are inappropriate. \\

\indent Stormer et al. \cite{SF-203} used also a MoG model \cite{CF-1}, where depth and infrared features are combined to detect foreground objects
in case of close or overlapping objects. Two independent background models are built. Each pixel is classified as background or foreground only if the two models matching conditions agree. But a failure of one of the models affects the final pixel classification. \\

\indent  Leens et al. \cite{SF-204} combined color and depth features, obtained with a low resolution ToF camera in a multi-camera system. The ViBe
algorithm \cite{FS-51} is applied independently to the color and the depth features. Then, the obtained foreground masks are then combined with logical operations and then post processed with morphological operations. \\

\indent Hu et al. \cite{SF-210} realized the foreground detection by using a weighted average on the probabilities obtained from the MOG model \cite{CF-1}. The different weights are updated adaptively for each output of the classifier by considering foreground detections in the previous frames and the depth feature. Experimental results \cite{SF-210} show that the proposed approach can effectively solve the limitations of color-based or depth-based detection.

\subsubsection{Depth from RGB-D Cameras}
Camplani et Salgado \cite{SF-300} proposed a combination of classifiers by jointly considering color and depth features. This combination is based on a weighted average that allows to adaptively modifying the support of each classifier in the ensemble by using the foreground detections in the previous frames and the depth and color edges. Thus, Camplani et al. \cite{SF-300} developed a weights selection scheme. For all those pixels for which the depth measurements is not available, the depth-based classifier weight is set to $0$, and the color-based classifier weight is set to $1$. So, when depth data is not available in the current frame or in the background model, only the color feature is considered for the final pixel classification. For the pixels that do not belong to the nmd set, the weights are assigned following a function of the depth-image edges as depth data guarantee compact detection of moving object regions except that for the very noisy depth values at object boundaries. To reduce this effect, the influence of the color based classifiers is increased in these regions. For all those pixels that have valid depth data, the weights are assigned following a function of the depth-image edges to limit the effect of noisy depth values at object boundaries by using the color information in these zones. On the contrary, the depth information is more reliable in the regions far from depth-edges, since it guarantees compact foreground and it is resilient to shadows and illumination changes. This method reduced false detections due to challenges such as noise in depth measurements, moved background objects, color and depth camouflage, illumination changes and shadows. \\

\indent Camplani et al. \cite{SF-301} proposed a foreground detection that combines depth and color information and reduces the noise present
in the depth maps to improve their accuracy. Indeed, this method reduced the distance-dependent spatial noise and the nmd pixel effect while accurately preserving object depth boundaries, thanks to an adaptive filtering strategy. Temporal fluctuations are reduced by iteratively building a reliable color/depth model of the static elements in the scene. The parameters of the proposed filtering process and the temporal model are continuously adapted to the distance-dependent noise. The background model is based on two independent MoG models in color and depth, respectively. Binary image operations are used to combine foreground detection results in depth and color to obtain a binary mask which preserves depth-based compactness and color-based accuracy in the final detection. A real-time implementation on GPU architecture was developed by Camplani et al. \cite{SF-305}.\\

\indent Camplani et al. \cite{SF-302} used a multiple region-based classifiers in a mixture of experts fashion to improve the final foreground detection. It is based on multiple background models that provide a description at region and pixel level by considering the color and depth features. As Camplani et Salgado \cite{SF-300}, the combination of the four models (pixel-color, region-color, pixel-depth, region-depth) is based on a weighted average to efficiently adapt the contribution of each classifier to the final classification.\\

\indent Camplani et al. \cite{SF-303} used also a weigthed average scheme in a Bayesian framework. Thus, probabilities from color and depth features are combined. The weights are chosen as a function of the each input to increase the support of the most reliable classifier as in \cite{SF-300}. \\

\indent Camplani et al. \cite{SF-304} proposed a combination of two algorithms to obtain a high-quality foreground detection using only depth feature acquired by the first generation of Microsoft Kinect. The first algorithm is the original MOG algorithm \cite{CF-1} adapted for depth feature. The second algorithm is based on a Bayesian network, which explicitly exploits the spatial characteristic of the depth feature. This Bayesian network is able to accurately predict the foreground/background regions between consecutive frame using two dynamic models, which encode the spatial and depth evolution of the foreground/background regions regions. \\

\indent Fernandez-Sanchez et al. \cite{SF-400} used the Codebook model \cite{302}, which has been extended to integrate depth features. This method is called Depth-Extended Codebook (DECB). To combine depth and color information, depth cues are used to bias the foreground detection based on color. The inclusion of depth information is made in two different ways: \textbf{\textit{1)}} the first one considers depth as the fourth channel of the codebook, which has an independent mechanism from color and brightness, and \textbf{\textit{2)}} the second one biases the distance in chromaticity associated to a pixel according to the depth distance. The first approach called 4D-DECB improved the robustness of the color-based algorithm to sudden illumination changes, highlighted regions and shadows. As depth features are more robust to lighting artifacts and shadows, dependence between RGB and depth have been used in the second approach which can be interpreted in the following way: if an input pixel is considered to be foreground, but it is close enough to the threshold, the classification will take into account the knowledge about the depth value for that pixel. This modification produced less false positive detections than 4D-DECB without biasing the color threshold.  In the same idea, Fernandez-Sanchez et al. \cite{SF-401} improved this approach called DECB-LF (DECB- Late Fusion) by refining the foreground mask obtained from DECB (with biasing the color threshold) using the output of the color-based algorithm. Since depth images tend to have more noise than color ones, Fernandez-Sanchez et al. \cite{SF-401} evaluated a fusion method that reduces the impact of that noise in the resultant segmentation without using erosion or small region suppression. \\

\indent Gallego and Pardas \cite{SF-500} combined color and depth features to perform a more complete Bayesian segmentation between foreground and background classes. For the background, the model consists of two independent Gaussians per pixel, one in the RGB domain and the second one in the depth domain. For the foreground, two parametric region-based foreground models combined color, space and depth domains, called Spatial Color Gaussian Mixture Model (SCGMM) and the Spatial Depth Gaussian Mixture Model (SDGMM), respectively. Then, the method combined the spatial-color and spatial-depth region-based models for the foreground as well as color and depth pixel-wise models for the background in a Logarithmic Opinion Pool fusion framework used to correctly combine the likelihoods of each model. A posterior enhancement step based on a trimap analysis is also proposed in order to correct the precision errors that the depth sensor introduces. \\

\indent Greff et al. \cite{SF-600} provided a comparison of background models with only the depth feature. These models are the following ones: First frame without foreground objects, single Gaussian \cite{CF-50} and Codebook model \cite{302}. The best performance is obtained by the Codebook
model which eliminates the errors of uncertain and alternating regions without missing the true foreground. \\

\indent Ottonelli et al. \cite{SF-700}\cite{SF-701} used a logical operations to fuse the results coming from the color and depth, respectively. For each features, the background model is the original MOG \cite{CF-1}. A depth-based compensation factor is computed  using a logical AND applied on the depth mask, and the difference mask between the depth mask and the RGB mask. Then, the final mask is obtained by using a logical OR between the RGB mask and the depth-based compensation factor. \\

\indent To make the background and foreground models more robust to effects such as camouflage and illumination changes, Spampinato et al. \cite{SF-705} explicitly models the scene's background and foreground with a Kernel Density Estimation approach in a quantized x-y-hue-saturation-depth space after a preprocessing stage for aligning color and depth data and for filtering/filling noisy depth measurements. Experimental results in three different indoor environments, with different lighting conditions, showed that this approach achieves an accuracy in foreground segmentation over 90\% that the combination of depth data and illumination-independent color space proved to be very robust against noise and illumination changes. \\

\indent Song et al. \cite{SF-710} used the original MOG \cite{CF-1} with  color and depth information. For combining color and
depth information, Song et al. \cite{SF-710} did not add depth to a color vector but designed two probabilistic background
models corresponding to color and depth based on MOG and denoised  depth  image. Thus, they address solving the color camouflage problem and depth denoising. Experimental datasets made on their own dataset \cite{SF-710} show robustness in case of camouflage in color and depth.\\

\indent Liang et al. \cite{SF-720} developed a refinement framework based on LUV color space and depth. As the foreground detection may be very inaccurate in some cases such as shadowing and color camouflage, Liang et al. \cite{SF-720} refined the inaccurate results by a supervised learning way. Thus, features are re-extracted from the source when it is detected. Since the depth data is not accuracy especially at the edge region, the edge of color map is used to detect the pixel wrongly classified as foreground by depth data. The re-extrated features with the initial detection results are fed to classifiers to obtain a better foreground detection. Experimental results on the RGB-D Object Detection Dataset \cite{SF-300} show that the refinement method outperforms the following color-depth methods: Camplani and Salgado \cite{SF-300}, Camplani et al. \cite{SF-302} and Nguyen et al. \cite{SF-706}\cite{SF-706-1} in presence of color camouflage and shadowing. The code is available at Github\protect\footnotemark[4]. \\
 
\indent Amamra et al. \cite{SF-791} designed on the GPU a GMMM based background subtraction algorithm for joint RGB and depth. This parallel algorithm benefits from asynchronous data exchange between the host (CPU) and the device (GPU). In addition, the data structures are organized in the GPU to permit a higher memory coalescing. Practically, this algorithm works at 30 fps. \\

\footnotetext[4]{{https://github.com/leonzfa/RBGS}}

\subsection{Discussion on Stereo Features}
Stereo features can not be used alone and needs to be carefully used following their properties as developed in Nghiem and Bremond \cite{SF-750}. There is no study about the influence on how the depth is acquired (Stereo cameras, ToF Cameras, RGB-D Cameras) and the robustness in the foreground detection. \\

\section{Motion Features}
\label{sec:MotionFeatures}
The motion features provide \textit{\textbf{temporal information}} and are usually obtained via optical flow to deal with irrelevant motion in the background. Theoretically, the most robust optical flow strategy should be employed to obtain the best performance. But most of the optical flow algorithms are computationally slow. Three alternative approaches are then used to introduce temporal attributes: \textbf{1)} the ones based only on the difference between consecutive frames. Then, the background model is only computed in stationary regions of the scene, \textbf{2)} optical flow (computed on all the pixels) which is used to detect moving areas. As in the previous approaches, the background model is only computed in stationary areas, and \textbf{3)} optical flow is only computed on moving areas after foreground detection. In this case, optical flow allows the algorithm to distinguish the unimportant moving areas from the moving objects. Some fast optical flow algorithms such as the fast nearest neighbour field based optical flow algorithms EPPM \cite{MF-18-1} run in near real-time on state-of-the-art GPU. We review in the following paragraphs the different exiting approaches and the reader can see how the motion features are fused with other features in Section \ref{sec:MultipleFeatures}. \\

\indent Tang et al. \cite{MF-1} used the motion speed value in addition to the intensity in the MOG model \cite{CF-1}. The motion speed value at each pixel is obtained through saliency motion filtering. First, a motion map is obtained by consecutive frame difference of Gaussian from each frame, from which a number of feature points are extracted with Monte Carlo importance sampling. Their corresponding velocities are computed using  an optical flow algorithm. The consecutive frame difference allows to detect slow
moving objects, and speed up the algorithm by only applying the optical flow to the regions of change which are detected by
consecutive frame differenc. In that region, for each pixel, the motion is considered as salient motion if the pixel and its neighborhood move in the same direction in a period of time. \\

\indent Huang et al. \cite{MF-11}\cite{MF-14}\cite{MF-15}\cite{MF-16} used dense optical flow for describing motion vectors. Regions with coherent motion are then extracted as initial motion markers. Pixels not assigned to any region are labelled uncertain ones. Finally, a watershed algorithm based on motion and color is utilized to associate uncertain pixels to the nearest similar mark. Further, MRFs are used to formulate foreground detection as a labelling problem. The optimization over the MRF model is then performed. The posterior probabilities initialized with the ones computed with the MOG model \cite{CF-1} are maximized to obtain the final classification result. Finally, regions which have the same classification label and similar colors are merged to derive a more consistent foreground mask. Experimental results \cite{MF-14} on gradual illumination changes and shadows demonstrate the robustness of this method, but the computational complexity of the technique has not been mentioned. In similar studies, Huang et al. \cite{MF-10}\cite{MF-12}\cite{MF-13} used motion information captured through the difference of consecutive frames to model the background in stationary areas. \\

\indent Zhou and Zhang in \cite{MF-17} used Lucas-Kanade gradient-based method for computing optical flow. As this method can only be applied for small displacements and as the displacements of moving objects between consecutive frames is anticipated to be more than $15$ pixels, Zhou and Zhang \cite{MF-17} used a hierarchical coarse-to-fine warping technique based on a Gaussian pyramid decomposition. The original MOG \cite{CF-1} with the color features is used to model the background. Then, in the fusion step, Zhou and Zhang \cite{MF-17} only considered those foreground objects as moving objects, where the amplitude and direction of the optical flow are within the ranges of consideration. \\

\indent Using EPPM \cite{MF-18-1}, Chen et al. \cite{MF-18} ensured temporally-consistent background subtraction with optical flow estimation by tracking the foreground pixels. Here, motion information is integrated with a temporal $M$-smoother. A similarity measurement is obtained directly from optical flow estimation with the assumption that the background estimate for the same object appearing in the difference video frames should be identical. As the direct implementation of EPPM \cite{MF-18-1} is extremely slow as optical flow estimation is required between any two video frames, Chen et al. \cite{MF-17} developed a recursive implementation so that optical flow estimation is required only between every two successive frames. As described in previous approaches, the background model is initially obtained using the MOG model \cite{CF-1}. Then, a spatial and a temporal $M$-smoother are employed to obtain a spatially-temporally-consistent foreground mask. Experimental results \cite{MF-18} on the ChangeDetection.net dataset \cite{901} and SABS dataset \cite{902} show that this algorithm outperforms most of state- of-the-art algorithms. \\

\indent In an other approach, Dou and Li \cite{MF-19} proposed a moving object detection method based on SIFT flow \cite{MF-19-1}. SIFT flow addressed the problem of image registration by aligning a query image to a target image at the scene level by spatially warping the query image to match the target image. This alignment is achieved by using dense pixel-wise SIFT descriptors. One aim of this algorithm is to counteract object motion between two scenes. This allows SIFT flow to be well suited to correcting object plane motion. Thus, Dou and Li \cite{MF-19} proposed to tp model each pixel with a pixel-wise SIFT descriptor and facilitate dynamically updating the model. The background model is a K histogram of the SIFT flow. Experimental results on the I2R dataset \cite{900} demonstrate that the proposed approach provides an effective and efficient way for background modeling and foreground detection. \\

\indent Using multiple features, Zhong et al. \cite{TF-19} proposed to fuse texture (LBP\cite{TF-10}) and motion patterns. A temporal operator to obtain the motion pattern is formulated as follows: \\
\begin{equation}
T_{motion}(x,y)= \sum_{i=0}^7 b_{t_i}(x,y) 2^i
\label{EquationMotion1}
\end{equation}
where the function $b_{t_i}(x,y$) keeps the sign of the difference between the central pixel at location $(x,y)$
attime $t$ and its $i^{th}$ neighboring pixel in previous $(t-1)^{th}$ frame as follows:
\begin{equation}
b_{t_i}(x,y)= 1 ~~ \text{if} ~~ I_{t-1}(x_i,y_ui)>I_t(x,y) \\
b_{t_i}(x,y)= 0 ~~ \text{otherwise}
\label{EquationMotion2}
\end{equation}
where $I_t(x,y$) is the intensity value at pixel location $(x,y)$ at time $t$. For each pixel, its probability to be either a background or foreground is computed from the histogram of each feature. Then, the results are combined using a weighted average mechanism. Experimental results \cite{TF-19} show that the combination of LBP and motion pattern outperforms the original LBP \cite{TF-10} in presence of dynamic backgrounds. \\

\indent Martins et al. \cite{OT-10} used a bio-inspired feature called Magno channel in addition with color features. Thus, this approach merges information from two inherently different methods: \textbf{(1)} bio-inspired motion detection method using the Magno channel, and \textbf{(2)} a background subtraction algorithm based on pixel color information. Thus, the background subtraction can be any one existing algorithms such as MOG \cite{CF-1}, Flux Tensor with Split Gaussian model (FTSG) \cite{CF-1-1}, KDE \cite{204}, AMBER \cite{9100} and SuBSENSE \cite{9110}. The foreground detection is obtained by merging the detection of the two methods. Experimental results \cite{OT-10} on the CD.net 2014 dataset show that the hybrid approach always substantially improves the performance of the original background subtraction methods. \\

\section{Local Histogram Features}
\label{sec:LHistogramF}

\subsection{Local Histogram of Color}
\label{sec:LHistogramColor}
Local histograms on intensity or color provides color information about the neigborhood of the pixel that the one color value do not give. In addition to their invariance to image rotation and translation, histograms are easy to compute. However, without data quantization to reduce size, histograms need more time to compute. Furthermore, histograms are sensitive to quantization errors and require a more complex similarity measure for the comparison. The different local histograms of color developed in the literature can be classified as crisp and fuzzy local histograms.

\subsubsection{Crisp Local Color Histograms}
\label{subsec:CrispLHistogramColor}
\begin{itemize}
\item \textbf{Local Color/Edge Histograms:} First, Mason and Duric \cite{StF-2} used local color histograms in the RGB color space with local edge features. First, a depth reduction formula is applied to transform 24-bit color to 12-bit color to reduce computation time, and to reduce the complexity because 24-bit histograms are more hard to compare since 1-bit change  color value places the corresponding pixel into a different histogram bin. Furthermore, the lower four bits obtained by low-cost cameras that capture 24-bit video are often very noisy, and 12-bit pixel representations only require 4096 bins for each histogram. Thus, Mason and Duric \cite{StF-2} obtained smaller histograms which are much easier to build and compare. Then, edge histograms are composed of 36 bins. For each edge pixel, the bin index is computed using the edge orientation ($10^0$ per bin). When the bin index is dertermined, the bin is incremented with the dege magnitude. Finally, the histograms in color are compared by using an intersection measure while the histograms in edge are computed with a chi-squared measure. Experimental results \cite{StF-2} show that the Local Gradient Histogram (LGH) with edge allow the method a better outline of the objects than the Local Color Histograms (LCH). \\
\item \textbf{Local Kernel Color Histograms:} Noriega et al. \cite{StF-3} proposed Local Kernel Color Histograms (LK-CH) to keep the advantages of histograms avoiding their drawbacks. First, each image is segmented into overlapped local squares with a histogram for each one to collect spatial information. To reduce the amount of data, the color space is quantized according to the most representative colors extracted from the scene. Second, Noriega et al. \cite{StF-3} used a double Gaussian kernel, that are one in the image space and one in the color space, to be robust against noise. Then, local kernel histograms are computed from image overlapped regions using the two Gaussian kernels. Finally, a pixel scale probability map is obtained by using the Bhattacharyya distance. Experimental results \cite{StF-3} on the Wallflower dataset show that LK-CH is less affected by noise, camera vibrations and swaying trees than the mean, the median and the LCH \cite{StF-2}. But, local kernel histograms with color features cannot handle illumination changes. \\
\item \textbf{Estimated Local Kernel Histograms:} Li et al. \cite{StF-7} developed a nonparametric Estimated Local Kernel Histogram (ELKH) for moving objects detection in presence of dynamic backgrounds. First, Local Kernel Histogram (LKH) are built by using the correlation and texture of spatially proximal pixels. Then, the probability distribution of LKH is estimated with a nonparametric technique. The  Bhattacharyya  distance is used to measure the similarity of LKH between the estimated background model and the current frame. Practically, Li et al. \cite{StF-7} used the Estimated LKH which is a block-wise version of the LKH, and in which there are only several nonzero bins, and the others are all zero. That means intensity of color is congregative in a pixel block. The EKLH contains some texture information, and it is more robust to disturbing noise than LKH. Experimental results \cite{StF-7} show that ELKH reduced false detections due to dynamic texture robustly, but also allow to detect the small moving objects. \\
\item \textbf{Local Color Difference Histograms:} Li \cite{StF-1} used histograms in the YUV color space to obtain difference image of color  distance. Histograms are obtained by statistics of the difference image. According to the mono-modal feature of histogram of the difference image, Ji \cite{StF-1} employed an adaptive clustering method, and removed noise with morphological filtering. Finally, an updating scheme is used to adapt the model to the illumination changes and environmental conditions. Experimental results \cite{StF-1} show that LCH with the YUV color space offer robustnees in presence of illumination changes. \\
\item \textbf{Local Dependency Histograms:} Zhang et al. \cite{StF-10} \cite{StF-11} proposed the Local Dependency Histograms (LDH) to model the spatial dependencies between a pixel and its neighboring pixels for dynamic background subtraction. LDH is computed using the direct neighbors and the indirect neighbors. The direct dependencies can be along any directions. The indirect spatial dependencies are confined to be only along the horizontal or vertical direction. Based on LDH, Zhang et al. \cite{StF-10} \cite{StF-11} developed a dynamic background subtraction in which each pixel is modeled as a group of weighted LDHs. Foreground detection  is obtained by comparing the LDH computed in current frame against its model LDHs with a histogram intersection measure and an adaptive thresholding method. Then, The model LDHs are adaptively updated by the new LDH. Experimental results \cite{StF-10} \cite{StF-11} on the I2R dataset \cite{900} and CMU dataset \cite{905} show that LDH is more robust in presence of dynamic backgrounds and camera jitter than the original MOG \cite{CF-1}.\\   
\item \textbf{Spatiotemporal Condition Information:} First, Wang et al. \cite{FSI-3} built a spatiotemporal neighborhood based on the center surround visual saliency model. To reduce the false detection in the foreground mask, a neighborhood weighted spatiotemporal condition information (NWSCI) is used to classify pixel based on the similarity of neighborhood pixels. Second, a joint cascade and hierarchical framework reduced computational cost by rejecting the unchanged regions before foreground detection. Experimental results \cite{FSI-3} show that NWSCI based background subtraction effectively outperforms the original MOG \cite{CF-1}, the original KDE \cite{204} and ViBE \cite{FS-51}, and gives similar performance than \cite{FSI-3-1} and \cite{FSI-3-2}. 
\end{itemize}

\subsubsection{Fuzzy Local Color Histograms}
\label{subsec:FuzzyLHistogramColor}

\begin{itemize}
\item \textbf{Local Fuzzy Color Histograms:} Kim and Kim \cite{StF-200} adopted a clustering-based feature called Fuzzy Color Histogram (FCH)
(FCH) to attenuate color variations due to background motions and still highlight moving objects. FCH \cite{StF-200-1} is a fuzzy version of the conventional color histogram. In CCH, the quantized color feature is assumed to be into exactly one color bin and it often lead to abrupt changes even though color variations are actually small. On the other hand, FCH uses the fuzzy membership \cite{StF-200-2} to relax this crisp condition. Kim and Kim \cite{StF-200} computed FCH on the CIELab color space arguing that CIELab color correctly quantifies the perceptual color similarity. Then, the colors in the CIELab color space are classified into clusters using the FCM clustering technique \cite{FF-5-1}. The foreground detection is made by thresholding the difference between the local FCH in the background and the background. The similarity measure used is the normalized histogram
intersection for simple computation. Then, the background model is updated an online procedure. Experimental results \cite{StF-200} on the I2R dataset \cite{900} show that the Local FCH (LFCH) outperforms the original MOG \cite{CF-1}, the generalized MoG (g-MoG) \cite{CF-1-2}, STLBP \cite{TF-13} and the local CCH. In other works, Kanna and Murthy \cite{StF-210}, and Gutti and Shankar \cite{StF-240} employed the local FCH too while Kumar et al. \cite{StF-220} used it in an automated surveillance system. In an other approach, Yang et al. \cite{StF-230} added to local FCH spatial coherence and temporal consistency in a Markov random field statistical (MRF) framework. \\
\item \textbf{Local Fuzzy Color Difference Histograms:} First, Panda et al. \cite{StF-250} presented a background  subtraction  algorithm based on 
color  difference  histogram  (CDH) by  measuring the color difference between a pixel and its neighbors in a small local neighborhood. CDH reduced false detections due to the non-stationary background, illumination variations and camouflage. Secondly, the color difference is fuzzified with a Gaussian membership function. Finally, Panda et al. \cite{StF-250} proposed a Fuzzy Color Difference Histogram (FCDH) by using fuzzy c-means (FCM) clustering \cite{FF-5-1} and exploiting the CDH. FCM clustering algorithm applied to CDH  allow to reduce the large dimensionality of the histogram bins in the computation and also decrease the effect of intensity variation  due to the unimportant  motion or illumination changes. Experimental results \cite{StF-250} on the I2R dataset \cite{900} show that FCDH is more robust than the original MOG \cite{CF-1}, the original LBP, STLBP  \cite{TF-13}
LIBS \cite{SF-20}, FBS \cite{FA-15}, FST \cite{FF-10}, MKFC \cite{FF-6}, and LFCH \cite{StF-200}.
\end{itemize}

\subsection{Local Histogram of Gradient}
\label{sec:LHistogramGradient}

\begin{itemize}
\item \textbf{Local Gradient Histograms:} Please see the item "Local Color/Edge Histograms" in Section \ref{sec:LHistogramColor}. \\
\item \textbf{Local Histogram of Oriented Gradient:} Fabian \cite{StF-120} presented a MOG based background subtraction which exploits Histograms of Oriented Gradients (HoG) instead of intensity to deal with camera jitter, automatic iris adjustment and exposure control. Thus, Fabian \cite{StF-120} avoided false detections of foreground mask. Practically, the brightness value is replaced by a local image gradient because image gradient and its orientation are invariant against changes in brightness. This scheme applied only to a certain extent given by limited range of pixel values. In practice, this assumption is quite suitable and allow the method to reduce the effects of automatic control. Futhermore, to deal with camera jitter, Fabian \cite{StF-120} handled small image movements. For this, the gradient is measured for squared area of size $8 \times 8$ called cell. Finally, an on-line spatial rearrangement of cells minimized the variance of dominant gradient for every cell. Practically, the most significant orientation bin for every cell in the image is extracted and the mean HOG (MHOG) is computed to provide the reference values for minimizing the variance
of gradients and especially resulting bins. To measure the distance between two different bins, Fabian \cite{StF-120} defined a metric in a discrete metric space to obtain difference between two different bins, and such that a decision can be made when a same bin doesn't exist. Then, the HOG feature ares used in the MOG model. Experimental results \cite{StF-120} on traffic surveillance videos show the pertinence of the HOG against the intensity. In an other approach, Mukherjee et al. \cite{StF-131} used both RGB and HOG in the MOG model. First, a modified distance based on support weight is developed to compare RGB features, and a HOG distance is presented to compare HOG features in order to provide distinct cluster values. Second, a multi-layers model is employed to obtain the foreground mask. Experimental results \cite{StF-131} on five datasets show that this method is more robust in presence of illumination changes and dynamic backgrounds than the original MOG \cite{CF-1}, CRF-based MOG  \cite{CF-1-10}, self-adaptive MOG \cite{CF-1-20}, the Type-2 Fuzzy MOG (T2-FMOG) \cite{CF-1-100}, T2-FMOG with MRF  \cite{CF-1-101} and SOBS \cite{CF-203-1}. Javed et al. \cite{StF-100}. In an other work, Panda et al. \cite{StF-132} combined HOG and LBP for complex dynamic scenes.  Experimental results \cite{StF-132} on the I2R dataset show that LBP-HOG is more robust than LBP, STLBP \cite{TF-13}, the Choquet integral with RGB features \cite{FA-15} MKFC \cite{FF-6} in presence of dynamic backgrounds. \\
\item \textbf{Local Adaptive HOG:} Hu et al. \cite{StF-130} proposed to use HOG in a coarse-to-fine strategy. First, a HOG-based background model 
is constructed with a group of adaptive HOG. Then, the foreground detection is achieved by comparing HOG of each pixel in the current frame and the background model. To deal with some missing foreground regions, the foreground mask is  is improved by using the pixel-wise detection provided by MOG algorithm and morphological  operations. In  the  refinement  step, the foreground maskis  refined based  on  the  distinction  in  color  feature  to  eliminate  errors  such as  noises and shadows. In the experimental results \cite{StF-130} on the Wallflower dataset \cite{StF-130}, this method outperforms the original MOG \cite{CF-1} and the codebook model \cite{CF-310}. \\
\item \textbf{Local Orientation Histograms:} Orientation histograms were applied with success for visual tracking, and keep the probability density  function of local gradients which is robust against illumination changes and easy to compute. In this context, Jang et al. \cite{StF-6} developed  Local Orientation Histograms (LOH) with Gaussian kernel, thus obtaining 1D orientation histograms to reduce the quantization error. LOHs allow to compared the background and the foreground by dividing each frame into small local regions. Each local region has the foreground probability given by comparing LOH between background and foreground image. In a multi-scaled approach, Jang et al. \cite{StF-6} used an foreground detection
algorithm that dynamically partition and compare regions with a recursive partitioning algorithm. Because it requires multiple extractions of histograms from multiple rectangular cells, Jang et al. \cite{StF-6} used the integral histogram \cite{FS-1-2} to provide fast extraction of histogram over the multiple overlapped cells. Experimental results \cite{StF-6} show that LOH suppress local false detection in presence of illumination changes.\\
\item \textbf{Local Kernel Histograms of Oriented Gradient:} Noriega and Bernier \cite{StF-4} proposed Local Kernel Histograms of Oriented Gradient (LK-HOG). First, contour-based features are extracted for local kernel histograms. Experimental results \cite{StF-4} on the Wallflower dataset show that LK-HOG is more robust to illumination changes than the mean, the median, the LCH \cite{StF-2}, LK-CH \cite{StF-3}, LGH \cite{StF-2}, and local HOG. In all the videos, Gaussian kernels improved the quantization error rate to reduce both false positives and negatives. \\
\end{itemize}

\subsection{Local Histograms of Figure/Ground}
\label{sec:LHistogramFG}
Zhong et al. \cite{StF-12} represented each pixel as a local histogram of figure/ground segmentations, which combines several prospective solutions that are generated with simple background algorithms to get a more reliable and robust feature for background subtraction. The background model of each pixel is constructed as a group of weighted adaptive local histograms of figure/ground segmentations, which describe the structure properties of the surrounding region. The goal of using LH-FGs is to make up for deficiencies each individual algorithm, thus achieving a better overall performance than each single algorithm. First, a local histogram of figure/ground segmentations  is computed over a squared fixed-size $N \times N$ neighborhood  for each algorithm with integral histogram \cite{FS-1-2}. Then, LH-FGs of each algorithm are concatenated together to provide the final  representation of each pixel. The feature extraction step thus yields to $S$ (2-bins) histograms where $S$ denote the number of foreground detecion maps. Finally, the $S$ histograms are concatenated  together to give a final $2S$-bins histogram, which is normalized to sum to one, so that it is also a probability distribution. The similarity between two histograms is computed with the Bhattacharya distance. \\

\section{Local Histon Features}
\label{sec:LHistonF}
Histon and its associated measure Histon Roughness Index (HRI) were applied to still image segmentation with good performance. In this context, Chiranjeevi and Sengupta \cite{FF-1} introduced histon and rough set theory for foreground detection, extended the histon concept to a 3D histon, and incorporated fuzziness into the 3D HRI measure, and thus obtained 3D Fuzzy Histon. Then, the labeling decision is based on Bhattacharyya distance between the model HRI and the corresponding measure in the current frame. Practically, histon and these variants are defined as follows: \\
\begin{itemize}
\item \textbf{Histon:} Histon allows to visualize the color information for the evaluation of similar colored regions. For each intensity value in the histogram, the number of pixels which are in the similar color sphere is computed, and this value is added to the histogram value, to obtain the histon value of the corresponding intensity. Histogram and histon distributions give color information and spatial information. Identical distributions of both the histogram and the histon in a region indicate that the region present lacks in its spatial homogeneity or its spatial similarity. So, Chiranjeevi and Sengupta \cite{FF-1} used this property to model the pixel in the center of the region which is formulated by Histon Roughness Index (HRI), after correlating the histon concepts with the rough set theory. Finally, HRI incorporates both the color and the spatial information, and it is used for comparison between the background model and the current frame. \\
\item \textbf{3D Histon:} In the original formulation of histon, each color channel is considered separately, rather independently, to obtain the histon for each color channel. But, exploiting the three color component values of each pixel with a 3D spatial distribution is better to get more information than exploiting the spatial distribution of the three independent color planes. So, Chiranjeevi and Sengupta \cite{FF-1} proposed an integrated 3D histon, where the histon distribution is computed by using the color value on three channels jointly. Then, the 3D HRI distribution for a region, centered at a pixel, is calculated using the 3D histon and the 3D histogram. Experimental results \cite{FF-1} show the effectiveness of 3D histon compared to the basic histon. \\
\item \textbf{3D Fuzzy Histon:} In 3D histon, whether a pixel is similar to its neighbors or not, is determined crisply. By determining the extent of similarity using Gaussian membership function, Chiranjeevi and Sengupta \cite{FF-1} proposed the 3D fuzzy histon as an extension of the basic fuzzy histon. 3D Fuzzy histon is subsequently used to compute 3D Fuzzy Histon Roughness Index (3D FHRI). Experimental results \cite{FF-1} on the I2R dataset \cite{900} show the effectiveness of 3D Fuzzy histon compared to the 3D histon. \\
\end{itemize}

\section{Local Correlogram Features}
\label{sec:LCorrelogramF}
Correlogram is an image of correlation statistics, and it can also be used as feature as follows: \\
\begin{itemize}
\item \textbf{Correlogram:} Correlogram captures inter-pixel relationships in a block or a region, and alleviates the drawbacks of histogram, which   only considers the pixel intensities for calculating the distribution \cite{611-10}. So, it is suitable for modeling dynamic backgrounds as developed in Chiranjeevi and Sengupta \cite{FF-5}. But, computation of correlogram for RGB color components involves huge computations as the correlogram size is $256^3 \times 256^3$. Even with single color channel, it requires significant computations as  the  correlogram  size  then  becomes  $256 \times 256$. Hence, the single channel is quantized to a finite number of levels $l$. Due to this, the correlogram size  is further reduced $l \times l$ with $l \ll l$. \\ 
\item \textbf{Fuzzy correlogram:} Crisp assignment of quantized intensity pair to a particular correlogram  bin is sensitive to quantization noise \cite{FF-5}. To reduce  computational time and to address the crisp assignment limitations, Chiranjeevi and Sengupta \cite{FF-5} developed fuzzy correlogram, composed by applying fuzzy c-means (FCM) algorithm \cite{FF-5-1} on correlogram. Thus, each intensity pair is related to all the bins by their respective membership values. The membership matrix is obtained by using fuzzy c-mean algorithm where the Euclidean distance is used as a measure between the cluster centers and the data points. Fuzzy correlogram greatly reduces the number of correlogram bins and the quantization noise. Experimental results \cite{FF-5} on the I2R dataset \cite{900} show that fuzzy correlogram can handle multi-modal distributions better that approaches which used multiple model features.\\
\item \textbf{Multi-channel fuzzy correlogram:} In Chiranjeevi and Sengupta \cite{FF-5}, correlogram was obtained from a single color channel, and thus the color information is loosed. The dependency across the color components was also ignored. Because there is more information in the color  distribution than a single monochrome distribution, Chiranjeevi and Sengupta \cite{FF-6} proposed a inter-channel correlogram, which captures both the color information and the dependency across the color planes. However, inter-channel correlograms exclude the inter-pixel relations within the same color plane. To use this information too, Chiranjeevi and Sengupta \cite{FF-6} combined both inter-channel and intra-channel correlograms, called multi-channel correlogram which exploits both the color information by taking into account the color dependencies and the inter-pixel relations across and within the color planes, unlike the correlogram proposed in Chiranjeevi and Sengupta \cite{FF-5} that captures only the inter-pixel relations on a single color plane. Then, the correlograms are mapped to a space of reduced dimensionality by using a transformation  based on a fuzzy membership matrix, whose elements indicate the belongingness of each intensity pair to the new bins. Instead of fuzzy correlogram which used FCM to computed membership values with the disavandages that Euclidian distance is only suitable for spherical clusters and that FCM is sensitive to the outliers, Chiranjeevi and Sengupta \cite{FF-6} obtained the membership values by using Kernel Fuzzy C-Mean (KFCM) algorithm \cite{FF-7}. The membership values applied over multi-channel correlogram results in a feature called Multi-channel Kernel Fuzzy Correlogram
(MKFC). KFCM uses a kernel which is based on a metric distance in place of Euclidean distance, which made it more robust than FCM. Furthermore, MKFC involves less number of bins ($ k \ll l^2$) than the original correlogram of dimensionality $l^2$ and this significantly reduces the computational complexity in distance computation for foreground detection. \\
\end{itemize}

\subsection{Discussion on Histogram/Correlogram Features}
Local histograms/correlograms on color or edge gives color or edge information about the neigborhood of the pixel that the one color or edge value do not give. Thus, these features allow the background model to be more robust in presence of dynamic backgrounds and illumination changes, and their main drawback is the number of bins. Furthermore, several approaches were developed to address this problem. Furthermore, fuzzy Histogram/Correlograms allow to avoid crisp assignements in the clustering step. Practically, LFCH  \cite{StF-200} and FCDH  \cite{StF-250} provide good performances in the case of histograms while MKFC \cite{FF-6} appears as the best descriptor in the case of correlograms.

\section{Haar-like Features} 
\label{sec:HaarFeatures}
Haar-like features \cite{TDF-3000} have been used in different approaches in background modeling and foreground detection as follows:  \\
\begin{itemize}
\item \textbf{Haar-like Features in hierarchical approaches:} Zhao et al. \cite{TDF-3100} used Haar-like features in a hierarchical codebook approach. In the block-based level, four Haar-like features and a block average value are used to represent a block. These features are computed  using integral image to reduce computation time and are less
sensitive  to dynamic backgrounds. Thus, most of the background blocks are removed at this step without reducing the true positive rate. n  the  block that has  been detected as foeground, a pixel-based codebook is applied to increase the precision. Experiment results show \cite{TDF-3100} on the I2R dataset show that the Haar-like feature allow to be more robust in presence of dynamic backgrounds.\\
\item \textbf{Haar-like Features in multi-features approaches:} 
In an other approach, Klare \cite{FA-1} and Klare and Sarkar \cite{FA-2} used the Haar-like features in an extended feature set. In a similar way, Han and Davis \cite{FR-20} employed Haar-like features in a multi-feature approach with density based background subtraction. In an other way, Lopez-Rubio and Lopez-Rubio \cite{FS-20} employed Haar-like features in a method based on stochastic optimization \cite{FS-20-1}. Please see Section \ref{sec:MultipleFeatures} for details. \\
\item \textbf{Haar-like Features in feature selection approaches:} Grabner and Bischof \cite{FS-1}, Grabner et al. \cite{FS-3} and Lee et al. \cite{TF-25} used Haar-like features in an online boosting algorithms which select the best combination of features. Please see Section \ref{sec:FeatureSelection} for details. \\
\end{itemize}

\section{Location Features} 
\label{sec:LocationFeatures}
Pixel-wise models ignore the dependencies between proximal pixels and it is asserted here that these dependencies are important. Thus, several authors proposed to use the location $(x,y)$ of the pixel to add to the \textit{\textbf{spatial information}}. This location information is used directly in a pixel manner \cite{LF-1}\cite{LF-2} or indirectly via invariant moments in a region manner  \cite{FeatureM-1}:
\begin{itemize}
\item \textbf{Location (x,y):} First, Sheikh and Shah \cite{LF-1}\cite{LF-2} used the location (x,y) in addition to the color features (Normalized RGB) in a joint representation called "domain-range representation" that provides a direct means to model and exploit the dependency between the pixels. Thus, a kernel density estimation (KDE) model represents the background and foreground processes by combining the three color dimensions and two spatial dimensions into a five-dimensional joint space. But, this method was found to be dependent on the size of the image. Indeed, the classification criterion, based on the ratio of likelihoods in this five-dimensional space, has an undesirable dependence on the size of the image.  Similar to the work of Sheikh and Shah \cite{LF-1}\cite{LF-2}, Narayana et al. \cite{LF-10} modelled the foreground and background likelihoods with a KDE using pixel samples from previous video frames but by building an explicit model of the prior probability of the background and the foreground at each pixel. Moreover, Narayana et al. \cite{LF-10} modelled the processes using a three-dimensional color distribution at each pixel. In addition, Narayana et al. \cite{LF-10} incorporated spatial priors for the background and foreground processes by using classification labels from the previous frame. Large spatial covariance allows neighbouring pixels to contribute more to the score at a given pixel location. Color covariance allows for some color appearance changes at a given pixel location. In this framework, the distributions are conditioned on spatial location, rather than being joint distributions over location and color. So, the model of Narayana et al. \cite{LF-10} avoids the dependence on the image size and yields better results. Furthermore, Narayana et al. \cite{LF-10} used Lab and SILTP \cite{TF-70} as color and spatial features respectively. Furthermore, instead of an uniform kernel \cite{LF-1}\cite{LF-2}, an adaptive kernel scheme is used and is nearly as accurate as the full procedure but runs much faster. Narayana et al. \cite{LF-11}\cite{LF-12} improved this scheme by using a separation of the foreground process into "previously seen" and "previously unseen" foreground processes, and by using explicit spatial priors for the three processes - background, previously seen foreground, and previously unseen foreground. The probabilistic formulation with likelihoods and a spatially dependent prior for each process leads to a posterior distribution over the processes.  Instead of a constant kernel bandwidth as in Sheikh and Shah \cite{LF-1}\cite{LF-2}, Antic and Crnojevic \cite{LF-20} adapted the kernel bandwidth according to the local image structure. Thus, image gradient is used to adaptively change the orientation and dimensions of the kernel at the borders of the region. This adaptive scheme provides more accurate modeling of non-stationary
background containing regions with different texture and illumination and suppress structural artifacts present in detection results when the kernel density estimation with constant bandwidth is used. \\
\item \textbf{Moments:} Marie et al. \cite{FeatureM-1} used invariant moments based on the Hu set moments \cite{FeatureM-1-1}). Each pixel is modeled as a set of moments computed from its neighborhood and stored using a codebook model \cite{CF-310}. Hu \cite{FeatureM-1-1} defined 7 invariants, but their complexity increases dramatically. Moreover, the results obtained using the three first are very similar than the one with all the moments. Practically,  Marie et al. \cite{FeatureM-1}  retained only the moment named "I1", which is computationally the fastest. Experimental results \cite{FeatureM-1} on the Sheikh sequence \cite{905} demonstrate that the codebook with moment outperforms the original codebook \cite{CF-310}, the original MOG \cite{CF-1}, and approximated median \cite{101}.
\end{itemize}

\section{Tranform Domain Features}
\label{sec:TransformDomainFeatures}

\subsection{Features from Frequency Domain Transform}
Transform domain features obtained through frequency analysis can provide valuable \textit{\textbf{spectral and spatial information}} that can in-turn lead to better background modeling and foreground detection. Some of the frequency domain features include:\\
\begin{itemize}
\item \textbf{Fourier Transform Features:} Fourier transform features encapsulate \textit{\textbf{spectral information}} which are suitable for scenes that contain periodic patterns. That is, scene where is significant correlation between structures in the scene and the observations across time. For example, a tree swaying in the wind or a wave lapping on a beach is not just a collection of randomly shuffled appearances, but a physical system that has characteristic frequency responses associated with its dynamics.  In this context, Wren and Porikli \cite{TDF-1} estimated the background model that captures spectral signatures of multi-modal backgrounds using Fast Fourier Transform (FFT) features through a method called Waviz. Here, FFT features are then used to detect changes in the scene that are inconsistent over time. Experimental results \cite{TDF-1} show robustness to low-contrast foreground objects in dynamic scenes. In an other work, Tsai and Chiu \cite{TDF-1-1} presented a background subtraction method using two-dimensional (2D) discrete Fourier transform (DFT). This 2D-DFT based method first converts input frames to gradient images, and then Tsai and Chiu \cite{TDF-1-1} applied the 2D-DFT on each spatial-temporal slice of the gradient image sequence and removed the vertical line pattern of the static backgrounds. In this way, this 2D-DFT based method can detect foreground objects without a training phase. However, the foreground masks contain only boundaries of moving objects and also present ringing around object boundaries. \\
\item \textbf{Discrete Cosine Transform (DCT) Features:} Porikli and Wren \cite{TDF-2} developed an algorithm called Wave-Back that generated a representation of the background using the frequency decompositions of pixel history. The Discrete Cosine Transform (DCT) coefficients are used as features are computed for the background and the current images. Then, the coefficients of the current image are compared to the background coefficients to obtain a distance map for the image. Then, the distance maps are fused in the same temporal window of the DCT to improve the
robustness against noise. Finally, the distance maps are thresholded to achieve foreground detection. This algorithm is efficient in the presence of waving trees. An other approach developed by Zhu et al. \cite{TDF-3} used a set of DCT-based features to exploit spatial and temporal correlation using a single Gaussian model. Thus, Zhu et al. \cite{TDF-3} used two features: the DC and the low frequency AC parameters. Each of which focuses on intensity and texture information, respectively. The AC feature parameter consists of the sum of the low frequency coefficients. The high frequency coefficients are not used because they are more sensitive to noise and they concern fine details that are more susceptible to small illumination changes. The use of these two features is equivalent to using both intensity and texture information, and produces more robust and reliable foreground detection masks. In an other approach, Wang et al. \cite{TDF-4} used only the information from DCT coefficients at block level to construct background models at pixel level. \cite{TDF-4} reports the implementation of running average, median and MOG in the DCT domain. Evaluation results show that these algorithms have much lower computational complexity in the DCT domain than in the spatial domain with the same accuracy. \\
\item \textbf{Wavelet Transform Features:} First, Huang and Hsieh \cite{WTDF-25}\cite{WTDF-26} proposed the use of Discrete Wavelet Transform (DWT) to obtain features that are used in a change detection based method for interframe-difference but DWT is not suitable for video applications as the use of DWT makes the method shift sensitive \cite{TDF-20}. In an other work, Gao et al. \cite{WTDF-1}\cite{WTDF-1-1}\cite{WTDF-1-2}\cite{WTDF-1-3} proposed a Marr  wavelet kernel and a background subtraction technique based on Binary Discrete Wavelet Transforms (BDWT). Thus, the  background and current frames are transformed in  the binary discrete wavelet domain, and background subtraction is performed in each sub-band. Experiments results \cite{WTDF-1} on several traffic video sequences show that this BDWT method produces better results with much lower computational complexity than the original MOG. To detect the moving objects more accurately, Hsia and Guo \cite{WTDF-9} proposed to use a modified directional lifting-based $9/7$ discrete wavelet transform (MDLDWT), which is based on the coefficient of lifting-based $9/7$ discrete wavelet transform (LDWT). Furthermore, to overcome that clear shape information of moving objects may not be available from multiple-level decomposition image such as LL3, Hsia and Guo \cite{WTDF-9} preserved the shape of objects in the low resolution image. Thus, the MDLDWT detects foreground moving objects in spatial domain. DLDWT not only retains the features of the flexibilities for multiresolution, but also achieves low computing cost when it is applied for LL-band images. MDLDWT based method presents the advantages of low critical path and fast computational speed. Moreover, the LL3-band is used solely to reduce the image transform computing cost and to remove noise. Experimental results \cite{WTDF-9} show that MDLDWT based method better retains slow motion of objects than DWT based method \cite{WTDF-25}\cite{WTDF-26}. In an other approach, Gao et al. \cite{WTDF-1-4} used orthogonal non-separable wavelet transformation for background modeling, and extracted the approximate information to reconstruct information frames. If the background present gradual changes, weighted superposition of multi background modeling images with time is applied to update the background. If the background presents sudden changes, the background is remodeled from this frame. In an other approach, Guan et al. \cite{WTDF-2}\cite{WTDF-2-1}\cite{WTDF-2-2}\cite{WTDF-2-3}\cite{WTDF-2-4} used wavelet multi-scale transform for detecting foreground moving objects and suppressing shadows. Practically, 2-D dyadic WT coefficients are obtained. Experimental results \cite{WTDF-2} show that the 2-D dyadic WT approach gives less false detections than the original MOG. Although, these wavelet based methods shown promising results \cite{WTDF-5}\cite{WTDF-10}, they are not adaptive in nature and tested against simple scenarios. Indeed,  it addressed  moving objects detection in presence of static backgrounds, but not effectively in presence of dynamic background changes. Practically, these previous wavelet based methods present various problems such as ghostlike appearance, object shadows, and noise. Also discrete real wavelet transform (DWT) presents shift-sensitivity \cite{WTDF-4}. To alleviate these weaknesses, Jalal and Singh \cite{WTDF-4}\cite{WTDF-4-1}\cite{WTDF-4-2} used Daubechies complex wavelet transform which is approximately shift-invariant and presents better directionality information compared with DWT. Thus, the noise resilience nature of wavelet domain is addressed, as the lower frequency sub-band of the wavelet transform presents the ability of a low-pass filter. So, Jalal and Singh \cite{WTDF-4}\cite{WTDF-4-1}\cite{WTDF-4-2} developed a background subtraction based on the low frequency sub-band characteristics of the object image, and exploited the local spatial coherence of the foreground objects to achieve a more robust foreground detection in presence camera jitter and illumination changes. In a similar way, Khare et al. \cite{WTDF-7-2} used a single change detection method based on Daubechies complex wavelet while Kushwaha and Srivastava \cite{WTDF-7-1} used a median filter to model the background.
In a further work, Khare et al. \cite{WTDF-7} used a double change detection method based on Daubechies complex wavelet coefficients of three consecutive frames. In an othe work, Kushwah and Srivastava \cite{WTDF-20}\cite{WTDF-20-1}\cite{WTDF-20-2} proposed a framework for dynamic background modeling based on Daubechies complex wavelet domain. First, wavelet decomposition of frame is obtained using the Daubechies complex wavelet transform. The change detection is then achieved by using detail coefficients (LH, HL, and HH), and the dynamic background model is obtained by an improved Gaussian mixture-based model applied on the approximate coefficient (LL subband). Experimental results \cite{WTDF-20-2} show that the Daubechies complex wavelet based method outperforms in presence of dynamic backgrounds the original codebook model \cite{CF-310}, the single change detection method based on Daubechies complex wavelet \cite{WTDF-7-2} and MDLDWT \cite{WTDF-9}.In an other approach, Mendizabal and Salgado \cite{WTDF-6} proposed to model the background at the region-levelin a wavelet based multi-resolution framework. Practically, the background model is obtained for each region independently as a mixture of $K$ Gaussian modes, by considering both the model of the approximation coefficients and the model of the detail coefficients at the different decomposition levels. Experimental results \cite{WTDF-6} show the robustness of this approach in presence of sudden illumination changes and strong shadows. All of the previous wavelet based methods are based on the same well-established moving object detection framework in which foreground objects are detected according to the differences of features between adjacent frames or between the current frame and background models.  Furthermore, the previous works only use various two-dimensional (2D) WTs to extract approximate coefficients and wavelet coefficients as features, in order to calculate differences between adjacent frames \cite{WTDF-30}\cite{WTDF-30-1}\cite{WTDF-30-2}\cite{WTDF-9}\cite{WTDF-7}\cite{WTDF-7-1}\cite{WTDF-7-2} or between the current frame and background models \cite{WTDF-1}\cite{WTDF-1-1}\cite{WTDF-1-2}\cite{WTDF-1-3} \cite{WTDF-2}\cite{WTDF-4}\cite{WTDF-4-1}\cite{WTDF-4-2}\cite{WTDF-6}. In this context, Han et al. \cite{WTDF-21} proposed a completely different method from the abovementioned framework of the other works and developed a background subtraction based on Three-Dimensional Discrete Wavelet Transform (3D-DWT). After analyzing frequency domain characteristics of the intensity temporal consistency of static backgrounds, the 3D-DWT based method decomposes the data cube built with a set of consecutive frames into multiple 3D DWT sub-bands, and then the relationship between static backgrounds and certain 3D-DWT sub-bands is established. Thus, the background and the foreground are separated in different 3D-DWT sub-bands. Practically, the 3D-DWT based method directly removes the background and retains the foreground by discarding sub-bands corresponding to the backgrounds. 3D-DWT presents the advantage of the frequency domain characteristics of intensity temporal consistency. Experimental results \cite{WTDF-21} show that the 3D-DWT based method rapidly produces accurate foreground masks in challenging situations lacking training opportunities and outperforms ViBe \cite{FS-51}, 2D-DFT \cite{TDF-1-1} and 2D-UWT \cite{WTDF-30}.\\
\item \textbf{Curvelet Transform Features:} Because the wavelet transform can not describe curve discontinuities and can present inaccurate segmentation of moving object due to non-removal of noise in consecutive frames, Khare et al. \cite{TDF-20} proposed to use curvelet transform. Thus, moving objects are detected with a change detection applied on curvelet coefficients of two consecutive frames. Experimental results \cite{TDF-20} show that the change detection based on curvelet transform is more robust to noise than the change detection with Discrete Wavelet Transform (DWT) \cite{WTDF-25}. \\
\item \textbf{Walsh Transform Features:} Tezuka and Nishitani \cite{TDF-30}\cite{TDF-31}\cite{TDF-32} modelled the background using the original MOG \cite{CF-1} applied on multiple block sizes obtained through the Walsh transform (WT). Walsh spectrum feature parameters are determined by using a set of coefficients from vertical, horizontal and diagonal directions, exhibiting strong spatial correlation among them. By using WT of the luminance component, four features are computed. The spectral nature of WT also reduces the computational steps required in feature extraction. Furthermore, Tezuka and Nishitani \cite{TDF-30}\cite{TDF-31}\cite{TDF-32}  developed a Selective Fast Walsh Transform (SFWT) with WT parameters consisting of only the low frequency coefficients. Experimental results \cite{TDF-30} show that the MOG with the Walsh Transform Features outperforms the MOG with RGB features \cite{CF-1} and the single Gaussian with DCT features \cite{TDF-3}. \\
\item \textbf{Hadamard Transform Features:} Baltieri et al. \cite{TDF-50} proposed a fast background initialization method designed at the block-level in a non-recursive manner to obtain the best background model using the least number of frames as possible. For this, each frame is split into blocks, producing a history of blocks and searching among them for the most reliable ones. In this last phase, the method works at a super-block level evaluating and comparing the spectral signatures of each block component. These spectral signatures are obtained using the Hadamard Transform which is faster than DCT. Experimental results \cite{TDF-50} demonstrate that this method outperforms its DCT counterpart \cite{TDF-10}. \\
\item \textbf{Slant Transform Features:} Haberdar and Shah  \cite{TDF-200}\cite{TDF-201} proposed a framework for detecting relevant changes
in dynamic background scenes. Practically, the changes are classified into two main classes called ordinary changes and relevant changes. Thus, this framework is based on a set of orthogonal linear transforms which allow to capture spatiotemporal features of local ordinary change patterns  and subsequently employ them in the detection of relevant changes. Three orthogonal linear transforms as the base transforms are used: \textbf{1)} discrete cosine transform (DCT), \textbf{2)} Walsh-Hadamard transform (WHT) and \textbf{3)} Slant transform (ST). Because DCT, WHT, and ST provide  complementary basis vectors, their combination allows to capture different types of ordinary change patterns. DCT is a sinusoidal transform that is widely used to obtain compact representations. WHT is a non-sinusoidal transform providing basis vectors that are rectangular or square wave. WHT can represent patterns with sharp discontinuities more accurately with fewer values than DCT. ST provided basis vectors derived from sawtooth wave-forms and are a good complement to WHT. Depending on the ordinary change patterns in each set, elements of the data sequence are assigned to one of these three base transforms. Experimental results \cite{TDF-200}\cite{TDF-201} on the ChangeDetection.net 2012 dataset that this approach is more robust in the dynamic background category than DPGMM \cite{TDF-200-1}, Spectral-360 \cite{TDF-200-2}, and \cite{TDF-200-3}. \\
\item \textbf{Gabor Features:} In the works of Xue et al. \cite{TDF-100}\cite{TDF-101}, the input image is first convolved with local Gabor filters so that each pixel has a group of features containing multiple amplitudes and corresponding phase values. Then, the features with the most effective phase information is selected according to the criteria that higher amplitude value in the feature group means more accurate local structure information has been captured, and its corresponding phase information is more representative. This phase feature is then defined as the sum of the selected phase values. This phase feature presents several advantages for background modeling and foreground detection. First, it is insensitive to illumination changes as it is an
inherent property of phase information. Second, the feature is relatively stable. Although noise exists in real videos, such feature change very little. Third, the wide value range makes it more suitable for background modeling and foreground detection. Finally, the feature is discriminative. When the true foreground appears in the scene, its value changes rapidly and drastically. For the bagkround model,  Xue et al. \cite{TDF-100}\cite{TDF-101} used the original MOG \cite{CF-1} and refined the foreground mask with blob aggregation using the Euclidean distance transform. Experimental results \cite{TDF-100}\cite{TDF-101} on I2R dataset \cite{900} that the MOG with the phase feature is more robust in presence of dynamic backgrounds and illumination changes than the original MOG \cite{CF-1}, KDE \cite{204} and LBP \cite{TF-10}. In an other approach, considering the advantages of Gabor filters which include the robustness to illumination changes,  Wei et al. \cite{TDF-110} adopted Gabor filters at multi-scale and multi-orientation to decompose an input video for sequential spatial feature extraction. Thus, a spatial feature vector (SFV) is computed for each pixel. Then, Wei et al. \cite{TDF-110} used the MOG model \cite{CF-1} on the SFVs. Experimental results \cite{TDF-110} on the Wallflower dataset \cite{500} show that this feature is more robust than the RGB one in the presence of illumination changes.\\
\end{itemize}

\subsection{Features from Video Domain Transform}
The primary purpose of pixel-level background modeling is to detect moving objects. However, in video coding and transmission applications, its use is to compress video data without degrading image quality and to transmit as less data as possible (only the data that have changed \cite{VC-1}\cite{VC-2}\cite{VC-3}\cite{VC-10}\cite{TDF-2110}\cite{VC-12}). Although features from the pixel domain are not effective for compression, several features from the video compressed domain are efficient for background/foreground separation. Furthermore, motion information is directly available without incurring cost of estimating a motion field. This set of features include: \\
\begin{itemize}
\item \textbf{Features in MPEG Domain:} Babu et al. \cite{TDF-2000} proposed to extract moving objects from MPEG compressed video by using motion vectors (MVs) which are sparse in MPEG. Thus, MVs are used for automatically estimating the number of objects and extracting independently moving objects. Thus, MVs are accumulated over few frames to enhance the motion information, and are further spatially interpolated to get dense motion vectors. The foreground mask is obtained by using an expectation maximization (EM) algorithm. To determine the number of motion models used in the EM step, Babu et al. \cite{TDF-2000} used a block-based affine clustering method, and the segmented objects are temporally tracked. Finally, precise object boundaries are obtained with an edge refinement. In an other approach, Zeng et al. \cite{TDF-2010} proposed
a moving object extraction which discriminated background and moving objects by means of the higher-order statistics applied on 
the inter-frame differences of Discrete Cosinus Transform (DCT) image. Thus, the DCT image is partially decoded from the compressed video for a rapid reconstruction of image data. The background is detected by the moment preserving thresholding technique. Based on the background statistic, an optimal threshold based on the background variance allows to  extract the final object mask by comparison the fourth moment measure and the variance. In a further work, Wang et al. \cite{TDF-2021} presented three algorithms (running average, median, MOG) which model the background  directly from compressed  video.  Each algorithm used DCT coefficients including AC coefficients at block level to represent background, and  adapted the background by updating DCT coefficients. The foreground mask is obtained with pixel accuracy through 
two-stage approach. Thus, the block regions fully or partially occupied by moving objects are identified in the DCT domain, and then pixels from these foreground blocks are further classified in the spatial domain. Experimental results \cite{TDF-2021} show the three algorithms achieved comparable accuracy to their counterparts in the  spatial domain. Furthermore, the  computational cost of the proposed median and MoG algorithms are only 40.4\% and 20.6\% of their counterparts in the spatial domain. In an other work, Porikli et al. \cite{TDF-2020}\cite{TDF-2021} exploited the macro-block structure to decrease the spatial resolution of the processed data, which exponentially reduces the computational time. Furthermore, Porikli et al. \cite{TDF-2020}\cite{TDF-2021} used temporal grouping of the intra-coded and estimated frames into a single feature layer. To achieve foreground detection, the DCT coefficients for I-frames and block motion vectors for P-frames are combined and a frequency-temporal data structure is constructed. From the blocks where the AC-coefficient energy and local inter-block DC-coefficient variance is small, the homogeneous volumes are enlarged by evaluating the distance of candidate vectors to the volume characteristics. Affine motion models are fit to volumes. 
Finally, the foreground mask is generated with a hierarchical clustering stage which iteratively merges the most similar parts. \\
\item \textbf{Features in H.264/AVC Domain:} Dey and Kundu \cite{TDF-2030} used the temporal statistics of feature vectors, describing macroblock (MB) units in each frame. Thus, feature vectors are used to select potential candidates containing moving objects. From the candidate macroblocks, foreground pixels are determined by comparing the colors of corresponding pixels pair-wise with a background model. This approach allows each macroblock to have a different quantization parameter, satisfying the requirements of both variable and fixed bit-rate applications. Additionally, a low-complexity technique for color comparison is used to obtain pixel-wise precision at a negligible computational cost as compared to classical approaches. A similar approach can be found in Pope et al. \cite{TDF-2040}. But, these methods \cite{TDF-2030}\cite{TDF-2040} depend only on the number of encoding bits and fail to detect motion/activity when the bitrate is severely constrained and rate-distortion optimization is enabled. To solve this problem, Dey and Kundu \cite{TDF-2031} proposed enhanced MB features which use both encoding bits as well as the quantization step-sizes of individual coefficients in a MB. This is particularly important as the quantization parameters vary widely between simple and complex sections of an encoded image at lower bitrates. Then, the method is a two-steps hybrid foreground detection where MBs covering prospective foreground regions are first identified. An adaptive thresholding technique selected the
candidate MBs corresponding to foreground objects. In the second step, pixels constituting the selected MBs are classified into background or foreground. Because the presence of shadows usually causes false classification, Dey and Kundu \cite{TDF-2031} estimated pixel differences
using the Luv color  instead of using the YCbCr color space \cite{TDF-2030}. Experimental results \cite{TDF-2031} shows that the enhanced MB approach with Luv color space outperforms both enhanced MB approach with YCrCb and the classical MB approach \cite{TDF-2030}. In an other approach, Tong et al. \cite{TDF-2050} first constructed the background model using the average MB RDCost over the initial $N$ successive frames. After the MB mode selection during encoding, each MB RDCost which is available is compared to the background model. If an MB has RDCost larger than a pre-defined threshold, it is regarded as motion MB. But, holes may appear inside large moving objects because MBs with less change in large moving objects have small RDCost, making those MBs to be misclassified as background. To solve this problem, Tong et al. \cite{TDF-2050} used spatial refinement. If more than half of the eight adjacent MBs of a background MB are moving, then the MB under consideration is regarded to be in motion. Further, moving objects are subjected to temporal refinement.  \\
\item \textbf{Features in HEVC Domain:} For surveillance video coding, the rate-distortion analysis shows that a larger Lagrange multiplier should be used if the background in a coding unit took a larger proportion. An adaptive Lagrange multiplier is better for rate distortion optimization. Thus, Zhao et al. \cite{TDF-2100} developed a background proportion adaptive Lagrange multiplier selection method based on HEVC. From the analysis of the relationship between the Lagrange multiplier and the background proportion, a Lagrange multiplier selection model for surveillance video is used. However, there is a large amount of static background regions which provide motion vectors equal to zero. Because motion search is very  time-consuming in the process of video coding, Zhao et al. \cite{TDF-2110} proposed a background-foreground division based search algorithm (BFDS) to accelerate the motion search in surveillance video coding by utilizing the background and foreground information of coding units. The idea is  to classify a predicting unit into a background predicting unit or a foreground predicting unit and then adopt different search strategies for each unit.\\
\end{itemize}

\subsection{Features from Compressive Cameras}
While the previous works performed background subtraction on compressed images, they do no addressed compressive features from cameras that record MPEG video directly. Thus, several works have been developed for Compressive Sensing (CS) imaging, and not compressed video files. Compressive features are then obtained by using a basis which provides a K-sparse representation. Practically, many different basis can provide sparse approximations of  images such as  wavelets, curvelets adn Gabor transformations. So, an image does not result in an exactly K-sparse representation and its transform coefficients decay exponentially to zero. In the literature, several approaches based on compressive features have been used and differ from the basis, the optimization method and the background model. \\

\indent First, Cevher et al. \cite{CS-1} used an orthonormal basis and a Basis Pursuit Denoising (BDP) algorithm for the minimization problem while Cevher et al. \cite{CS-2} used Lattice Matching Pursuit (LaMP) \cite{CS-2} in a further work. This algorithm needs a large amount of storage and computation for training the object silhouette, which is not suitable for real-time background/foreground separation. In an other work, Mota et al. \cite{CS-3}\cite{CS-4} proposed a  $l_1$-$l_1$ minimization. Needell and Tropp \cite{CS-10} developed an algorithm called (CoSaMP) while He et al.  \cite{CS-20} obtained compressive features through linear compressive measurements and used an improved CoSAMP algorithm called CoSaMP-Subspace. In an other work, Warnell et al. \cite{CS-30}\cite{CS-31} proposed two methods: one based on cross-validation measurements with an algorithm named (ARCS-CV), and a second one based low-resolution measurements with an algorithm named (ARCS-LR). In an other approach, Li et al. \cite{CS-40} employed an orthonormal wavelet basis, a linear programming technique for the minimization problem and the running average for the background model.  Wang et al. \cite{CS-50} developed a Gradient projection for sparse reconstruction (GPSR). Xu and Lu \cite{CS-60} proposed two approaches: one based on canonical sparsity basis, and the second based on wavelet sparsity basis with a K-cluster-valued CoSaMP algorithm. For linear compressive measurements, Davies et al. \cite{CS-70} compared the Basis Pursuit (BP) and Orthogonal Matching Pursuit (OMP) algorithms.  Wang et al. \cite{CS-90} used wavelets transform to obtain a sparse representation. Shah et al. \cite{CS-100} used  an algorithm called Convex Lattice Matching Pursuit (CoLaMP). For embedded camera networks, Shen et al. \cite{CS-101} developed two algorithms in the case of random projections basis called CS-MoG and Colour Space Compressed Sampling (CoSCS)-MoG with YCrCb components, respectively. Liu and Pados \cite{CS-102} used Walsh-Hadamard measurements ($l_1$-PCA). \\

\indent In the Robust Principal Component Analysis (RPCA) framework \cite{630}\cite{JournalCVIU2014-1}, several works used compressive sensing features too. Waters et al. \cite{RPCA-1}\cite{RPCA-2} proposed to recover the low-rank and sparse matrices from compressive measurements in method called SpaRCS for SPArse and low Rank decomposition via Compressive Sensing. In a further work, Kyrillidis and Cevher \cite{RPCA-2} provided a real-time implementation for SpaRCS while Ramesh and Shah \cite{RPCA-50} developed a Regularized version of the SpaRCS algorithm called R-SpaRCS. In an other approach, Zonoobi and Kassi \cite{RPCA-30} proposed a modification to the SpaRCS algorithm in order to incorporate priori-knowledge of the sparse component. In an other approach, Jiang et al. \cite{RPCA-10}\cite{RPCA-11} proposed a background subtraction based on low-rank and sparse decomposition by using the compressed measurement too. Although this model is suitable to the limited bandwidth of multimedia sensor networks, it is not enough robust to the movement turbulences and sudden illumination changes because the wavelet transform coefficients are not sparse in the case of turbulences. In a further work, Yang et al. \cite{RPCA-20} developed an adaptive CS-based algorithm which can exactly and simultaneously reconstruct the video foreground and background by using only 10\% of sampled measurements. However, it still uses the wavelet transform as Jiang et al. \cite{RPCA-10}\cite{RPCA-11}. This causes false detection in presence of turbulences and sudden illumination changes.  In an other work, Li and Qi \cite{RPCA-40} proposed a recursive Low-rank and Sparse estimation through Douglas-Rachford splitting (rLSDR) algorithm by recursively estimating low-rank and sparse components in the reconstructed video frames from CS measurements.  \\

\indent In a low-rank minimization (LRM) framework, Shu et al. \cite{LRM-1}\cite{LRM-2} proposed a  three-dimensional
compressive sampling (3DCS) approach to decrease the required sampling rate of the CI camera. So, a generic three-dimensional sparsity
measure(3DSM) is decoded a video from incomplete samples by exploiting its 3D piecewise smoothness and temporal low-rank property. Furthermore, a decoding algorithm is used for this 3DSM with guaranteed convergence. Experimental results \cite{LRM-1}\cite{LRM-2} show that the 3DCS based method requires a much lower sampling rate than the existing CS methods with the same accuracy. In an other work, Kang et al. \cite{LRM-10}\cite{LRM-11}  usee a three-dimensional circulant compressive sampling method to obtain sampled measurement, based on which the video foreground and background are reconstructed by solving an optimization problem. Experimental results \cite{LRM-10}\cite{LRM-11} on the I2R dataste demonstrate that this method outperforms both the original RPCA \cite{630} and the original MOG \cite{CF-1} in presence of dynamic backgrounds. 

\section{Multiple Characteristics}
\label{sec:MultipleCharacteristics}
The use of multiple characteristics consists to use additional values of the directed value of the concerned feature such as the mean, the median or high-order statistics. For example, Thongkamwitoon et al. \cite{CF-302}\cite{CF-303} and Amnuaykanjanasin et al. \cite{CF-304} used four tuples which contains the expected color vector, the color covariance matrix, the brightness distorsion and the color distorsion. An other representative approach is the codebook model in which minimum brightness, maximum brightness, frequency with which the code word has occurred and maximum negative run-length (MNRL) defined as the longest interval during the training period that the codeword has not reccured, and the first and last access times, respectively, that the codeword has occurred. So, the codebook model contains color (RGB \cite{CF-310}, YUV \cite{CF-341} and HSV \cite{CF-210} in the cylinder color model, YUV \cite{CF-343} in the spherical color model, and HSV in the hybrid cone cylinder color model \cite{CF-315}\cite{CF-316}), brightness and temporal information even if it uses color features. \\

\section{Multiple Features}
\label{sec:MultipleFeatures}
The use of multiple features (also called Bag-of-Features (BoF) \cite{EnF-1}\cite{EnF-2}\cite{EnF-3}) for background modeling has become a promising solution to improving robustness in real applications. The fundamental idea is to add spatial and/or temporal dimensions to the already existing spectral information available from the visual scene. The different features can be obtained from the same sensor (that is one camera) or from differents sensors such as IR cameras or RGB-D cameras. Table \ref{MFOverview-1}, Table \ref{MFOverview-2} and Table \ref{MFOverview-3} summarize different strategies based on multiple features in terms of number of features, fusion operators and background models. 

\subsection{Two features}
One of the popular choice of multiple features has been the combination of an alternative spatial feature to the already existing color features. Some common extensions of the color features to multiple features include:
\begin{itemize}
\item \textbf{Color-Gradient:} As the gradients of an image are relatively less sensitive to changes in illumination, some previous works have added it to the color feature to obtain quasi illumination invariant foreground detection. Jabri et al. \cite{EF-0} were the first authors who combined the intensity with the gradient features. The foreground detection mask was obtained by an union operator and the features were considered as independent. Holtzhausen et al. \cite{EF-60} used the same fusion scheme but with RGB features instead of intensity features. In an other way, Javed et al.  \cite{EF-1} used the RGB features with the gradient but with the AND operator. To take into account that the dependence within the features, Zang and Xu \cite{FA-10} used Ohta and gradient with the Sugeno integral, and Ding et al. \cite{FA-22} YCrCb and gradient with the Choquet integral which is more suitable than the Sugeno integral for the application of foreground detection. Approaches with fuzzy integrals are more robust than approaches with AND or OR operators as can be seen in \cite{FA-10}\cite{FA-22}. \\
\item \textbf{Color-Texture:} An other and one of the more common approaches of extending the feature space beyond color is the addition of texture features. It has been shown that the addition of texture to the color feature could improve robustness towards illumination changes and shadows as in \cite{TF-125}. For texture features, most of the authors used 1) the LBP \cite{TF-125}\cite{MulF-2} or one of its variants (ULBP \cite{TF-26},DLBP \cite{TF-60}), 2) the LTP \cite{TF-71}\cite{TF-72} or one of its variants (SILTP \cite{LF-10}\cite{TF-76}), or 3) statistical feature \cite{FA-33}\cite{FA-35} or fuzzy statistical features \cite{FF-10}\cite{FA-40}. In literature, different fusions scheme were used such as AND operator \cite{MulF-4}, weighted average \cite{TF-125}\cite{MulF-5}\cite{TF-26}\cite{TF-72} \cite{TF-76}, and fuzzy integrals (Sugeno integral \cite{FA-11}, Choquet integral \cite{FA-12}\cite{FA-14}\cite{FA-31}\cite{FA-32}\cite{FA-33}\cite{FA-34}\cite{FA-35}\cite{FF-10}, Interval valued Choquet Integral \cite{FA-40}). Even if there is no rigorous study of the best combination and fusion scheme, the YCrCb color space and LBP features aggregated with the Choquet integral \cite{FA-12}, and intensity and FST features aggregated with the Interval valued Choquet Integral \cite{FA-40} seem to be the more suitable and robust solutions for foreground detection in presence of illumination changes and dynamic backgrounds. \\
\item \textbf{Color-Depth:} With the increased use of RGB-D types of sensor, the combination of color together with depth information has emerged as a popular strategy for improved foreground detection. This combination of color and depth features has proven to deal with the camouflage in color and in some specific studies stereo features (disparity, depth) have also been used. Because depth features are different of color features as its distribution is different, the fusion scheme needs to be suitably chosen as the independence aspect \cite{SF-750}. The most representative works in color-depth are the ones of Camplani et al. \cite{SF-300}\cite{SF-301}\cite{SF-302}\cite{SF-303}, Fernandez-Sanchez \cite{SF-400}\cite{SF-401}, and Gallego and Pardas \cite{500}. More generally, different fusions scheme were used in literature as AND or OR operator \cite{SF-400}\cite{SF-401}, weighted average \cite{SF-300}\cite{SF-302}\cite{SF-303} and logarithmic opinion pools \cite{SF-500}. Most of the time, the color space is the RGB one \cite{SF-210}\cite{SF-300}\cite{SF-301}\cite{SF-302}\cite{SF-303}\cite{SF-400}\cite{SF-401}\cite{SF-500}\cite{SF-700} but HSV and LUV color spaces can be found in \cite{SF-705} and \cite{SF-720}, respectively. \\
\item \textbf{Color-Motion:} In addition to color information, the use of motion can allow the foreground detection to deal with unimportant movements such as in dynamic backgrounds. Zhou and Zhang \cite{MF-17} combined intensity with optical flow information. Gong and Chen \cite{MF-20} used both RGB color components and 2D motion vector obtained by optical flow in a 1-SVMs background model. The GPU implementation processes QVGA-sized video sequences at 39.3 fps on a laptop. Lin et al. \cite{MF-19-2}\cite{MF-19-3} used the optical flow value and the mean of inter-frame image difference as features in a probalistic SVM approach for background initialization. Martins et al.\cite{OT-10} used a bio-inspired hybrid segmentation which merges information from two inherently different methods: (1) bio-inspired motion detection method using a feature called Magno channel based from the modeling of the human visual system, and a background subtraction algorithm based on pixel color information. The foreground detection is obtained by merging with a logical AND the detection of the two methods.\\
\item \textbf{Color-Location:} Sheikh and Shah \cite{LF-1} proposed to use the location in addition to normalized RGB to exploit the dependency between pixel and thus spatial coherence can be taken into account. \\
\item \textbf{Texture-Motion:} Zhong et al. \cite{TF-19} used LBP and motion information obtained by a temporal operator (See Section \ref{sec:MotionFeatures}). \\
\end{itemize}

\subsection{More than two features}
The combination of multiple features, beyond two, seems useful particularly if the combination exploits spectral, spation and temporal information. However, such larger combination of features requires careful consideration to avoid redundancy that could impact computational demand and also preserve discriminative ability of features to ensure wide range of adaptation capbilities. In literature, there are different combination of three features: \textit{1)} color-edge-texture  \cite{FA-30}, \textit{2)} color-edge location \cite{EnF-1}, \textit{3} color-texture-motion \cite{EnF-2}, and \textit{4)} color-texture-location \cite{TF-210}. Thus, Azab et al. \cite{FA-30} aggregated three features, i.e color (RGB), edge (obtained with the Sobel operator) and texture (LBP) using the Choquet integral to deal with illumination changes and dynamic backgrounds. The Choquet integral is used in aggregating color, edge and texture confidence maps. Experimental results of \cite{FA-30} on the PETS 2006 dataset showed that the addition of the edge feature increases the performance of the similar method  \cite{FA-12} which used only the color and the texture features.\\

\indent More than three features can be found in ensemble of features based approaches \cite{FA-2}\cite{FR-20}\cite{FS-20}, feature selection schemes (See Section \ref{sec:FeatureSelection}) or bag-of-features approaches  \cite{EnF-1}\cite{EnF-2}\cite{EnF-3}. For example, Klare and Sarkar \cite{FA-2} proposed an algorithm that incorporates multiple instantiations of the MOG algorithm with 13 features including: \textbf{\textit{1)}} color features (RGB), and \textbf{\textit{2)}} edge features which are the gradient and magnitude obtained with a Canny edge detector, eight texture feature (Haar features). The fusion method used is the average rule. Similarly, Han and Davis \cite{FR-20} performed background subtraction using a Support Vector Machine over background likelihood vectors for a set of features which consist of 11 features: color featurees (RGB), gradient (horizontal and vertical) and six Haar-like features. Then, Han and Davis \cite{FR-20} used a SVM classifier over the background probability vectors for the feature set. The aim of integrating the classifier for foreground/background segmentation is the selection of the discriminative features. Moreover, it also reduced the feature dependency problem. Otherwise, highly correlated non-discriminative features may dominate the classification process regardless of the states of other features.  A radial basis function kernel is used to deal with non-linear input data. As Han and Davis \cite{FR-20} trained the classifier based on probability vectors rather than feature vectors directly, a universal SVM is used for all sequences and not a separate SVM for each pixel nor for each sequence. \\

\indent Lopez-Rubio and Lopez-Rubio \cite{FS-20} developed a method based on stochastic optimization \cite{FS-20-1} which overcomes the limitations of some features along with a set of relevant features that yields adequate results. Thus, a probabilistic model, in addition to handling any number of pixel features, could also account for the correlations among the features, so that a more realistic model is obtained. This method is called Multiple Feature Background Model (MFBM). Since the number of features $N_f$ is not restricted, a fast and numerically stable implementation is of great importance to the practical applicability of MFBM. The slowest part of MFBM is the computation of the Gaussian probability density. It is $O(N_f)$ due to the inversion of the covariance matrix, and the computation of its determinant \cite{FS-20}. The other equations are $O(N_f)$ or lower. The following 24 features are used: \textbf{\textit{1)}} Normalized RGB are used due to their robustness with respect to illumination changes, \textbf{\textit{2)}} six Haar-like features which are robust features that convey texture information, \textbf{\textit{3)}} previous features $1$ to $9$ processed with the bidimensional median filter of  window size $5\times5$ pixels, \textbf{\textit{4)}} Gradient features estimated with Sobel operator  are less affected by illumination changes while they provide local texture information, \textbf{\textit{5)}} two features which consider both color information and a pixel adjacent to the pixel at hand, as a local indication of color texture, and \textbf{\textit{6)}} two small filters of size $3\times3$ pixels to extract the texture information. Experimental results \cite{FS-20} on several large scale dataset (I2R dataset \cite{900}, ChangeDetection.net \cite{901}, BMC 2012 dataset \cite{903}) tested  all possible combinations of pairs of features and triples of the $24$ features. The conclusion is that Haar-like features are not well suited for MFBM which needs few features with the best possible discriminative power. The performance of MFBM increases as more relevant features are added, but the improvement decreases until a point where inserting more features is useless ($N_f=5$). Normalized color channels and median filtered feature present particularly good results. This is due to their robustness against illumination changes for normalized features and their ability to filter the background noise for median filtered features. State-of-the-art background modeling approaches are outperformed by MFBM method but it seems that is due to the number and kind of features and not due to the model itself. In a further work, Molina-Cabello et al. \cite{FS-21} used the same set of features with a background subtraction method called Features based FSOM (FFSOM) which is an extension of Foreground Self-Organizing Maps (FSOM) \cite{FS-21-1}.\\

\indent In Subudhi et al. \cite{EnF-3}, six local features are taken into consideration: \textit{1)} Three existing features: brightness (Bs), inverse contrast ratio (ICR) and average sharpness (AS)), and \textit{2)} three proposed features: absolute relative height (ARH), local weighted variance (LWV) and integrated modal variability (IMV)). As most of the background subtraction methods consider non bi-unique model (which is the fact that local changes corresponding to the moving objects are obtained by making a combination of multiple features rather than combination of decision on individual features), Subudhi et al. \cite{EnF-3} suggested the use of a bi-unique model as in Bovolo et al. \cite{EnF-3-1} where individual spectral properties are combined or results of the spectral channels are fused to obtain better results.Experimental resultson the ChangeDetection.net dataset \cite{901} show the robustness of this method in the categories "Dynamic Backgrounds", "Camera Jitter" and "Thermal". \\

\indent Wang and Wan \cite{EnF-5} used 10 features: three in color (RGB components), six in gradient computed via Prewitt operator, and one in Gabor filter \cite{EnF-5-1}. Then, Wang and Wan \cite{EnF-5} proposed a Multi-Task Robust Principal Component Analysis (MTRPCA) model which integrate multi-feature jointly into the RPCA framework \cite{630}. Experimental results on the ChangeDetection.net dataset \cite{901} show that the MTRPCA with all the features outperforms MTRPCA with color and gradient, RPCA with color, and RPCA with gradient. Gan et al. \cite{EnF-4} proposed a similar method called Multi-feature Robust Principal Component Analysis (MFRPCA). \\

\indent Exploiting multiple cues, Huerta et al. \cite{MulF-30} first combined both intensity and color (chromatic and brightness
distortion) features in order to solve some of the color motion segmentation problems such as saturation or the lack of the color when the background model is constructed. Nonetheless, some colors problems are still unsolved such as dark and light camouflage. Then, to solve these problems, Huerta et al. \cite{MulF-30} used edge features obtained with the Sobel operator. Finally, intensity, color and edge cues are combined using the logical operators AND and OR. In a futher work, Huerta et al. \cite{MulF-31} improved these multiple cues approach by using a chromatic-invariant cone model for the color features. \\

\newpage
\begin{landscape}
\begin{table}
\scalebox{0.79}{
\begin{tabular}{|l|l|l|l|l|l|} 
\hline
\scriptsize{Features}           &\scriptsize{Categories (number of papers)} &\scriptsize{Background Model}    &\scriptsize{Fusion (number of papers)}    &\scriptsize{Dependence-independence} &\scriptsize{Authors - Dates} \\
\hline
\hline
\scriptsize{Two Features}       &\scriptsize{\textbf{Color-Edge (5)}}         &\scriptsize{}                    &\scriptsize{}          
&\scriptsize{}                  &\scriptsize{}   \\
\scriptsize{}                   &\scriptsize{Intensity-Gradient}              &\scriptsize{-}               &\scriptsize{Union (Foreground, Boundary)(1)} 
&\scriptsize{Independent}       &\scriptsize{Jabri et al. (2000) \cite{EF-0}}   \\
\scriptsize{}                   &\scriptsize{RGB-Gradient}                &\scriptsize{-}                    &\scriptsize{AND (Foreground, Boundary) (1)} &\scriptsize{Independent}       &\scriptsize{Javed et al. (2002) \cite{EF-1}}   \\
\scriptsize{}                   &\scriptsize{RGB-Gradient}                &\scriptsize{-}                    &\scriptsize{Union (Foreground, Boundary)(1)}   
&\scriptsize{Independent}       &\scriptsize{Holtzhausen et al. (2012) \cite{EF-60}}   \\
\scriptsize{}                   &\scriptsize{RGB-HOG}                     &\scriptsize{MOG \cite{StF-131}}   &\scriptsize{AND, OR}       &\scriptsize{Independent}        &\scriptsize{Mukherjee  et al. (2010) \cite{StF-131}}   \\
\scriptsize{}                   &\scriptsize{Ohta-Gradient}               &\scriptsize{-}                    &\scriptsize{Sugeno Integral (1)}        
&\scriptsize{Dependent}         &\scriptsize{Zhang and  Xu (2006) \cite{FA-10}}   \\
\scriptsize{}                   &\scriptsize{YCrCb-Gradient}              &\scriptsize{-}                    &\scriptsize{Choquet Integral (1)}       &\scriptsize{Dependent}         &\scriptsize{Ding et al. (2009) \cite{FA-22}}   \\
\cline{2-6}
\scriptsize{}                   &\scriptsize{\textbf{Color-Texture (34)}} &\scriptsize{}                     &\scriptsize{}          
&\scriptsize{}                  &\scriptsize{}   \\
\scriptsize{}                   &\scriptsize{LCP-LBP}                     &\scriptsize{-}                    &\scriptsize{Weigthed Average (2)}       &\scriptsize{Independent}      &\scriptsize{Chua et al. (2012) \cite{TF-125}}   \\
\scriptsize{}                   &\scriptsize{Intensity-SKR Texture}       &\scriptsize{-}                    &\scriptsize{Boundary Curvature (1)}     &\scriptsize{Independent}      &\scriptsize{Guo and  Yu (2012) \cite{MulF-1}}   \\
\scriptsize{}                   &\scriptsize{RGB-BM}                      &\scriptsize{-}                    &\scriptsize{Conditions (Foreground) (1)}  &\scriptsize{Independent} &\scriptsize{Lai et al. (2013) \cite{MulF-2}} \\
\scriptsize{}                   &\scriptsize{Photometric Invariant Color-LBP}    &\scriptsize{-}             &\scriptsize{Boundary Texture (1)}       &\scriptsize{Independent}      &\scriptsize{Shang et al. (2012) \cite{MulF-3}} \\
\scriptsize{}                   &\scriptsize{RGB-LBP}                     &\scriptsize{-}                    &\scriptsize{AND (Foreground) (1)}      
&\scriptsize{Independent}       &\scriptsize{Lin et al. (2010) \cite{MulF-4}} \\
\scriptsize{}                   &\scriptsize{RGB-LBP}          &\scriptsize{K Multi-layers \cite{MulF-5} }  &\scriptsize{Weigthed Average (1)}      &\scriptsize{Independent}       &\scriptsize{Jian and Odobez (2007) \cite{MulF-5}} \\
\scriptsize{}                   &\scriptsize{Intensity-SCS-LTP}           &\scriptsize{MOG \cite{CF-1}, Pattern Adaptive KDE \cite{TF-71}}            &\scriptsize{Product Likelihoods Foreground (1)}                          &\scriptsize{Independent}    &\scriptsize{Zhang et al. (2011) \cite{TF-71}} \\
\scriptsize{}                   &\scriptsize{Intensity-SCS-LTP}           &\scriptsize{MOG \cite{CF-1}}      &\scriptsize{Weighted Average (1)}       &\scriptsize{Independent}      &\scriptsize{Zhang et al. (2012) \cite{TF-72}} \\
\scriptsize{}                   &\scriptsize{HSV-ULBP}     &\scriptsize{K Histograms \cite{TF-10}}      &\scriptsize{Weighted Average (1)}          
&\scriptsize{Independent}       &\scriptsize{Yuan et al. (2011) \cite{TF-26}} \\
\scriptsize{}                   &\scriptsize{Color-DLBP}     &\scriptsize{-}      &\scriptsize{-}  
&\scriptsize{Independent}       &\scriptsize{Xu et al. (2009) \cite{TF-60}} \\
\scriptsize{}                   &\scriptsize{RGB-ST-SILTP}     &\scriptsize{K Histograms}      &\scriptsize{Weighted Average (1)}  
&\scriptsize{Independent}       &\scriptsize{Ji and Wang (2014) \cite{TF-76}} \\
\scriptsize{}                   &\scriptsize{Lab-SILTP}                   &\scriptsize{KDE \cite{LF-10}}    &\scriptsize{ - (3)}         &\scriptsize{Independent}       &\scriptsize{Narayana et al. (2012) \cite{LF-10}}   \\
\scriptsize{}                   &\scriptsize{Normalized RGB-SILTP}        &\scriptsize{MOG \cite{CF-1}}    &\scriptsize{ - (1)}         &\scriptsize{Independent}       &\scriptsize{Shi and Liu (2016) \cite{TF-129-20}}   \\
\scriptsize{}                   &\scriptsize{Brightness-Chromaticity-Texture Variations} &\scriptsize{Conditional Probability Densities \cite{MulF-9}} &\scriptsize{ - (1)}         &\scriptsize{Independent}       &\scriptsize{Wang et al. (2016) \cite{MulF-9}}   \\
\scriptsize{}                   &\scriptsize{Ohta-LBP}                    &\scriptsize{-}                   &\scriptsize{Sugeno Integral (1)}        &\scriptsize{Dependent}         &\scriptsize{Zhang and  Xu (2006) \cite{FA-11}}   \\
\scriptsize{}                   &\scriptsize{YCrCb-LBP}                   &\scriptsize{-}                   &\scriptsize{Choquet Integral (3)}
&\scriptsize{Dependent}         &\scriptsize{El Baf et al. (2008) \cite{FA-12}}   \\
\scriptsize{}                   &\scriptsize{IR-LBP}                       &\scriptsize{-}                  &\scriptsize{Choquet Integral (1)}       &\scriptsize{Dependent}         &\scriptsize{El Baf et al. (2008) \cite{FA-14}}   \\
\scriptsize{}                   &\scriptsize{HSI-LBP}                      &\scriptsize{-}                  &\scriptsize{Choquet Integral (4)}       &\scriptsize{Dependent}         &\scriptsize{Ding et al. (2009) \cite{FA-21}}   \\
\scriptsize{}                   &\scriptsize{RGB-LBP}                      &\scriptsize{-}                  &\scriptsize{Choquet Integral (1)}       &\scriptsize{Dependent}         &\scriptsize{Lu et al. (2012) \cite{FA-31}}   \\
\scriptsize{}                   &\scriptsize{CrCb-LBP}                   &\scriptsize{-}                   &\scriptsize{Choquet Integral (1)}       
&\scriptsize{Dependent}         &\scriptsize{Balcilar et al. (2013) \cite{FA-32}}   \\
\scriptsize{}                   &\scriptsize{YCrCb-ULBP}                   &\scriptsize{-}                 &\scriptsize{Choquet Integral (1)}       
&\scriptsize{Dependent}         &\scriptsize{Lu et al. (2014) \cite{FA-34}}   \\
\scriptsize{}                   &\scriptsize{Intensity-ST}                   &\scriptsize{-}               &\scriptsize{Choquet Integral (1)}     &\scriptsize{Dependent}         &\scriptsize{Chiranjeevi et al. (2014) \cite{FA-33}}   \\
\scriptsize{}                   &\scriptsize{Intensity-ST}                   &\scriptsize{-}               &\scriptsize{Choquet Integral (1)}     &\scriptsize{Dependent}         &\scriptsize{Gayathri et al. (2014) \cite{FA-35}}   \\
\scriptsize{}                   &\scriptsize{Intensity-FST}                   &\scriptsize{-}              &\scriptsize{Choquet Integral (1)}     &\scriptsize{Dependent}         &\scriptsize{Chiranjeevi and Sengupta  (2012) \cite{FF-10}}   \\
\scriptsize{}                   &\scriptsize{Intensity-FST}                   &\scriptsize{-}              &\scriptsize{Interval valued Choquet Integral (1)}     
&\scriptsize{Dependent}         &\scriptsize{Chiranjeevi and Sengupta  (2016) \cite{FA-40}}   \\
\cline{2-6}
\scriptsize{}                   &\scriptsize{\textbf{Color-Stereo Features (Part 1) (30)}} &\scriptsize{}         &\scriptsize{} 
&\scriptsize{}                  &\scriptsize{}   \\
\scriptsize{}                   &\scriptsize{RGB-Disparity}                   &\scriptsize{Fist Image}       &\scriptsize{AND (Background), AND (Foreground) (3)}               &\scriptsize{Independent}  &\scriptsize{Ivanov et al. (1998) \cite{SF-1}}  \\
\scriptsize{}                   &\scriptsize{RGB-Disparity}                     &\scriptsize{MOG \cite{CF-1}} &\scriptsize{OR (Foreground) (1)}   
&\scriptsize{Independent}       &\scriptsize{Gordon et al. (1999)\cite{SF-11}}  \\
\scriptsize{}                   &\scriptsize{YUV-Depth}                         &\scriptsize{MOG \cite{CF-1}} &\scriptsize{OR (Foreground)(3)}    
&\scriptsize{Independent}       &\scriptsize{Harville et al. (2001)  \cite{SF-100}}  \\
\scriptsize{}                   &\scriptsize{Intensity-Depth}                   &\scriptsize{MOG \cite{CF-1}} &\scriptsize{OR (Foreground) (1)}    
&\scriptsize{Dependent}         &\scriptsize{Song et al. (2014)  \cite{SF-103}}  \\
\scriptsize{}                   &\scriptsize{Intensity-Depth}                   &\scriptsize{MOG \cite{CF-1}} &\scriptsize{AND (Foreground) +  Conditions (1)}    
&\scriptsize{Dependent}         &\scriptsize{Silvestre (2007)  \cite{SF-200}}  \\
\scriptsize{}                   &\scriptsize{Intensity-Depth}                   &\scriptsize{KDE \cite{204}} &\scriptsize{AND (Foreground) (1)}    
&\scriptsize{Dependent}         &\scriptsize{Silvestre (2007)  \cite{SF-200}}  \\
\scriptsize{}                   &\scriptsize{YCrCb-Depth-Modulation}            &\scriptsize{MOG \cite{CF-1}} &\scriptsize{OR (Foreground) (2)}    
&\scriptsize{Dependent}         &\scriptsize{Langmann et al. (2010)  \cite{SF-201}}  \\
\scriptsize{}                   &\scriptsize{IR-Depth}                          &\scriptsize{MOG \cite{CF-1}} &\scriptsize{Boundary Depth (1)}    
&\scriptsize{Independent}       &\scriptsize{Stormer et al. (2010) \cite{SF-203}}  \\
\scriptsize{}                   &\scriptsize{Intensity-Depth}                   &\scriptsize{ViBe \cite{FS-51}} &\scriptsize{OR (Foreground) (1)}    
&\scriptsize{Independent}       &\scriptsize{Leens et al. (2009) \cite{SF-204}}  \\
\hline
\end{tabular}}
\caption{Multiple Features: An Overview (Part 1).} \centering
\label{MFOverview-1}
\end{table}
\end{landscape}

\newpage
\begin{landscape}
\begin{table}
\scalebox{0.75}{
\begin{tabular}{|l|l|l|l|l|l|} 
\hline
\scriptsize{Features}           &\scriptsize{Categories}                  &\scriptsize{Background Model}    &\scriptsize{Fusion (number of papers)}  &\scriptsize{Dependence-independence} &\scriptsize{Authors - Dates} \\
\hline
\hline
\scriptsize{Two features}       &\scriptsize{\textbf{Color-Stereo Features (Part 2)}}      &\scriptsize{}                   &\scriptsize{}          
&\scriptsize{}                  &\scriptsize{}   \\
\scriptsize{}                   &\scriptsize{RGB-Depth}                         &\scriptsize{-}               &\scriptsize{Weighted average (1)}    
&\scriptsize{Independent}       &\scriptsize{Hu et al. (2014) \cite{SF-210}}  \\
\scriptsize{}                   &\scriptsize{RGB-Depth}                          &\scriptsize{MOG \cite{CF-1}} &\scriptsize{Weighted average (1)}    
&\scriptsize{Independent}       &\scriptsize{Camplani and Salgado (2013) \cite{SF-300}}  \\
\scriptsize{}                   &\scriptsize{RGB-Depth}                          &\scriptsize{MOG \cite{CF-1}} &\scriptsize{Combination (1)}    
&\scriptsize{Independent}       &\scriptsize{Camplani et al. (2013) \cite{SF-301}}  \\
\scriptsize{}                   &\scriptsize{RGB-Depth}                          &\scriptsize{MOG \cite{CF-1}} &\scriptsize{Weighted average (1)}    
&\scriptsize{Independent}       &\scriptsize{Camplani et al. (2013) \cite{SF-302}}  \\
\scriptsize{}                   &\scriptsize{RGB-Depth}          &\scriptsize{2 MOG Temporal (Background)}    &\scriptsize{Weighted average (1)}    
&\scriptsize{Independent}       &\scriptsize{Camplani et al. (2013) \cite{SF-303}}  \\
\scriptsize{}                   &\scriptsize{}                  &\scriptsize{2 MOG Spatial (Foreground) }     &\scriptsize{}    
&\scriptsize{}                  &\scriptsize{}  \\
\scriptsize{}                   &\scriptsize{RGB-Depth (Approach 1)}           &\scriptsize{Codebook \cite{302}} &\scriptsize{AND (Background) (1)}  
&\scriptsize{Independent}       &\scriptsize{Fernandez-Sanchez et al. (2013) \cite{SF-400}}  \\
\scriptsize{}                   &\scriptsize{RGB-Depth (Approach 2)}          &\scriptsize{Codebook \cite{302}}  &\scriptsize{AND (Background) (1)}  
&\scriptsize{Dependent}         &\scriptsize{Fernandez-Sanchez et al. (2013) \cite{SF-400}}  \\
\scriptsize{}                   &\scriptsize{RGB-Depth}                      &\scriptsize{Codebook \cite{302}}   &\scriptsize{AND (Background) (1)}  
&\scriptsize{Dependent}         &\scriptsize{Fernandez-Sanchez et al. (2014) \cite{SF-401}}  \\
\scriptsize{}                   &\scriptsize{RGB-Depth}                       &\scriptsize{2 MOG Temporal (Background)}                &\scriptsize{Logarithmic Opinion Pool (1)}  
&\scriptsize{Independent}       &\scriptsize{Gallego and Pardas (2013) \cite{SF-500}}  \\
\scriptsize{}                   &\scriptsize{}                  &\scriptsize{2 MOG Spatial (Foreground) }      &\scriptsize{}    
&\scriptsize{}                  &\scriptsize{}  \\
\scriptsize{}                   &\scriptsize{RGB-Depth}                       &\scriptsize{MOG \cite{912}}     &\scriptsize{Combination (2)}    
&\scriptsize{Independent}       &\scriptsize{Ottonelli et al. (2013) \cite{SF-700}}  \\
\scriptsize{}                   &\scriptsize{xy-HSV-Depth}                    &\scriptsize{KDE \cite{SF-705}}  &\scriptsize{}    
&\scriptsize{}                   &\scriptsize{Spampinato et al. (2014) \cite{SF-705}}  \\
\scriptsize{}                   &\scriptsize{Color-Depth}                     &\scriptsize{MOG \cite{CF-1}}     &\scriptsize{Two probabilistic models (1)}    &\scriptsize{Independent}       &\scriptsize{Song et al. (2014) \cite{SF-710}}  \\
\scriptsize{}                   &\scriptsize{LUV-Depth}                         &\scriptsize{MOG \cite{CF-1}}   &\scriptsize{-}   
&\scriptsize{Independent}       &\scriptsize{Liang et al. (2016) \cite{SF-720}}  \\
\scriptsize{}                   &\scriptsize{LUV-Depth}                         &\scriptsize{FuzzyAdaptiveSOM \cite{CF-203-2}}   &\scriptsize{-}    
&\scriptsize{Independent}       &\scriptsize{Liang et al. (2016) \cite{SF-720}}  \\
\scriptsize{}                   &\scriptsize{RGB-Depth}                         &\scriptsize{KDE \cite{204}}    &\scriptsize{- (1)}    
&\scriptsize{Independent}       &\scriptsize{Moya-Alcover et al. (2016) \cite{SF-730}}  \\
\scriptsize{}                   &\scriptsize{Colorimetric Invariant-Depth}      &\scriptsize{-}   &\scriptsize{- (1)}    
&\scriptsize{Independent}       &\scriptsize{Murgia et al. (2014) \cite{SF-790}}  \\
\scriptsize{}                   &\scriptsize{RGB-Depth}      &\scriptsize{GMM \cite{CF-1}}   &\scriptsize{- (1)}    
&\scriptsize{Independent}       &\scriptsize{Amamra et al. (2014) \cite{SF-791}}  \\
\cline{2-6}
\scriptsize{}                   &\scriptsize{\textbf{Color-Motion (5)}}      &\scriptsize{}                   &\scriptsize{}          
&\scriptsize{}                  &\scriptsize{}   \\
\scriptsize{}                   &\scriptsize{Intensity-Optical Flow}      &\scriptsize{MOG \cite{CF-1}}       &\scriptsize{Conditions (1)}       &\scriptsize{}                  &\scriptsize{Zhou and Zhang \cite{MF-17}}   \\
\scriptsize{}                   &\scriptsize{RGB-2D motion vector}        &\scriptsize{1-SVM}                 &\scriptsize{Weighted average (1)}     &\scriptsize{Dependent}         &\scriptsize{Gong and Cheng  (2011) \cite{MF-20}}   \\
\scriptsize{}                   &\scriptsize{Interframe Difference-Optical Flow}      &\scriptsize{SVM}       &\scriptsize{-(2)}       &\scriptsize{Dependent}         &\scriptsize{Lin et al. (2002) \cite{MF-20}}   \\
\scriptsize{}                   &\scriptsize{Color-Magno Channel}      &\scriptsize{MOG \cite{CF-1}}       &\scriptsize{AND (1)}       &\scriptsize{Independent}         &\scriptsize{Matins et al. (2016) \cite{OT-10}}   \\
\cline{2-6}
\scriptsize{}                   &\scriptsize{\textbf{Color-Location (1)}}      &\scriptsize{}                 &\scriptsize{}          
&\scriptsize{}                  &\scriptsize{}   \\
\scriptsize{}                   &\scriptsize{Normalized RGB-(x,y)}      &\scriptsize{KDE \cite{LF-1}}     &\scriptsize{ - (2)}          
&\scriptsize{}                  &\scriptsize{Sheikh and Shah \cite{LF-1}}   \\
\cline{2-6}
\scriptsize{}                   &\scriptsize{\textbf{Color-Intensity (2) }}                 &\scriptsize{}    &\scriptsize{}          
&\scriptsize{}                  &\scriptsize{}   \\
\scriptsize{}                   &\scriptsize{RGB-Intensity (Multi-cues)}                    &\scriptsize{}    &\scriptsize{AND-OR (1)}        
&\scriptsize{}                  &\scriptsize{Huerta et al. (2007) \cite{MulF-32}}   \\
\scriptsize{}                   &\scriptsize{Color invariant H-Intensity (Multimodal)}      &\scriptsize{}    &\scriptsize{- (1)}        
&\scriptsize{}                  &\scriptsize{Cocorullo et al. (2016) \cite{MulC-1}}   \\
\cline{2-6}
\scriptsize{}                   &\scriptsize{\textbf{Intensity-IR (11)}}        &\scriptsize{}    &\scriptsize{}          
&\scriptsize{}                  &\scriptsize{}   \\
\scriptsize{}                   &\scriptsize{Intensity - IR (Multi-sensors)}   &\scriptsize{}    &\scriptsize{Contour Saliency (4)}        
&\scriptsize{}                  &\scriptsize{Davis and Sharma (2005) \cite{IF-13}}   \\
\scriptsize{}                   &\scriptsize{Intensity - IR (Multi-sensors)}   &\scriptsize{}    &\scriptsize{Physics-based fusion (1)}        
&\scriptsize{}                  &\scriptsize{Nadimi and Bhanu (2003) \cite{IF-30}}   \\
\scriptsize{}                   &\scriptsize{Intensity - IR (Multi-sensors)}   &\scriptsize{}    &\scriptsize{Evolutionary-based fusion (2)}        
&\scriptsize{}                  &\scriptsize{Nadimi and Bhanu (2004) \cite{IF-31}}   \\
\scriptsize{}                   &\scriptsize{Intensity - IR (Multi-sensors)}   &\scriptsize{}    &\scriptsize{Transferable Belief Model (3)}        
&\scriptsize{}                  &\scriptsize{Conaire et al.  (2006) \cite{IF-32}}   \\
\scriptsize{}                   &\scriptsize{Intensity - IR (Multi-sensors)}   &\scriptsize{}    &\scriptsize{Joint Sample Consensus Model (1)}     &\scriptsize{}                  &\scriptsize{Han et al. (2013) \cite{IF-40}}   \\
\cline{2-6}
\scriptsize{}                   &\scriptsize{\textbf{Color-IR (1)}}        &\scriptsize{}    &\scriptsize{}          
&\scriptsize{}                  &\scriptsize{}   \\
\scriptsize{}                   &\scriptsize{RGB - IR (Multi-sensors)}   &\scriptsize{}    &\scriptsize{- (1)}        
&\scriptsize{}                  &\scriptsize{Pejcic et al. (2009) \cite{TF-179-1}}   \\
\scriptsize{}                   &\scriptsize{\textbf{Texture-Motion (1)}}     &\scriptsize{}                  &\scriptsize{}          
&\scriptsize{}                  &\scriptsize{}   \\
\scriptsize{}                   &\scriptsize{LBP-Temporal Operator}      &\scriptsize{}                   &\scriptsize{Weighted average (1)}        
&\scriptsize{}                  &\scriptsize{Zhong et al. (2008) \cite{TF-19}}   \\
\cline{2-6}
\scriptsize{}                   &\scriptsize{\textbf{Edge-Texture (1)}}     &\scriptsize{}                  &\scriptsize{}          
&\scriptsize{}                  &\scriptsize{}   \\
\scriptsize{}                   &\scriptsize{Gradient-LPTP \cite{OT-40-1}}     &\scriptsize{}                   &\scriptsize{- (1)}        
&\scriptsize{Dependent}         &\scriptsize{Kim et al. (2015) \cite{OT-40}}   \\
\cline{2-6}
\scriptsize{}                   &\scriptsize{\textbf{Texture-Location (1)}}      &\scriptsize{}                 &\scriptsize{}          
&\scriptsize{}                  &\scriptsize{}   \\
\scriptsize{}                   &\scriptsize{LBP-(x,y)}      &\scriptsize{-}     &\scriptsize{Weighted average likelihood (1)}          
&\scriptsize{}                  &\scriptsize{Rodrigues et al. \cite{MulF-100}}   \\
\scriptsize{}                   &\scriptsize{\textbf{Texture-Local Gradient Histograms (1)}}    &\scriptsize{}         &\scriptsize{}       
&\scriptsize{}                  &\scriptsize{}   \\
\scriptsize{}                   &\scriptsize{LBP-HOG)}                               &\scriptsize{MOG \cite{CF-1}}     &\scriptsize{- (1)}          
&\scriptsize{}                  &\scriptsize{Panda and  Meher \cite{StF-132}}   \\
\hline
\end{tabular}}
\caption{Multiple Features: An Overview (Part 2).} \centering
\label{MFOverview-2}
\end{table}
\end{landscape}

\newpage
\begin{landscape}
\begin{table}
\scalebox{0.75}{
\begin{tabular}{|l|l|l|l|l|l|} 
\hline
\scriptsize{Features}           &\scriptsize{Categories}                  &\scriptsize{Background Model}    &\scriptsize{Fusion (number of papers)}  &\scriptsize{Dependence-independence} &\scriptsize{Authors - Dates} \\
\hline
\hline
\scriptsize{Three Features}     &\scriptsize{\textbf{Color-Intensity-Edge}}     &\scriptsize{}              &\scriptsize{} 
&\scriptsize{}                  &\scriptsize{}   \\
\scriptsize{}     							&\scriptsize{Color-Intensity-Edge}                              &\scriptsize{}              &\scriptsize{OR-AND (1)} 
&\scriptsize{Independent}       &\scriptsize{Huerta et al. \cite{MulF-30}}   \\
\scriptsize{}                   &\scriptsize{Chromatic-invariant cone model-Intensity-Edge}     &\scriptsize{}              &\scriptsize{OR-AND (1)} 
&\scriptsize{Independent}       &\scriptsize{Huerta et al. \cite{MulF-31}}   \\
\scriptsize{}     &\scriptsize{\textbf{Color-Intensity-Texture}}     &\scriptsize{}              &\scriptsize{} 
&\scriptsize{}                  &\scriptsize{}   \\
\scriptsize{}     							&\scriptsize{Color-Intensity-LRP}   &\scriptsize{Codebook \cite{CF-310}}              &\scriptsize{} 
&\scriptsize{Independent}       &\scriptsize{Zaharescu and Jamieson \cite{MulF-40}}   \\
\scriptsize{}                   &\scriptsize{\textbf{Color-Edge-Texture}}     &\scriptsize{}              &\scriptsize{}          
&\scriptsize{}                  &\scriptsize{}   \\
\scriptsize{}                   &\scriptsize{RGB-Gradient-LBP}                &\scriptsize{SG \cite{CF-50}} &\scriptsize{Choquet Integral (1)}  &\scriptsize{Dependent}         &\scriptsize{Azab et al. \cite{FA-30}} \\
\scriptsize{}                   &\scriptsize{RGB-Gradient-Gabor filter}       &\scriptsize{MFRPCA \cite{EnF-4}} &\scriptsize{-} 
&\scriptsize{Independent}       &\scriptsize{Gan et al. (2013) \cite{EnF-4}} \\
\scriptsize{}                   &\scriptsize{RGB-Gradient-Gabor filter}       &\scriptsize{MTRPCA \cite{EnF-5}} &\scriptsize{-} 
&\scriptsize{Independent}       &\scriptsize{Wang and Wan (2014) \cite{EnF-5}} \\
\scriptsize{}                   &\scriptsize{Intensity-Gradient-LBP}       &\scriptsize{Covariance \cite{TF-18-1}} &\scriptsize{-} 
&\scriptsize{Dependent}         &\scriptsize{Zhang et al. (2008) \cite{TF-18}} \\
\scriptsize{}                   &\scriptsize{RGB-Gradient- LBP}             &\scriptsize{Codebook \cite{CF-310}}   &\scriptsize{-}     
&\scriptsize{Independent}         &\scriptsize{Lin et al.  (2016) \cite{CF-345}}   \\
\cline{2-6}
\scriptsize{}                   &\scriptsize{\textbf{Color-Edge-Location (1)}}       &\scriptsize{}                  &\scriptsize{}          
&\scriptsize{}                  &\scriptsize{}   \\
\scriptsize{}                   &\scriptsize{RGB-Gradient-(x,y)}             &\scriptsize{KDE \cite{EnF-1}}   &\scriptsize{}  
&\scriptsize{Independent}       &\scriptsize{Yoo and Kim (2013)\cite{EnF-1}} \\
\cline{2-6}
\scriptsize{}                   &\scriptsize{\textbf{Color-Texture-Motion (1)}}      &\scriptsize{}                     &\scriptsize{}          
&\scriptsize{}                  &\scriptsize{}   \\
\cline{2-6}
\scriptsize{}                   &\scriptsize{RGB-LBP-Optical Flow}        &\scriptsize{Nested model  \cite{EnF-2}}     &\scriptsize{}  
&\scriptsize{Independent}       &\scriptsize{Li et al. (2014) \cite{EnF-2}} \\
\scriptsize{}                   &\scriptsize{\textbf{Color-Texture-Location (2)}}    &\scriptsize{}                     &\scriptsize{}          
&\scriptsize{}                  &\scriptsize{}   \\
\scriptsize{}                   &\scriptsize{RGB-Textons-(x,y)}                      &\scriptsize{KDE  \cite{LF-1}}     &\scriptsize{}  
&\scriptsize{Independent}       &\scriptsize{Spampinato et al. (2014) \cite{TF-210}} \\
\scriptsize{}                   &\scriptsize{Lab-U-LBP-(x,y)}                      &\scriptsize{KDE  \cite{LF-1}}     &\scriptsize{}  
&\scriptsize{Independent}       &\scriptsize{Yan et al. (2016) \cite{TF-114-10}} \\
\scriptsize{}                   &\scriptsize{\textbf{Other features (1)}}      &\scriptsize{}                     &\scriptsize{} 
&\scriptsize{}                  &\scriptsize{}   \\    
\cline{2-6}     
\scriptsize{}                   &\scriptsize{Object size-Direction-Speed}      &\scriptsize{}                     &\scriptsize{}          
&\scriptsize{Independent}       &\scriptsize{Jodoin et al. (2012) \cite{FSI-210}} \\
\hline
\scriptsize{More than Three Features}    &\scriptsize{\textbf{Ensembled based approaches (5)}} &\scriptsize{}                  &\scriptsize{}        &\scriptsize{}     &\scriptsize{}   \\ 
\scriptsize{4 features}       &\scriptsize{YUV-Edge Deviation-Vector Deviation-magnitude}   &\scriptsize{} &\scriptsize{OR (1)}    &\scriptsize{Independent}     &\scriptsize{Kamkar-Parsi et al. (2005) \cite{MulF-7}} \\
\scriptsize{13 features}      &\scriptsize{Color-Edge-Haar Features}   &\scriptsize{MOG \cite{CF-1}} &\scriptsize{Weighted average (1)}    &\scriptsize{Independent}     &\scriptsize{Klare and Sarkar (2009) \cite{FA-2}} \\
\scriptsize{11 features}      &\scriptsize{Color-Gradient-Haar Features}  &\scriptsize{KDA \cite{FR-20}}       &\scriptsize{SVM Classifier (1)}   
&\scriptsize{Independent}     &\scriptsize{Han and Davis (2011) \cite{FR-20}} \\
\scriptsize{24 features}      &\scriptsize{Color-Edge-Texture-Haar Features}   &\scriptsize{Stochastic Approximation \cite{FS-20-1}} &\scriptsize{} 
&\scriptsize{Independent}     &\scriptsize{Lopez-Rubio and Lopez-Rubio (2014) \cite{FS-20}} \\
\scriptsize{24 features}      &\scriptsize{Color-Edge-Texture-Haar Features}   &\scriptsize{FSOM \cite{FS-21-1}} &\scriptsize{} 
&\scriptsize{Independent}     &\scriptsize{Molina-Cabello et al. (2016) \cite{FS-21}} \\
\scriptsize{5 features}       &\scriptsize{Intensity-Color-Edge-Texture-Laplacian Features}  &\scriptsize{OR-PCA with MRF \cite{FS-42}} &\scriptsize{Weighted similarity measure (1)} 
&\scriptsize{Independent}     &\scriptsize{Javed et al. (2015) \cite{FS-42}} \\
\scriptsize{}      &\scriptsize{\textbf{Feature selection approaches (9)}}                &\scriptsize{}                  &\scriptsize{}          
&\scriptsize{}     &\scriptsize{}   \\ 
\scriptsize{11 features}       &\scriptsize{RRGB-Gradients,Color co-occurrences}   &\scriptsize{Bayesian framework (1)} &\scriptsize{} 
&\scriptsize{Independent}     &\scriptsize{Li et al. (2004) \cite{900}} \\
\scriptsize{9 features}       &\scriptsize{RGB-Gradients (Horizontal/Vertical)}   &\scriptsize{KDE (1)} &\scriptsize{} 
&\scriptsize{Independent}     &\scriptsize{Parag et al. (2006) \cite{FS-10}} \\
\scriptsize{9 features}       &\scriptsize{RGB-HSV-YCbCr}   &\scriptsize{ViBe (1)} &\scriptsize{} 
&\scriptsize{Independent}     &\scriptsize{Braham et al. (2015) \cite{FS-50}} \\
\scriptsize{9 features}       &\scriptsize{HOG-LBP-Haar Features}   &\scriptsize{online boosting (2)} &\scriptsize{} 
&\scriptsize{Independent}     &\scriptsize{Grabner and Bischof (2006) \cite{FS-1}} \\
\scriptsize{9 features}       &\scriptsize{HOG-LBP-Haar Features}   &\scriptsize{Time dependent online boosting (1)} &\scriptsize{} 
&\scriptsize{Independent}     &\scriptsize{Grabner et al. (2008) \cite{FS-3}} \\
\scriptsize{9 features}       &\scriptsize{HOG-OCLBP\cite{TF-25}-Haar Features}  &\scriptsize{Hierarchical online boosting (1)}  &\scriptsize{} 
&\scriptsize{Independent}     &\scriptsize{Lee et al. (2011) \cite{TF-25}} \\
\scriptsize{7 features}       &\scriptsize{Color-Gradient-HOG-LBP\cite{TF-10}}  &\scriptsize{OR-PCA (1)}  &\scriptsize{} 
&\scriptsize{Independent}     &\scriptsize{Javed et al. (2015) \cite{FS-40}} \\
\scriptsize{19 features}      &\scriptsize{R,G,B, H,S,V, gray-scale, XCS-LBP, OC-LBP}  &\scriptsize{SVM (1)}  &\scriptsize{} 
&\scriptsize{Independent}     &\scriptsize{} \\
\scriptsize{}                 &\scriptsize{gradient orientation and magnitude, optical flow}  &\scriptsize{}  &\scriptsize{} 
&\scriptsize{Independent}     &\scriptsize{} \\
\scriptsize{}                 &\scriptsize{7 spectral narrow bands}                           &\scriptsize{}  &\scriptsize{} 
&\scriptsize{Independent}     &\scriptsize{} \\
\scriptsize{}                 &\scriptsize{\textbf{Bag of features (1)}} &\scriptsize{} &\scriptsize{}   
&\scriptsize{}     &\scriptsize{}   \\ 
\scriptsize{6 features}       &\scriptsize{Bs-ICR-AS-ARH-LWV-IMV}   &\scriptsize{ViBe \cite{FS-51}} &\scriptsize{}    
&\scriptsize{Independent}     &\scriptsize{Subudhi et al. (2016) \cite{EnF-3}} \\
\hline 
\end{tabular}}
\caption{Multiple Features: An Overview (Part 3).} \centering
\label{MFOverview-3}
\end{table}
\end{landscape}

\begin{table*}
\scalebox{0.90}{
\begin{tabular}{|l|l|l|} 
\hline
\scriptsize{Fusion Scheme} &\scriptsize{Categories} &\scriptsize{Authors}  \\
\hline
\hline
\scriptsize{Classical Operators} &\scriptsize{AND}  &\scriptsize{\textit{Foreground:} Javed et al. \cite{EF-1}, Lin et al. \cite{MulF-4}}   \\
\scriptsize{}                    &\scriptsize{}     &\scriptsize{Ivanov et al. \cite{SF-1}, Ivanov et al. \cite{SF-2}, Ivanov et al. \cite{SF-3} }   \\
\scriptsize{}                    &\scriptsize{}     &\scriptsize{\textit{Distances:} Silvestre \cite{SF-200} }   \\
\scriptsize{}                    &\scriptsize{}     &\scriptsize{\textit{Background:} Fernandez-Sanchez et al. \cite{SF-400}, Fernandez-Sanchez et al.  \cite{SF-401}}                \\
\cline{2-3}
\scriptsize{}                    &\scriptsize{OR} &\scriptsize{\textit{Foreground:} Gordon et al. \cite{SF-11}, Harville et al. \cite{SF-100}}   \\
\scriptsize{}                    &\scriptsize{}   &\scriptsize{Harville et al. \cite{SF-101}, Harville  \cite{SF-102}}                           \\
\scriptsize{}                    &\scriptsize{}   &\scriptsize{Langmann et al.  \cite{SF-201}, Leens et al \cite{SF-204}}                        \\
\cline{2-3}
\scriptsize{}                    &\scriptsize{Combination}  &\scriptsize{\textit{OR-AND-Morphological Operators:} Camplani et al. \cite{SF-301}}   \\
\scriptsize{}                    &\scriptsize{}             &\scriptsize{Ottonelli et al. \cite{SF-700}, Ottonelli et al. \cite{SF-701}}  \\
\scriptsize{}                    &\scriptsize{}             &\scriptsize{}                \\
\cline{2-3}
\scriptsize{}      &\scriptsize{Conditions} &\scriptsize{\textit{Boundaries:} (Edge) Javed et al. \cite{EF-1}, (Edge) Guo and  Yu  \cite{MulF-1}}  \\
\scriptsize{}                    &\scriptsize{}         &\scriptsize{(Texture) Shang et al. \cite{MulF-3}, (Depth) Stormer et al. \cite{SF-203}}   \\
\scriptsize{}                    &\scriptsize{}         &\scriptsize{\textit{Weak/Strong Pixels:} Lai et al. \cite{MulF-2}}                        \\
\scriptsize{}                    &\scriptsize{}         &\scriptsize{\textit{Weak/Strong Motion Vectors:} Zhou and Zhang \cite{MF-17}}             \\
\cline{2-3}
\scriptsize{}         &\scriptsize{Weigthed Average}  &\scriptsize{\textit{Similarities:} Chua et al.  \cite{TF-125}, Chua et al.  \cite{TF-125-1}}  \\
\scriptsize{}         &\scriptsize{}                  &\scriptsize{Ji and Wang \cite{TF-76}}                                                       \\
\scriptsize{}         &\scriptsize{}                  &\scriptsize{\textit{Distances:} Zhang et al. \cite{TF-72}, Yuan et al. \cite{TF-26}}        \\
\scriptsize{}         &\scriptsize{}                  &\scriptsize{Jian and Odobez (2007) \cite{MulF-5}}                                           \\
\scriptsize{}         &\scriptsize{}                  &\scriptsize{Gong and Cheng  (2011) \cite{MF-20}}                                            \\
\hline
\hline
\scriptsize{Statistical  Operators} &\scriptsize{Product of the likelihoods}            &\scriptsize{Zhang et al. \cite{TF-71}}   \\
\cline{2-3}
\scriptsize{}                       &\scriptsize{Weighted average of the likelihoods}   &\scriptsize{Hu et al.    \cite{SF-210}, Camplani and Salgado \cite{SF-300}}   \\
\scriptsize{}                       &\scriptsize{}   &\scriptsize{Camplani et al.  \cite{SF-302}, Camplani et al.  \cite{SF-303}}   \\
\scriptsize{}                       &\scriptsize{}   &\scriptsize{Klare and Sarkar \cite{FA-2}}   \\
\scriptsize{}                       &\scriptsize{}   &\scriptsize{Zhong et al. \cite{TF-19}}      \\
\cline{2-3}
\scriptsize{}                       &\scriptsize{Logarithm Opinion Pool}                &\scriptsize{Gallego and Pardas \cite{SF-500}}   \\
\cline{2-3}
\scriptsize{}                       &\scriptsize{SVM Classifier}                &\scriptsize{Han and Davis \cite{FR-20}}   \\
\hline
\hline
\scriptsize{Fuzzy Operators}      &\scriptsize{Sugeno Integral}     &\scriptsize{Zhang and  Xu \cite{FA-10}, Zhang and  Xu (2006) \cite{FA-10}} \\
\cline{2-3}
\scriptsize{}                     &\scriptsize{Choquet Integral}    &\scriptsize{El Baf et al. \cite{FA-12}, El Baf et al. \cite{FA-13}}        \\
\scriptsize{}                     &\scriptsize{}                    &\scriptsize{El Baf et al. \cite{FA-14}, El Baf et al. \cite{FA-15}}        \\
\scriptsize{}                     &\scriptsize{}                    &\scriptsize{Ding et al. \cite{FA-20}, Ding et al. \cite{FA-21}}            \\
\scriptsize{}                     &\scriptsize{}                    &\scriptsize{Ding et al. \cite{FA-22}, Li et al. \cite{FA-23}}              \\
\scriptsize{}                     &\scriptsize{}                    &\scriptsize{Ding et al. \cite{FA-24}, Azab et al. \cite{FA-30}}            \\
\scriptsize{}                     &\scriptsize{}                    &\scriptsize{Lu et al. \cite{FA-31}, Lu et al. \cite{FA-34}}                \\
\scriptsize{}                     &\scriptsize{}              &\scriptsize{Balcilar et al. \cite{FA-32}, Chiranjeevi and Sengupta \cite{FA-33}} \\
\scriptsize{}                     &\scriptsize{}                    &\scriptsize{Gayathri and Srinivasan \cite{FA-35}, Chiranjeevi and Sengupta \cite{FF-10}}                          \\
\cline{2-3}
\scriptsize{}                     &\scriptsize{Interval valued Choquet integral} &\scriptsize{Chiranjeevi and Sengupta \cite{FA-40}}            \\
\hline 
\end{tabular}}
\caption{Features Fusion: An Overview.} \centering
\label{FFOverview}
\end{table*}

\section{Feature Selection}
\label{sec:FeatureSelection}
Commonly, the background subtration methods do not take into account the properties of each features, and the same feature is used
globally for the whole scene. Indeed, all existing background subtraction methods operate with a uniform feature map in the
sense that the features used to model the background and perform the detection are the same for all pixels of the image, thus ignoring the non-uniformity of the spatial distribution of background properties and neglecting the foreground properties. But in practice for complex scene comprising of several elements such as waving trees, sky, soil and car, the most discriminant features for these elements are probably different, and therefore a single-feature background subtraction algorithm may not be appropriate. \\
\indent Despite the choice of the best features for each region is not an easy task as it requires a deep knowledge of the scene, it is possible  to automatically select the most relevant feature to improve the foreground segmentation in complex scenes thanks to their capability to select a  subset  of  highly discriminant features removing irrelevant and redundant ones. One can find ensemble-based approaches, minimizing the use of traditional feature selection methods (filters, wrappers and embedded) \cite{FS-100}. The objective of such approaches is to generate multiple  feature  selectors  and  then aggregating  their  outputs, producing a more robust classification.  The  main known re-sampling ensemble  methods  that generate  and  combine  a  diversity  of  learners  are:  bagging, boosting and random subspace \cite{FS-101}. The boosting approach and
its variants for feature selection has been used in \cite{FS-1}\cite{FS-2}\cite{FS-3}\cite{TF-25}\cite{FS-10}. However, usually only exemplars of one-class elements are available (i.e. the background component is  always  present), whereas the other classes are unknown (i.e. foreground objects
can appear/disappear several times in the scene). This is known as the one-class classification (OCC) problem, which specific nature is not taken into account in boosting approaches. The OCC approach has been used by Silva et al. \cite{FS-30}. Practically, the different feature selection scheme applied to background modeling and foreground detection can be classified as follows: \\

\begin{itemize}
\item \textbf{Predetermined feature selection:} Li et al. \cite{900} were the first to express the need for modeling distinct part of the image with different features, and described the background image as consisting of two pixel categories, static pixels and dynamic pixels. Unfortunately, this method lacks of generality as \textit{1)} the cardinality of the feature set, i.e. the number of candidate features or group of features that are predetermined, is limited to two (one group of features for static pixels and one other for dynamic pixels), and \textit{2)} the choice of features is not made dynamically but only the type of pixels. \\
\item \textbf{Feature selection via AdaBoost:} Grabner and Bischof \cite{FS-1}\cite{FS-2} introduced an on-line boosting based feature selection framework using the Adaboost algorithm\cite{FS-1-1}. This method was enabled to tackle a large variety of challenges and used large feature pools at reduced computational costs. This gives the feasibility to achieve real-time computational complexity. The feature set contained \textbf{\textit{1)}} the Haar-like features \cite{TDF-3000}, \textbf{\textit{2)}} orientation histograms with $16$ bins similar to Dalal and Triggs\ cite{TDF-4000}, and \textbf{\textit{3)}} a simple version with $4^{th}$ neighborhood of LBP \cite{TF-10}. Grabner and Bischof \cite{FS-1}\cite{FS-2} computed integral images and integral histograms \cite{FS-1-2} as efficient data structures for fast calculation of all features. The current image is divided into overlapping patches, and the patches are used as training data for foreground/background classifiers. As only background parts of the current image can be used as training data, Grabner and Bischof \cite{FS-1}\cite{FS-2} generated an arbitrary image patch, of which the mean pixel value is $128$ and the variance is $256^2/12$, as a foreground patch for initial training. As a new current image is given to the classifiers, the image is divided into patches and each patch is classified as either foreground or background using the trained classifiers. After classification, the patches classified as background with high confidence are used as training data to update the classifiers. The main advantage of this scheme is that it is robust to illumination changes and dynamic backgrounds since the classifiers are consistently updated. However, as it is an off-line learning algorithm, all labelled training samples are required to be available a-priori. In most real applications, training data should not always be sampled from one fixed distribution as it may not encapsulate all the complexities in real-scenes. However, since the method is based
on self-learning (i.e. classifier predictions feed model updates), the background model can end up in catastrophic state, as mentioned by Grabner and Bischof \cite{FS-3}. Therefore, Grabner and Bischof \cite{FS-3} proposed a time-dependent on-line boosting algorithm using exponential forgetting of the samples over time. Nevetherless, these two algorithms naturally lose the advantage of color information and still suffers from large computational complexity \cite{TF-25}. Thus, Lee et al. \cite{TF-25} proposed a hierarchial on-line boosting which uses block based Opponent Color Local Binary Pattern (OC-LBP). Thus, Lee et al. \cite{TF-25} divided a patch into three layers
of R,G, and B, and then each layer into $3\time 3$ blocks. $9$ different block based OC-LBP are generated by placing different colors to the center block and the surrounding blocks. For each block based OC-LBP, $8$ classifiers are constructed. Then, Lee et al. \cite{TF-25} considered $9$ classifier pools, each of which includes $8$ classifiers. While the original on-line boosting \cite{FS-1}\cite{FS-2}\cite{FS-3} should generate arbitrary foreground patches, the hierarchical on-line boosting  \cite{TF-25} only needs background patches. The computational time is shortened significantly by making the process of training and classification simple. As pointed out by Braham et al. \cite{FS-50}, the drawbacks of these approaches are unrealistic assumptions about the statistical distributions of foreground features that are used, i.e. a uniform distribution is assumed for the gray value of foreground objects and serves as a basis for computing other feature distributions.\\
\item \textbf{Feature selection via RealBoost:} Parag et al. \cite{FS-10} proposed a generic model that is capable of automatically selecting the features that obtain the best invariance to the background changes while maintaining a high detection rate for the foreground detection. In this study, Parag et al. \cite{FS-10} proposed the use of a Realboost algorithm.  Unlike Adaboost algorithm which combines weak hypotheses
having outputs in $\left\{-1, +1\right\}$, RealBoost algorithm computes real-valued weak classifiers given real numbered feature values, and generates a linear combination of these weak classifiers that minimizes the training error. To generate the background model, Elgammal et al. \cite{FS-10} used the Kernel Density estimation (KDE). Using the density estimates, the Realboost algorithm \cite{FS-10-1} is able to select the features most appropriate for any pixel. In the implementation phase, Parag et al. \cite{FS-10} used $9$ types of features, namely three color values R,G,B and spatial derivatives for each of these color channels in both $x$ and $y$ directions for each pixel of a color image. For intensity images, $3$ features are used, that are the pixel intensity value and its spatial gradients in horizontal and vertical directions. Experimental results \cite{FS-10} show that this approach outperforms the original KDE \cite{204} by a slight margin. While this framework can be used with an arbitrary number of candidate features, it has a serious drawback as pointed out by Braham et al. \cite{FS-50}. Indeed, the synthetic foreground examples used for boosting are generated randomly from a uniform distribution. Indeed, all candidate features are assumed to be uniformly distributed
for foreground objects. This assumption is not valid because of the wide variety of foreground statistical distributions among different features \cite{FS-50}. For example, gradient has a foreground probability density function concentrated around low values while RGB color components have a wider foreground probability density function. This may explain why the classification performance of this framework decreases when gradient features are added to the feature set. This limitation can be addressed by means of global statistical foreground models.\\
\item \textbf{Dynamic feature selection strategy:}  Javed et al. \cite{FS-40} used a dynamic feature selection strategy with Online Robust Principal Component Analysis (OR-PCA) framework. Feature statistics, in terms of means and variances, are used as a selection criterion. Unlike aforementioned purely spatial or spatiotemporal selection approaches, this method is exclusively temporal-based, which means that all pixels use the same features for the foreground segmentation. The non-uniformity of the spatial distribution of background properties is thus ignored. \\
\item \textbf{Feature selection via a generic method:} Like in Parag et al. \cite{FS-10}, Braham et al. \cite{FS-50} proposed that the feature selection process only occurs during the training phase, to avoid extra computations during normal background subtraction operations. Thus, the training phase is divided into three parts: \textit{1)} The first part consists to accumulate images free of foreground objects which are further processed to build local background statistical models, \textit{2)} The second part uses an other sequence of images, which include foreground objects and use to build a global foreground statistical model for each candidate feature of the feature set, and \textit{3)} the third part selects the best feature/threshold combination which is a local process, meaning that it is performed for each pixel individually. The goal of the selection process consists to detect which feature is most appropriate to discriminate between the local background and the foreground at the frame level.
Unlike previous approaches, the foreground feature statistical distributions are estimated at the frame level. Features are selected locally depending on their capability to discriminate between local background samples and global foreground samples for a specific background model and for a chosen performance metric. Experiments results \cite{FS-50} show that this feature selection scheme significantly improves the performance of ViBe \cite{FS-51}. \\
\item \textbf{Feature selection via One-class SVM:} Silva et al. \cite{FS-30} used an Incremental  Weighted  One-Class Support Vector
Machine (IWOC-SVM) which  select suitable features for each pixel to distinguish the foreground objects from the background. In addition, an  Online and Weighted version of the Random Subspace  (OW-RS) \cite{FS-30} is used to assign a degree  of  importance to each feature set, and  these weights are used directly  in the training step  of the  IWOC-SVM. Moreover, a heuristic  approach is used to
reduce the complexity of the background model maintenance while  maintaining the robustness  of the background model. Practically, the  features were chosen to have at least one feature in the five type of features: color  feature(R,G,B, H,S,V and gray-scale), texture feature (XCS-LBP \cite{TF-63}), color-texture (OC-LBP \cite{TF-63-1}), edge feature (gradient orientation and magnitude), and motion feature (optical flow). In  addition, multispectral  bands (a total of 7 spectral narrow bands) are also used. Experimental results on multispectral video sequences from the MSVS dataset \cite{906} show the pertinence of this selection scheme.
\end{itemize}
In these different methods, the feature selection is made at a different step: \textit{1)} in off-line step in a predetermined way (as in Li et al. \cite{900}) which is less adaptive over time because the set of features for each category of pixel is fixed over time and it only requires additional computation for the detection of the category of each pixel , \textit{2)} in the training step only (as in Parag et al. \cite{FS-10} and in Braham et al. \cite{FS-50}) which is more robust than the off-line way with a set of selected features more flexible and thus more diverse for each pixel, or \textit{3)} in an online way all over the process (as in Grabner and Bischof \cite{FS-1}, Javed et al. \cite{FS-40}, and Silva et al. \cite{FS-30}) which is the most adaptive scheme but it requires additional computation time. 

\begin{table*}
\scalebox{0.70}{
\begin{tabular}{|l|l|l|l|l|} 
\hline
\scriptsize{Algorithms} &\scriptsize{Background Model} &\scriptsize{Number} &\scriptsize{Features} &\scriptsize{Authors-Dates}  \\
\hline
\hline
\scriptsize{\textbf{Off-line feature selection}}  &\scriptsize{}      &\scriptsize{}       &\scriptsize{}  
&\scriptsize{}   \\
\scriptsize{Predetermined}  &\scriptsize{Bayesian framework \cite{900}}      &\scriptsize{5}       &\scriptsize{Static Background Pixels: RGB-Gradients}  &\scriptsize{Li et al. (2004) \cite{900}}   \\
\scriptsize{}  &\scriptsize{}      &\scriptsize{6}     &\scriptsize{Dynamic Background Pixels: Color co-occurrences}  &\scriptsize{}   \\
\hline
\scriptsize{\textbf{Feature selection in the training step}}  &\scriptsize{}      &\scriptsize{}       &\scriptsize{}  
&\scriptsize{}   \\
\scriptsize{Realboost \cite{FS-10-1}} &\scriptsize{KDE \cite{204}}     &\scriptsize{3}      &\scriptsize{Intensity-Gradients (Horizontal/Vertical)}   &\scriptsize{Parag et al. (2006) \cite{FS-10}}   \\
\scriptsize{Realboost \cite{FS-10-1}} &\scriptsize{KDE \cite{204}}     &\scriptsize{9}      &\scriptsize{RGB-Gradients (Horizontal/Vertical)}   &\scriptsize{Parag et al. (2006) \cite{FS-10}}   \\
\scriptsize{Generic method} &\scriptsize{ViBe \cite{FS-51}}            &\scriptsize{9}      &\scriptsize{RGB-HSV-YCbCr}   &\scriptsize{Braham et al. (2015) \cite{FS-50}} \\
\hline
\scriptsize{\textbf{On-line feature selection }}  &\scriptsize{}      &\scriptsize{}       &\scriptsize{}  
&\scriptsize{}   \\
\scriptsize{Adaboost \cite{FS-1-1}}  &\scriptsize{Single Gaussian \cite{CF-50}}      &\scriptsize{9}       &\scriptsize{Haar Features, HoG \cite{TDF-4000}, LBP \cite{TF-10}}  
&\scriptsize{Grabner and Bischof (2006) \cite{FS-1}}   \\
\scriptsize{Adaboost \cite{FS-1-1}}  &\scriptsize{Single Gaussian \cite{CF-50}}      &\scriptsize{9}       &\scriptsize{Haar Features, HoG \cite{TDF-4000}, LBP \cite{TF-10}}  
&\scriptsize{Grabner et al. (2006) \cite{FS-2}}   \\
\scriptsize{Time dependent Adaboost \cite{FS-1-1}}  &\scriptsize{Single Gaussian \cite{CF-50}}   &\scriptsize{9}   &\scriptsize{Haar Features, HoG \cite{TDF-4000}, LBP \cite{TF-10}}  
&\scriptsize{Grabner et al. (2008) \cite{FS-3}}   \\
\scriptsize{Hierarchical Adaboost \cite{FS-1-1}}  &\scriptsize{Single Gaussian \cite{CF-50}}   &\scriptsize{9}     &\scriptsize{Haar Features, HoG \cite{TDF-4000}, OCLBP \cite{TF-63-1}}  
&\scriptsize{Lee et al. (2011) \cite{TF-25}}   \\
\scriptsize{Selection criterion}                   &\scriptsize{OR-PCA \cite{FS-40}}  		   &\scriptsize{7}    &\scriptsize{Color-Gradient-HOG-LBP\cite{TF-10}}   &\scriptsize{Javed et al. (2015) \cite{FS-40}}   \\
\scriptsize{One-class SVM \cite{FS-30}}            &\scriptsize{SVM}                     &\scriptsize{19}      &\scriptsize{R,G,B, H,S,V, gray-scale, XCS-LBP, OC-LBP}   &\scriptsize{Silva et al. (2016) \cite{FS-30}}   \\
\scriptsize{ } &\scriptsize{}     &\scriptsize{}      &\scriptsize{gradient orientation and magnitude, optical flow}   &\scriptsize{}   \\
\scriptsize{ } &\scriptsize{}     &\scriptsize{}      &\scriptsize{7 spectral narrow bands}   &\scriptsize{}   \\
\hline 
\end{tabular}}
\caption{Features Selection: An Overview.} \centering
\label{FSOverview}
\end{table*}

\section{Resources, datasets and codes}
\label{sec:Resources}

\subsection{Features Web Site}
This website contains a full list of references in the field, links to available datasets and codes. In each case, the list
is regularly updated and classified following the type of features as in this paper. An overview of the content of the
Feature Website is given at the home page. In addition to the sections which concern the type of features, this website gives references and links to available implementations, datasets, conferences, workshops and websites.

\subsection{Datasets}
Several datasets available to evaluate and compare background subtraction algorithms have been developed in the last decade. We classified them in terms of IR datasets, color datasets and RGB-D datasets. IR and color datasets provide videos from one camera in IR or color space. On the other hand, RGB-D datasets provide color+depth videos obtained by stereo cameras or the Microsoft Kinect RGB-D camera or the Asus Xtion Pro Live camera. All these datasets are publicly available and their links are provided on the Features Web Site in the section Available Datasets.

\subsubsection{IR dataset}
The OTCBVS 2006 Dataset is related to the conference "Object Tracking and Classification in and
Beyond the Visible Spectrum" (OTCVBVS\protect\footnotemark[28]) contains sequences for person detection and face detection. Three sequences are then interesting for background subtraction: (1) Dataset 01 (OSU Thermal Pedestrian) which concerns person detection in thermal imagery, (2) Dataset 03 (OSU Color-Thermal Database) on fusion-based object detection in color and thermal imagery and (3) Dataset 05 (Terravic Motion IR Database ) which focus on detection and tracking with thermal imagery.

\subsubsection{Color datasets}
Color datasets are the majority in the available datasets and the reader can find the full list in Bouwmans \cite{JournalCOSREV2014-1}. Here, we focus on the recent large-scale datasets and the datasets in aquatic environments.  \\
\begin{enumerate}
\item \textbf{Large-scale datasets:} These datasets are realistic large-scale datasets with accurate ground-truth providing a balanced coverage of the range of challenges present in the real world. Three large-scale datasets are available and very interesting: \\
\begin{itemize}
\item \textbf{ChangeDetection.net Dataset:}  The CDW\protect\footnotemark[5] 2012 dataset \cite{901} presents a realistic, large-scale video dataset consisting of nearly 90,000 frames in 31 video sequences representing 6 categories selected to cover a wide range of challenges in 2 modalities (color and thermal IR). The main characteristic of this dataset is that each frame is annotated for ground-truth foreground, background, and shadow area boundaries. This allows an objective and precise quantitative comparison of background subtraction algorithms. This dataset was extended in 2014 to the CDW\protect\footnotemark[6] 2014 dataset which contains all the 2012 videos plus additional ones with the following challenges: challenging weather, low frame-rate, acquisition at night, PTZ capture and air turbulence. \\
\item \textbf{BMC 2012 Dataset:} The BMC\protect\footnotemark[7] (Background Models Challenge) \cite{903}\cite{903-1} is a workshop organized within ACCV (Asian Conference in Computer Vision) about the comparison of background subtraction techniques with both synthetic and real videos. This benchmark is first composed of a set of 20 synthetic video sequences with the corresponding ground truth, frame by frame, for each video (at 25 fps). The first part of 10 synthetic videos are devoted to the learning phase of the proposed algorithms, while the 10 others are used for evaluation.  BMC also contains 9 real videos acquired from static cameras in video-surveillance contexts for evaluation. 
This dataset has been built in order to test the algorithms' reliability during a certain time and in difficult situations such as outdoor scenes. 
Furthermore, this dataset allows us to test the influence of some difficulties encountered during the foreground detection step, as the presence of waving trees, cast shadows or sudden illumination changes in the scene. \\
\item \textbf{SABS Dataset:} The SABS\protect\footnotemark[8] (Stuttgart Artificial Background Subtraction) dataset \cite{902} represents an artificial dataset for pixel-wise evaluation of background models. Synthetic image data generated by modern ray-tracing makes realistic high quality ground-truth data available. The dataset consists of video sequences for nine different challenges of background subtraction for video surveillance. These sequences are further split into training and test data. For every frame of each test sequence ground-truth annotation is provided as color-coded foreground masks. This way, several foreground objects can be distinguished and the ground-truth annotation could also be used for tracking evaluation. The dataset contains additional shadow annotation that represents for each pixel the absolute luminance distance between the frame with and without foreground objects. The sequences have a resolution of $800 \times 600$ pixels and are captured from a fixed viewpoint. \\
\end{itemize}
\item \textbf{Datasets in aquatic environments} These datasets are very interesting to test and evaluate illumination invariant color features and texture features due to the dynamic background changes and illumination changes particular to aquatic environments. Three datasets are publicly available: \\
\begin{itemize}
\item \textbf{Fish4knowledge Dataset:} The Fish4knowledge \protect\footnotemark[9] dataset \cite{233}\cite{907}\cite{2072} is an underwater benchmark dataset for target detection against complex background which consists of 14 videos categorized into seven different classes representing complex challenges in background modeling. \\
\item \textbf{Aqu@theque Dataset:} The Aqu@theque\protect\footnotemark[10] dataset \cite{NETTIES2006}\cite{SIGMAP2007}\cite{IWSSIP2007} contains four different image sequences from the "Aqu@thèque" project \cite{2}, and one sequence from the experimental study of turbulent flow. The fish need to be well detected in order to allow features extraction which are used to recognize the species of the fish. One sequence is used for the experimental study of turbulent flow in vertical slot fishways to optimize their protection. Each of the five sequences shows different situations and challenges such as bootstrapping, occlusion, camouflage, light changes and dynamic background. \\
\item \textbf{MAR Dataset:} Maritime Activity Recognition (MAR\protect\footnotemark[11]) \cite{2062} is a dataset containing data coming from different video sources (fixed and Pan-Tilt-Zoom cameras) and from different scenarios. There are 10 videos from fixed cameras with ground-truth images and 15 form PTZ cameras. The aim of this dataset is to provide a set of videos that can be used to develop intelligent surveillance system for maritime environment. \\
\end{itemize}
\end{enumerate}

\subsubsection{RGB-D datasets}
Three RGB-D datasets are publicly available and can be described as follows: \\
\begin{itemize}
\item \textbf{RGB-D Object Detection Dataset:} The RGB-D\protect\footnotemark[12] dataset \cite{SF-300}\cite{SF-301} provides five sequences of indoor environments, acquired with the \textit{Microsoft Kinect RGB-D camera}. Each sequence contains different challenges such as cast shadows, color and depth camouflage. For each sequence a hand-labeled ground truth is provided. \\
\item \textbf{CITIC RGB-D Dataset:} The CITIC\protect\footnotemark[13] dataset \cite{SF-400}\cite{SF-401} contains sequences recorded with rectified \textit{stereo cameras}, and some frames have been hand-segmented to provide ground-truth information. \\
\item \textbf{RGB-D Rigid Multi-Body Dataset:} The RGB-D Rigid Multi-Body \protect\footnotemark[14] dataset \cite{9071} consists of 3 RGB-D videos of objects with different sizes (chairs, box/watering can, small box/teacan). The datasets have been recorded using an \textit{Asus Xtion Pro Live camera} in a resolution of 640x480 at 30 Hz frame rate. Ground truth for the camera pose has been obtained with an OptiTrack Motion Capture system. The moving objects are also manually annotated in frames at every 5 seconds. \\
\end{itemize}

\footnotetext[5]{http://www.changedetection.net/}
\footnotetext[6]{http://www.changedetection.net/}
\footnotetext[7]{http://bmc.univ-bpclermont.fr}
\footnotetext[8]{http://www.vis.uni-stuttgart.de/index.php?id=sabs}
\footnotetext[9]{http://www.fish4knowledge.eu}
\footnotetext[10]{http://sites.google.com/site/thierrybouwmans/recherche---aqu-theque-dataset}
\footnotetext[11]{http://labrococo.dis.uniroma1.it/MAR/}
\footnotetext[12]{http://www.gti.ssr.upm.es/~mac/}
\footnotetext[13]{http://atcproyectos.ugr.es/mvision/}
\footnotetext[14]{http://www.ais.uni-bonn.de/download/rigidmultibody/}

\subsection{LBP Library}
The LBP Library provided by Caroline Silva, Cristina Lazar and Andrews Sobral is a collection of eleven Local Binary Patterns (LBP) algorithms developed for background subtraction problem. The algorithms were implemented in C++ based on OpenCV. A CMake file is provided and the library is complatible with Windows, Linux and Mac OS X. The library was tested successfully with OpenCV 2.4.10.  The local texture pattern available are the following ones: original LBP \cite{TF-10}, CS-LBP \cite{TF-33}, SCS-LBP \cite{TF-21}, $\epsilon$-LBP \cite{TF-20}, HCS-LBP \cite{TF-23}, OCLBP \cite{TF-25}, VARLBP \cite{TF-69}, BG-LBP \cite{TF-66}, XCS-LBP \cite{TF-63}, SILTP \cite{TF-70} and CS-SILTP \cite{TF-77}. \\

\section{Conclusion}
\label{sec:Conclusion}
In conclusion, this review on the role and the importance of features for background modeling and foreground detection highlights the following points: \\
\begin{itemize}
\item Features can be classified following their size, their type in a specific domain, their intrinsic properties and their mathematical concepts. Each type of features presents different robustness against challenges met in videos taken by a fixed cameras. For the color feature, YCrCb color space seems to be the more suitable feature \cite{FA-13}\cite{CF-203}. For the texture feature, Silva et al. \cite{FS-30} provided a study on the LBP and its variants that show that XCS-LBP is the best LBP feature for this application in presence of illumination changes and dynamic backgrounds.
Although this study covered texture features, it is restricted to LBP features and then there is not a full study on the different texture features. For the depth feature, it needs to carefully used them following their properties as developed in Nghiem and Bremond \cite{SF-750}. Features in a domain transform are very useful to reduce computation times as in the case of compressive sensing features.\\
\item Several features have been used in other applications and none in background modeling and foreground detection such several variants of LBP  (Multi-scale Region Perpendicular LBP (MRP-LBP) \cite{TF-1000}, Scale-and Orientation Adaptive LBP (SOA-LBP) \cite{TF-1010}). Furthermore, statistical or fuzzy version of crisp feature could be investigated such as histograms of fuzzy oriented gradients \cite{StF-135}. It would be interesting to evaluate them for this application. \\
\item Because each feature has its strenghtness and weakness against each challenge, multiple features schemes are used to combine the advantages of their different robustness. Most of the time, gradient, texture, motion and stereo features are used in addition to the color feature to deal with camouflage in color, illumination changes, dynamic backgrounds and shadows. Different fusion operators can be used to combine these different features but fuzzy integrals such as the Choquet integral \cite{FA-12} and interval-valued Choquet \cite{FA-40} seem the best way to aggregate different features because dependency between features can be taken into account.\\
\item Because there is not a unique feature that performs better than any other feature independently of the background and foreground properties, feature selection allows to use the best feature or the best combination of features. Experimental results provided by the existing approaches show the pertinence of feature selection in background modeling and foreground detection. However, basic algorithms such as Adaboost and Realboost have  been used most of the time. The most advanced scheme is the IWOC-SVM algorithm developed by Silva et al. \cite{FS-30}, but more advanced selection schemes can be used such as statistical or fuzzy feature selection. 
\end{itemize}
Finally, the best approach seems to fuse multiple features with the intervalued fuzzy Choquet integral. The best set of features seems to be
illumination invariant color features combined with spatio-temporal texture features and depth features. Future research should concern \textbf{1)} a full evaluation of texture features, \textbf{2)} a full comparison of feature fusion schemes, \textbf{3)} feature selection schemes and \textbf{4)} reliability of features because it has been less investigated.
\section{Acknowledgment}

\bibliographystyle{plain}
\bibliography{mybib,myfeatures,mychangedetection,myrpca,mysparse,mypublication}
\end{document}